\documentclass[10pt,twocolumn,letterpaper]{article}

\usepackage{cvpr}
\usepackage{times}
\usepackage{epsfig}
\usepackage{graphicx}
\usepackage{amsmath}
\usepackage{xcolor}
\usepackage{amssymb}
\usepackage{url}
\usepackage{ownstyles}
\usepackage{array}
\usepackage{cellspace}
\usepackage{subcaption}
\usepackage{float}
\usepackage{footmisc}

\newif\ifcomments

\commentstrue

\ifcomments
\newcommand{\comments}[1]{#1}
\else
\newcommand{\comments}[1]{}
\fi

\newcommand{\layer}[1]{\ensuremath{\mathsf{#1}\xspace}}
\newcommand{\unit}[1]{\ensuremath{\mathsf{#1}\xspace}}

\newcommand{\thinplus}{\hspace*{-.3ex}+\hspace*{-.3ex}}
\newcommand{\thinminus}{\hspace*{-.3ex}-\hspace*{-.3ex}}
\newcommand{\thineq}{\hspace*{-.3ex}=\hspace*{-.3ex}}

\newcommand{\papertitle}{Plug \& Play Generative Networks:\\Conditional Iterative Generation of Images in Latent Space}
\title{\papertitle}

\usepackage[pagebackref=true,breaklinks=true,letterpaper=true,colorlinks,bookmarks=false]{hyperref}

\cvprfinalcopy 

\def\cvprPaperID{1822} 

\DefineFNsymbols{mySymbols}{{\ensuremath\dagger}{\ensuremath\ddagger}\S\P
   *{**}{\ensuremath{\dagger\dagger}}{\ensuremath{\ddagger\ddagger}}}
\setfnsymbol{mySymbols}

\begin{document}

\newcommand{\spacehackt}{.2cm}
\newcommand{\spacehackh}{.5cm}
\addtolength{\topmargin}{-\spacehackt}
\addtolength{\textheight}{\spacehackh}




\author{
	Anh Nguyen\\
	\small{University of Wyoming$^\dagger$}\\
	{\tt\small anh.ng8@gmail.com}
	\and
	Jeff Clune\\
	\small{Uber AI Labs$^\dagger$, University of Wyoming}\\
	{\tt\small jeffclune@uwyo.edu}
	\and
	Yoshua Bengio\\
	\small{Montreal Institute for Learning Algorithms}\\
	{\tt\small yoshua.umontreal@gmail.com}
	\and
	Alexey Dosovitskiy\\
	\small{University of Freiburg}\\
	{\tt\small dosovits@cs.uni-freiburg.de}
	\and
	Jason Yosinski\\
	\small{Uber AI Labs\thanks{This work was mostly performed at Geometric Intelligence, which Uber acquired to create Uber AI Labs.}}\\
	{\tt\small yosinski@uber.com}
}


\maketitle


\begin{abstract}
Generating high-resolution, photo-realistic images has been a long-standing goal in machine learning. Recently, Nguyen et al. \cite{nguyen-2016-synthesizing-the-preferred-inputs} showed one interesting way to synthesize novel images by performing gradient ascent in the latent space of a generator network to maximize the activations of one or multiple neurons in a separate classifier network.  In this paper we extend this method by introducing an additional prior on the latent code, improving both sample quality and sample diversity, leading to a state-of-the-art generative model that produces high quality images at higher resolutions ($227\times227$) than previous generative models, and does so for all 1000 ImageNet categories.
In addition, we provide a unified probabilistic interpretation of related activation maximization methods and call the general class of models ``Plug and Play Generative Networks.'' PPGNs are composed of 1) a generator network G that is capable of drawing a wide range of image types and 2) a replaceable ``condition'' network C that tells the generator what to draw.
We demonstrate the generation of images conditioned on a class (when C is an ImageNet or MIT Places classification network) and also conditioned on a caption (when C is an image captioning network). Our method also improves the state of the art of 	Multifaceted Feature Visualization \cite{nguyen-2016-arXiv-multifaceted-feature-visualization:}, which generates the set of synthetic inputs that activate a neuron in order to better understand how deep neural networks operate. 
Finally, we show that our model performs reasonably well at the task of image inpainting.
While image models are used in this paper, the approach is modality-agnostic and can be applied to many
types of data.
\end{abstract}

\begin{figure}
	\centering
	\includegraphics[width=1.0\columnwidth]{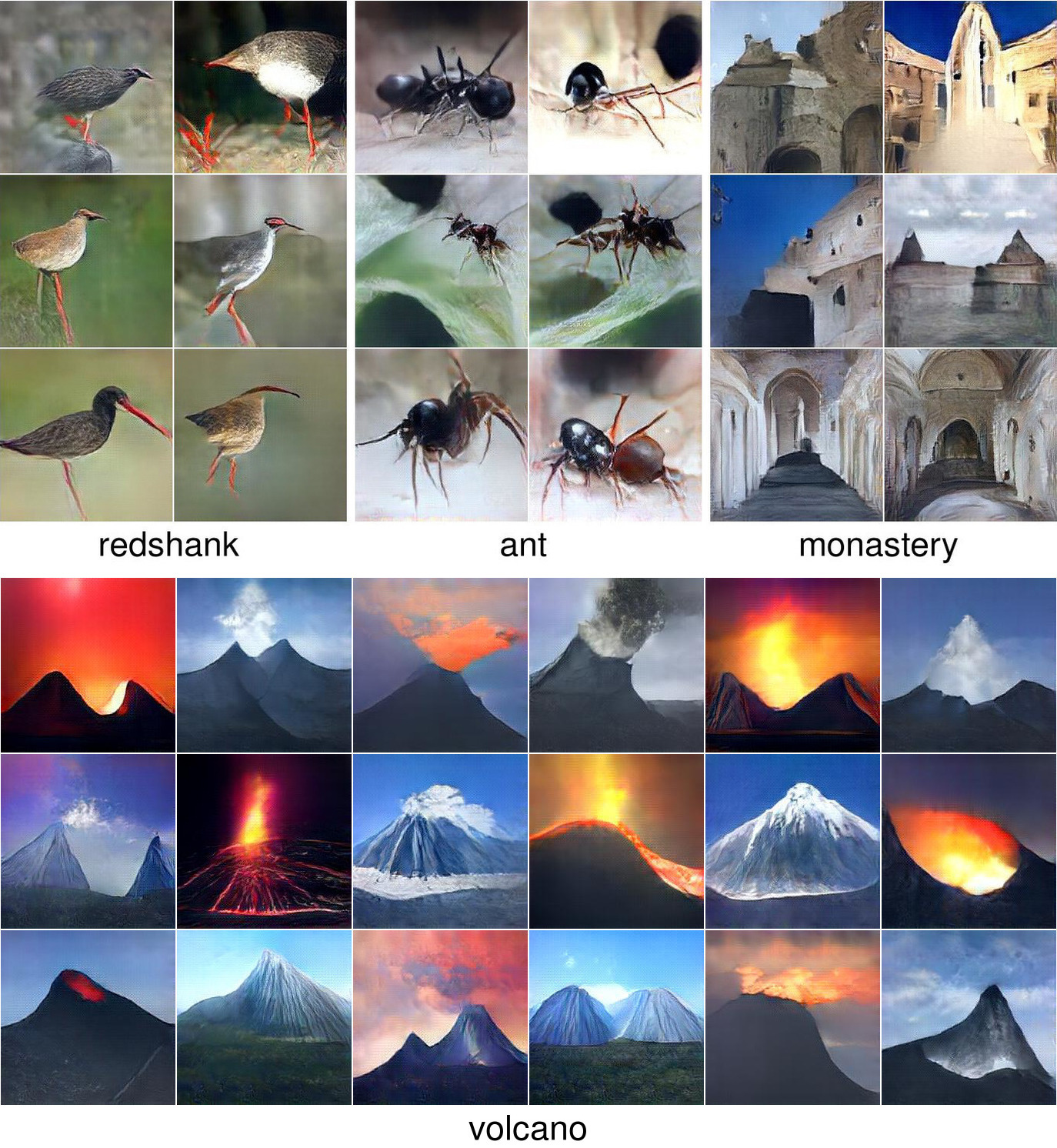}
	\vspace*{-0.4em}
	\caption{
	  Images synthetically generated by Plug and Play Generative Networks at high-resolution (227x227) for four ImageNet classes. Not only are many images nearly photo-realistic, but samples within a class are diverse.
	}
	\figlabel{teaser}
	\vspace*{-0.9em}
\end{figure}


\vspace*{-1.7em}
\section{Introduction}

\begin{figure*}
  (a) Real: top 9                                                 \hspace{4.8em}
  (b) DGN-AM \cite{nguyen-2016-synthesizing-the-preferred-inputs} \hspace{3.8em}
  (c) Real: random 9                                               \hspace{4.4em}
  (d) PPGN (\textbf{this})
  \centering
	\includegraphics[width=2.0\columnwidth]{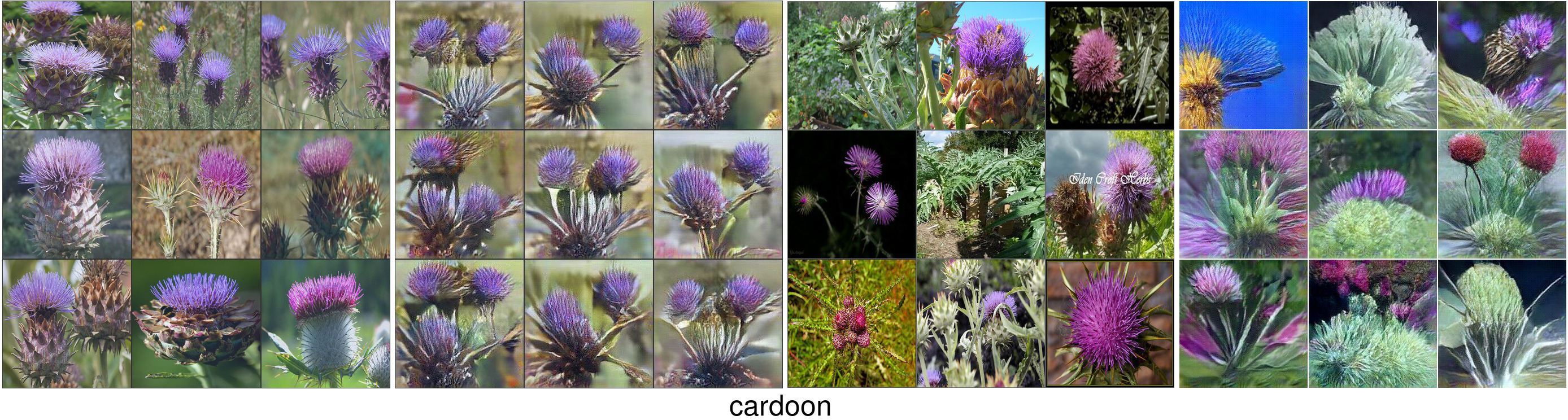}
	\caption{
		For the ``cardoon'' class neuron in a pre-trained ImageNet classifier, we show: a) the 9 real training set images that most highly activate that neuron; b) images synthesized by DGN-AM \cite{nguyen-2016-synthesizing-the-preferred-inputs}, which are of similar type and diversity to the real top-9 images; c) random real training set images in the cardoon class; and d) images synthesized by PPGN, which better represent the diversity of random images from the class. \figref{DGNAM_vs_PPGN} shows the same four groups for other classes.
	}
	\figlabel{cardoon}
	\vspace*{-.5em}
\end{figure*}

Recent years have seen generative models that are increasingly capable of synthesizing diverse, realistic images that capture both the fine-grained details and global coherence of natural images~\cite{van-den-oord-2016-arXiv-pixel-recurrent-neural,larochelle-2011-AISTATS-the-neural-autoregressive-distribution,dosovitskiy-2016-NIPS-generating-images-with,gregor2015draw,radford-2015-arXiv-unsupervised-representation-learning,kingma-2013-arXiv-auto-encoding-variational-bayes}. However, many important open challenges remain, including (1) producing photo-realistic images at high resolutions \cite{ledig2016photo}, (2) training generators that can produce a wide variety of images (e.g. all 1000 ImageNet classes) instead of only one or a few types (e.g. faces or bedrooms \cite{radford-2015-arXiv-unsupervised-representation-learning}), and (3) producing a diversity of samples that match the diversity in the dataset instead of modeling only a subset of the data distribution \cite{goodfellow2014generative-adversarial-networks, theis-2016-ICLR-a-note-on-the-evaluation-of-generative}. Current image generative models often work well at low resolutions (e.g. $32\times32$), but struggle to generate high-resolution (e.g. $128\times128$ or higher), globally coherent images (especially for datasets such as ImageNet \cite{deng2009imagenet:-a-large-scale-hierarchical} that have a large variability \cite{odena-2016-arXiv-conditional-image-synthesis,salimans-2016-arXiv-improved-techniques-for-training,goodfellow2014generative-adversarial-networks}) due to many challenges including difficulty in training \cite{salimans-2016-arXiv-improved-techniques-for-training,odena-2016-arXiv-conditional-image-synthesis} and computationally expensive sampling procedures \cite{van-den-oord-2016-arXiv-pixel-recurrent-neural,van-den-oord-2016-arXiv-conditional-image-generation}.

\addtolength{\topmargin}{\spacehackt}
\addtolength{\textheight}{-\spacehackh}



Nguyen et al.  \cite{nguyen-2016-synthesizing-the-preferred-inputs} recently introduced a technique that produces high quality images at a high resolution. Their Deep Generator Network-based Activation Maximization\footnote{ Activation maximization is a technique of searching via optimization for the synthetic image that maximally activates a target neuron in order to understand which features that neuron has learned to detect \cite{erhan2009visualizing-higher-layer-features}.} (DGN-AM) involves training a generator $G$ to create realistic images from compressed features extracted from a pre-trained classifier network $E$ (\figref{concept}f). To generate images conditioned on a class, an optimization process is launched to find a hidden code $h$ that $G$ maps to an image that highly activates a neuron in another classifier $C$ (not necessarily the same as $E$). Not only does DGN-AM produce realistic images at a high resolution (Figs.~\ref{fig:cardoon}b \& \ref{fig:DGNAM_vs_PPGN}b), but, without having to re-train $G$, it can also produce interesting new types of images that $G$ never saw during training. For example, a $G$ trained on ImageNet can produce ballrooms, jail cells, and picnic areas if $C$ is trained on the MIT Places dataset~(\figref{mit_places_DGNAM_vs_PPGN}, top).

A major limitation with DGN-AM, however, is the lack of diversity in the generated samples.
While samples may vary slightly (e.g. ``cardoons'' with two or three flowers viewed from slightly different angles; see \figref{cardoon}b), the whole image tends to have the same composition (e.g. a closeup of a single cardoon plant with a green background). It is noteworthy that the images produced by DGN-AM closely match the images from that class that most highly activate the class neuron (\figref{cardoon}a).
Optimization often converges to the same mode even with different random initializations, a phenomenon common with activation maximization \cite{erhan2009visualizing-higher-layer-features,nguyen-2016-arXiv-multifaceted-feature-visualization:,wei2015understanding}.
In contrast, real images within a class tend to show more diversity (\figref{cardoon}c).
In this paper, we improve the diversity and quality of samples produced via DGN-AM by adding a prior on the latent code that keeps optimization along the manifold of realistic-looking images (\figref{cardoon}d).

We do this by providing a probabilistic framework in which to unify and interpret
activation maximization approaches \cite{simonyan2013deep-inside-convolutional,yosinski-2015-ICML-DL-understanding-neural-networks,nguyen-2016-arXiv-multifaceted-feature-visualization:,nguyen-2016-synthesizing-the-preferred-inputs} as a type of  \emph{energy-based} model \cite{Goodfellow-et-al-2016-Book,lecun2006tutorial} where the energy function is a sum of multiple constraint terms: (a) priors (e.g. biasing images to look realistic) and (b) conditions, typically given as a category of a separately trained classification model (e.g. encouraging images to look like ``pianos'' or both ``pianos'' and ``candles''). We then show how to sample iteratively from such models using an approximate Metropolis-adjusted Langevin sampling algorithm.

We call this general class of models Plug and Play Generative Networks (PPGN). The name reflects an important, attractive property of the method: one is free to design an energy function, and ``plug and play'' with different priors and conditions to form a new generative model.
This property has recently been shown to be useful in multiple image generation projects that use
the DGN-AM generator network prior
and swap in different condition networks \cite{zhou2016places,yahoo2016nsfw}.
In addition to generating images conditioned on a class, PPGNs can generate images conditioned on text, forming a text-to-image generative model that allows one to describe an image with words and have it synthesized. We accomplish this by attaching a recurrent, image-captioning network (instead of an image classification network) to the output of the generator, and performing similar iterative sampling. 
Note that, while this paper discusses only the image generation domain, the approach should generalize to many other data types.
We publish our code and the trained networks at \url{http://EvolvingAI.org/ppgn}.

\section{Probabilistic interpretation of iterative image generation methods}




Beginning with the Metropolis-adjusted Langevin algorithm \cite{roberts1996exponential,roberts-1998-JRSS-optimal-scaling-of-discrete} (MALA), it is possible to define a Markov chain Monte Carlo (MCMC) sampler whose stationary distribution approximates a given distribution $p(x)$. We refer to our variant of MALA as \emph{MALA-approx}, which uses the following transition operator:\footnote{ We abuse notation slightly in the interest of space and denote as $N(0, \epsilon_3^2)$ a sample from that distribution. The first step size is given as $\epsilon_{12}$ in anticipation of later splitting into separate $\epsilon_1$ and $\epsilon_2$ terms.}

\vspace*{-.5em}
\beq
x_{t+1} = x_t + \epsilon_{12}\nabla\log p(x_t) + N(0, \epsilon_3^2)
\eqnlabel{mala-approx}
\eeq

\noindent A full derivation and discussion is given in \secref{samplers}.
Using this sampler we first derive a probabilistically interpretable
formulation for activation maximization methods (\secref{mcmc_interpretation}) and then
interpret other activation maximization algorithms in this framework (\secref{previous_models}, \secref{previous_models_si}).


\subsection{Probabilistic framework for Activation\\Maximization}
\seclabel{mcmc_interpretation}

Assume we wish to
sample from a joint model
$p(x, y)$,
which can be decomposed into an image model and a classification model:

\beq
p(x, y) = p(x) p(y | x)
\eqnlabel{joint_x_y}
\eeq


This equation can be interpreted as a ``product of experts'' \cite{hinton1999products} in which each expert determines whether a soft constraint is satisfied.
First, a $p(y|x)$ expert determines a condition for image generation (e.g. images have to be classified as ``cardoon'').
Also, in a high-dimensional image space, a good $p(x)$ expert is needed to ensure the search stays in the manifold of image distribution that we try to model (e.g. images of faces \cite{brock2016neural,yeh2016semantic}, shoes \cite{zhu2016generative} or natural images \cite{nguyen-2016-synthesizing-the-preferred-inputs}),
otherwise we might encounter ``fooling'' examples that are unrecognizable but have high $p(y|x)$
\cite{nguyen-2015-CVPR-deep-neural-networks,szegedy2013intriguing-properties-of-neural}.
Thus, $p(x)$ and $p(y|x)$ together impose a complicated high-dimensional constraint on image generation.

We could write a sampler for the full joint $p(x, y)$, but because $y$ variables are categorical, suppose for now that we fix $y$ to be a particular chosen class $y_c$, with $y_c$ either sampled or chosen outside the inner sampling loop.\footnote{ One could resample $y$ in the loop as well, but resampling $y$ via the Langevin family under consideration is not a natural fit: because $y$ values from the data set are one-hot -- and from the model hopefully nearly so -- there will be a wide small- or zero-likelihood region between $(x, y)$ pairs coming from different classes. Thus making local jumps will not be a good sampling scheme for the $y$ components.
}
This leaves us with the conditional $p(x | y)$:
\vspace{-0.3cm}
\begin{align}
p(x | y = y_c) &=       p(x) p(y = y_c | x) / p(y = y_c) \nonumber \\
               &\propto p(x) p(y = y_c | x)
\eqnlabel{decomposition}
\end{align}

%
%
%

We can construct a MALA-approx sampler for this model, which produces the following update step:

\vspace{-0.3cm}
\begin{align}
x_{t+1} &= x_t + \epsilon_{12}\nabla\log p(x_t |  y = y_c) + N(0, \epsilon_3^2) \nonumber \\
        &= x_t \thinplus \epsilon_{12}\nabla\log p(x_t ) \thinplus \epsilon_{12}\nabla\log p(y \thineq y_c | x_t) \thinplus N(0, \epsilon_3^2)
\end{align}

\noindent Expanding the $\nabla$ into explicit partial derivatives and decoupling $\epsilon_{12}$ into explicit $\epsilon_1$ and $\epsilon_2$ multipliers, we arrive at the following form of the update rule:

\beq
x_{t+1} = x_t + \epsilon_1\frac{\partial\log p(x_t)}{\partial{x_t}}  +  \epsilon_2\frac{\partial \log p(y = y_c | x_t)}{\partial x_t} + N(0, \epsilon_3^2)
\eqnlabel{update_rule}
\eeq

We empirically found that decoupling the $\epsilon_1$ and $\epsilon_2$ multipliers works better. An intuitive interpretation of the actions of these three terms is as follows:

\begin{itemize}
  \vspace*{-.5em}
  \item $\epsilon_1$ term: take a step from the current image $x_t$ toward one that looks more like a generic image (an image from any class).
  \vspace*{-.5em}
  \item $\epsilon_2$ term: take a step from the current image $x_t$ toward an image that causes the classifier to output higher confidence in the chosen class. The $p(y = y_c | x_t)$ term is typically modeled by the softmax output units of a modern convnet, e.g. AlexNet \cite{krizhevsky2012imagenet-classification-with-deep} or VGG \cite{simonyan-2014-arXiv-very-deep-convolutional}.
  \vspace*{-.5em}
  \item $\epsilon_3$ term: add a small amount of noise to jump around the search space to encourage a diversity of images.
  \vspace*{-.5em}
\end{itemize}

\begin{figure*}
	\centering
	\includegraphics[width=2.0\columnwidth]{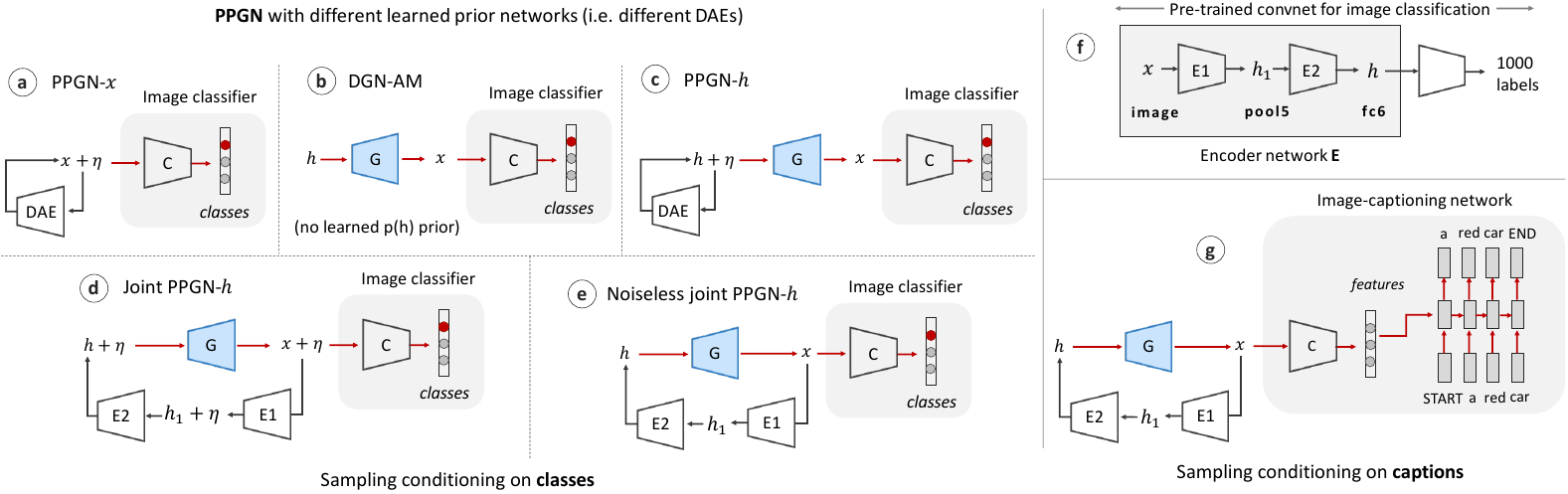}
	\caption{
		Different variants of PPGN models we tested. The Noiseless Joint PPGN-$h$ (e), which we found empirically produces the best images, generated the results shown in \figsref{teaser}{cardoon} \& Sections \ref{sec:noiseless_ppgn} \& \ref{sec:results}. In all variants, we perform iterative sampling following the gradients of two terms: the condition (red arrows) and the prior (black arrows).
		\textbf{(a)} PPGN-$x$ (\secref{ppgn_x}): To avoid fooling examples \cite{nguyen-2015-CVPR-deep-neural-networks} when sampling in the high-dimensional image space, we incorporate a $p(x)$ prior modeled via a denoising autoencoder (DAE) for images, and sample images conditioned on the output classes of a condition network $C$ (or, to visualize hidden neurons, conditioned upon the activation of a hidden neuron in $C$).
		\textbf{(b)} DGN-AM (\secref{dgnam}): Instead of sampling in the image space (i.e. in the space of individual pixels), Nguyen et al. \cite{nguyen-2016-synthesizing-the-preferred-inputs} sample in the abstract, high-level feature space $h$ of a generator $G$ trained to reconstruct images $x$ from compressed features $h$ extracted from a pre-trained encoder $E$ (f).
Because the generator network was trained to produce realistic images, it serves as a prior on $p(x)$ since it ideally can only generate real images. However, this model has no learned prior on $p(h)$ (save for a simple Gaussian assumption).
		\textbf{(c)} PPGN-$h$ (\secref{ppgn_h}): We attempt to improve the mixing speed and image quality by incorporating a learned $p(h)$ prior modeled via a multi-layer perceptron DAE for $h$.
		\textbf{(d)} Joint PPGN-$h$ (\secref{default_ppgn}): 
		To improve upon the poor data modeling of the DAE in PPGN-$h$, we experiment with treating $G+E_1+E_2$ as a DAE that models $h$ via $x$. In addition, to possibly improve the robustness of $G$, we also add a small amount of noise to $h_1$ and $x$ during training and sampling, treating the entire system as being composed of 4 interleaved models that share parameters: a GAN and 3 interleaved DAEs for $x$, $h_1$ and $h$, respectively.
		This model mixes substantially faster and produces better image quality than DGN-AM and PPGN-$h$ (\figref{sampling_class_from_random}).
		\textbf{(e)} Noiseless Joint PPGN-$h$ (\secref{noiseless_ppgn}): We perform an ablation study on the Joint PPGN-$h$, sweeping across noise levels or loss combinations, and found a Noiseless Joint PPGN-$h$ variant trained with one less loss (\secref{train_PPGN}) to produce the best image quality.
		\textbf{(f)} A pre-trained image classification network (here, AlexNet trained on ImageNet) serves as the encoder network $E$ component of our model by mapping an image $x$ to a useful, abstract, high-level feature space $h$ (here, AlexNet's fc6 layer). 
		\textbf{(g)} Instead of conditioning on classes, we can generate images conditioned on a caption by attaching a recurrent, image-captioning network to the output layer of $G$, and performing similar iterative sampling.
	}
	\figlabel{concept}
\end{figure*}

\subsection{Interpretation of previous models}
\seclabel{previous_models}

Aside from the errors introduced by not including a reject step,
the stationary distribution of the sampler in \eqnref{update_rule} will converge to the appropriate distribution if the $\epsilon$ terms are chosen appropriately \cite{welling2011bayesian}.
Thus, we can use this framework to interpret previously proposed iterative methods for generating samples, evaluating whether each method faithfully computes and employs each term.

There are many previous approaches that iteratively sample from a trained model to generate images
\cite{simonyan2013deep-inside-convolutional,yosinski-2015-ICML-DL-understanding-neural-networks,nguyen-2016-arXiv-multifaceted-feature-visualization:,nguyen-2016-synthesizing-the-preferred-inputs,wei-2015-understanding-intra-class-knowledge,arulkumaran2016improving,erhan2009visualizing-higher-layer-features,yeh2016semantic,zhu2016generative,brock2016neural,nguyen-2015-GECCO-innovation-engines:-automated,nguyen-2015-CVPR-deep-neural-networks,mahendran2016visualizing}, with methods designed for different purposes such as
activation maximization
\cite{simonyan2013deep-inside-convolutional,yosinski-2015-ICML-DL-understanding-neural-networks,nguyen-2016-arXiv-multifaceted-feature-visualization:,nguyen-2016-synthesizing-the-preferred-inputs,wei-2015-understanding-intra-class-knowledge,erhan2009visualizing-higher-layer-features,nguyen-2015-CVPR-deep-neural-networks,mahendran2016visualizing} or generating realistic-looking images by sampling in the latent space of a generator network \cite{yeh2016semantic,nguyen-2016-synthesizing-the-preferred-inputs,zhu2016generative,brock2016neural,arulkumaran2016improving,han2016alternating}. However, most of them are gradient-based, and can be interpreted as a variant of MCMC sampling from a graphical model \cite{koller2009probabilistic}. 

While an analysis of the full spectrum of approaches is outside this paper's scope, we do examine a
few representative approaches under this framework in \secref{previous_models_si}.
In particular, we interpret the models that lack a $p(x)$ image prior, yielding adversarial or fooling examples
\cite{szegedy2013intriguing-properties-of-neural,nguyen-2015-CVPR-deep-neural-networks} as setting $(\epsilon_1, \epsilon_2, \epsilon_3) = (0, 1, 0)$; and methods that use $L_2$ decay during sampling as using a Gaussian $p(x)$ prior with $(\epsilon_1, \epsilon_2, \epsilon_3) = (\lambda, 1, 0)$.
Both lack a noise term and thus sacrifice sample diversity.

\section{Plug and Play Generative Networks}

Previous models are often limited in that they use hand-engineered priors when sampling in either image space or the latent space of a generator network (see \secref{previous_models_si}).
In this paper, we experiment
with 4 different explicitly learned priors modeled by a denoising autoencoder
(DAE) \cite{vincent-2008-ICML-extracting-and-composing-robust}. 

We choose a DAE because,
although it does not allow evaluation of $p(x)$ directly, it \emph{does} allow approximation of the gradient
of the log probability when trained with Gaussian noise with
variance $\sigma^2$ \cite{alain-2014-what-regularized-auto-encoders};
with sufficient capacity and training time, the approximation is perfect in the limit as $\sigma \to 0$:

\beq
\frac{\partial \log p(x)}{\partial x} \approx \frac{R_x(x) - x}{\sigma^2}
\label{eqn:rx_x}
\eeq

\noindent where $R_x$ is the reconstruction function in $x$-space representing the DAE, i.e. $R_x(x)$ is a ``denoised'' output of the autoencoder $R_x$ (an encoder followed by a decoder) when the encoder is fed input $x$.
This term approximates exactly the $\epsilon_1$ term required by our sampler, so we can use
it to define the steps of a sampler for an image $x$ from class $c$. Pulling the $\sigma^2$ term into $\epsilon_1$, the update is: 

\beq
x_{t+1} = x_t + \epsilon_1\big(R_x(x_t) - x_t\big)  +  \epsilon_2\frac{\partial \log p(y = y_c | x_t)}{\partial x_t} + N(0, \epsilon_3^2)
\label{eqn:update_rule_DAE}
\eeq



\subsection{PPGN-$x$: DAE model of $p(x)$}
\label{sec:ppgn_x}

First, we tested using a DAE to model $p(x)$ directly (\figref{concept}a) and sampling from the entire model via Eq.~\ref{eqn:update_rule_DAE}. However, we found that
PPGN-$x$
exhibits two expected problems: (1) it models the data distribution poorly; and (2) the chain mixes slowly. More details are in Sec.~\ref{sec:SI_ppgn_x}.

\subsection{DGN-AM: sampling without a learned prior}
\label{sec:dgnam}

Poor mixing in the high-dimensional pixel space of PPGN-$x$ is consistent with previous observations that mixing on higher layers can result in faster exploration of the space \cite{bengio-2013-ICML-better-mixing-via-deep,luo-2013-AIStats-texture-modeling-with}.
Thus, to ameliorate the problem of slow mixing, we may reparameterize $p(x)$ as $\int_h p(h)p(x|h) dh$ for some latent $h$, and perform sampling in this lower-dimensional $h$-space.

While several recent works had success with this approach \cite{nguyen-2016-synthesizing-the-preferred-inputs,brock2016neural,yeh2016semantic}, they often hand-design the $p(h)$ prior. 
Among these, the DGN-AM method \cite{nguyen-2016-synthesizing-the-preferred-inputs} searches in the latent space of a generator network $G$ to find a code $h$ such that the image $G(h)$ highly activates a given neuron in a target DNN.
We start by reproducing their results for comparison.
$G$ is trained following the methodology in Dosovitskiy \& Brox \cite{dosovitskiy-2016-NIPS-generating-images-with} with an $L_2$ image reconstruction loss, a Generative Adversarial Networks (GAN) loss \cite{goodfellow2014generative-adversarial-networks}, and an $L_2$ loss in a feature space $h_1$ of an encoder $E$ (\figref{concept}f). The last loss encourages generated images to match the real images in a high-level feature space and is referred to as ``feature matching'' \cite{salimans-2016-arXiv-improved-techniques-for-training} in this paper, but is also known as ``perceptual similarity'' \cite{larsen-2015-arXiv-autoencoding-beyond-pixels,dosovitskiy-2016-NIPS-generating-images-with} or a form of ``moment matching'' \cite{li2015generative}.  Note that in the GAN training for $G$, we simultaneously train a discriminator $D$ to tell apart real images $x$ vs. generated images $G(h)$. More training details are in \secref{train_PPGN}.


The directed graphical model interpretation of DGN-AM is
$h \to x \to y$ (see \figref{concept}b) and the joint $p(h,x,y)$ can be decomposed into:

\beq
p(h,x,y) = p(h) p(x|h) p(y|x)
\eqnlabel{joint_h_x_y}
\eeq

\noindent where $h$ in this case represents features extracted from the first fully connected layer (called \layer{fc6}) of a pre-trained AlexNet \cite{krizhevsky2012imagenet-classification-with-deep} 1000-class ImageNet \cite{deng2009imagenet:-a-large-scale-hierarchical} classification network (see \figref{concept}f). $p(x|h)$ is modeled by $G$, an upconvolutional (also ``deconvolutional'') network \cite{dosovitskiy-2015-CVPR-learning-to-generate-chairs} with 9 upconvolutional and 3 fully connected layers. $p(y|x)$ is modeled by C, which in this case is also the AlexNet classifier. The model for $p(h)$ was an implicit unimodal Gaussian implemented via L2 decay in $h$-space \cite{nguyen-2016-synthesizing-the-preferred-inputs}. 


Since $x$ is a deterministic variable, the model simplifies to:

\beq
p(h,y) = p(h)p(y|h)
\eqnlabel{joint_h_y}
\eeq


From \eqnref{update_rule}, if we define a Gaussian $p(h)$ centered at 0 and set $(\epsilon_1, \epsilon_2, \epsilon_3) = (\lambda, 1, 0)$, pulling Gaussian constants into $\lambda$, we obtain the following noiseless update rule in Nguyen et al. \cite{nguyen-2016-synthesizing-the-preferred-inputs} to sample $h$ from class $y_c$:



\vspace{-0.3cm}
\begin{align}
h_{t+1} &= (1-\lambda)h_t + \epsilon_2\frac{\partial \log p(y = y_c | h_t)}{\partial h_t} \nonumber \\
&= (1-\lambda)h_t + \epsilon_2\frac{\partial \log C_c(G(h_t))}{\partial G(h_t)}\frac{\partial G(h_t)}{\partial h_t}
\label{eqn:dgnam_update_rule}
\end{align}
where $C_c(\cdot)$ represents the output unit associated with class $y_c$.
As before, all terms are computable in a single forward-backward pass. More concretely, to compute the $\epsilon_2$ term, we push a code $h$ through the generator $G$ and condition network $C$ to the output class $c$ that we want to condition on (\figref{concept}b, red arrows), and back-propagate the gradient via the same path to $h$.
The final $h$ is pushed through $G$ to produce an image sample.

Under this newly proposed framework, we have successfully reproduced the original DGN-AM results and their issue of converging to the same mode when starting from different random initializations (\figref{cardoon}b). 
We also found that DGN-AM mixes somewhat poorly, yielding the same image after many sampling steps (\figsref{sampling_DGNAM_image}{sampling_DGNAM_random}).

\subsection{PPGN-$h$: Generator and DAE model of $p(h)$}
\label{sec:ppgn_h}


We attempt to address the poor mixing speed of DGN-AM by incorporating a proper $p(h)$ prior learned via a DAE into the sampling procedure described in \secref{dgnam}. Specifically, we train $R_h$, a 7-layer, fully-connected DAE on $h$ (as before, $h$ is a \layer{fc6} feature vector).
The size of the hidden layers are respectively:
$4096 - 2048 - 1024 - 500 - 1024 - 2048 - 4096$. Full training details are provided in \ref{sec:train_h_DAE}.

The update rule to sample $h$ from this model is similar to \eqnref{dgnam_update_rule} except for the inclusion of all three $\epsilon$ terms:



\vspace{-0.3cm}
\begin{align}
  h_{t+1} &= h_t
  \thinplus \epsilon_1(R_h(h_t) \thinminus h_t) \nonumber
  \thinplus \epsilon_2\frac{\partial \log C_c(G(h_t))}{\partial G(h_t)}\frac{\partial G(h_t)}{\partial h_t} \nonumber \\
  &\quad\thinplus N(0, \epsilon_3^2)
\eqnlabel{ppgn_update_rule}
\end{align}



Concretely, to compute $R_h(h_t)$ we push $h_t$ through the learned DAE, encoding and decoding it (\figref{concept}c, black arrows). The $\epsilon_2$ term is computed via a forward and backward pass through both $G$ and $C$ networks as before (\figref{concept}c, red arrows). Lastly, we add the same amount of noise $N(0,{\epsilon^2_3})$ used during DAE training to $h$. Equivalently, noise can also be added before the encode-decode step.

We sample\footnote{ If faster mixing or more stable samples are desired, then $\epsilon_1$ and $\epsilon_3$ can be scaled up or down together. Here we scale both down to $10^{-5}$.}
using $(\epsilon_1, \epsilon_2, \epsilon_3) = (10^{-5}, 1, 10^{-5})$
and show results
in \figsref{sampling_h_image}{sampling_h_random}.
As expected, the chain mixes faster than PPGN-$x$, with subsequent samples appearing more qualitatively different from their predecessors.
However, the samples for PPGN-$h$ are qualitatively similar to those from DGN-AM (\figsref{sampling_DGNAM_image}{sampling_DGNAM_random}).
Samples still lack quality and diversity, which we hypothesize is due to the poor $p(h)$ model learned by the DAE.

\subsection{Joint PPGN-$h$: joint Generator and DAE}
\label{sec:default_ppgn}

The previous result suggests that the simple multi-layer perceptron DAE poorly modeled the distribution of \layer{fc6} features.
This could occur because the DAE faces the generally difficult unconstrained density estimation problem.
To combat this issue, we experiment with modeling $h$ via $x$ with a DAE: $h \to x \to h$.
Intuitively, to help the DAE better model $h$, we force it to generate realistic-looking images $x$ from features $h$ and then decode them back to $h$.
One can train this DAE from scratch separately from $G$ (as done for PPGN-$h$). However, in the DGN-AM formulation,
$G$ models the $h \to x$ (\figref{concept}b) and $E$ models the $x \to h$ (\figref{concept}f). Thus, the composition $G(E(.))$ can be considered an AE $h \to x \to h$ (\figref{concept}d).

Note that $G(E(.))$ is theoretically not a formal $h$-DAE because its two components were trained with neither noise added to $h$ nor an $L_2$ reconstruction loss for $h$ \cite{nguyen-2016-synthesizing-the-preferred-inputs} (more details in \secref{train_PPGN}) as is required for regular DAE training \cite{vincent-2008-ICML-extracting-and-composing-robust}. To make $G(E(.))$ a more theoretically justifiable $h$-DAE, we add noise to $h$ and train $G$ with an additional reconstruction loss for $h$ (\figref{training_diagram}c). 
We do the same for $x$ and $h_1$ (\layer{pool5} features), hypothesizing that a little noise added to $x$ and $h_1$ might encourage $G$ to be more robust \cite{vincent-2008-ICML-extracting-and-composing-robust}.
In other words, with the same existing network structures from DGN-AM \cite{nguyen-2016-synthesizing-the-preferred-inputs}, we train $G$ differently by treating the entire model as being composed of 3 interleaved DAEs that share parameters: one each for $h$, $h_1$, and $x$ (see \figref{training_diagram}c). 
Note that $E$ remains frozen, and $G$ is trained with 4 losses in total i.e. three $L_2$ reconstruction losses for $x$, $h$, and $h_1$ and a GAN loss for $x$. See \secref{train_PPGN_noise} for full training details. We call this the \emph{Joint PPGN-$h$} model.



We sample from this model following the update rule in \eqnref{ppgn_update_rule} with $(\epsilon_1, \epsilon_2)= (10^{-5}, 1)$, and with noise added to all three variables: $h$, $h_1$ and $x$ instead of only to $h$ (\figref{concept}d vs e). The noise amounts added at each layer are the same as those used during training. As hypothesized, we observe that the sampling chain from this model mixes substantially faster and produces samples with better quality than all previous PPGN treatments (\figsref{sampling_inter_image}{sampling_inter_random}) including PPGN-$h$, which has a multi-layer perceptron $h$-DAE.


\subsection{Ablation study with Noiseless Joint PPGN-$h$}
\label{sec:noiseless_ppgn}

While the Joint PPGN-$h$ outperforms all previous treatments in sample quality and diversity (as the chain mixes faster), the model is trained with a combination of four losses and noise added to all variables. This complex training process can be difficult to understand, making further improvements non-intuitive. To shed more light into how the Joint PPGN-$h$ works, we perform ablation experiments which later reveal a better-performing variant.

\textbf{Noise sweeps.} To understand the effects of adding noise to each variable, we train variants of the Joint PPGN-$h$ (1) with different noise levels, (2) using noise on only a single variable, and (3) using noise on multiple variables simultaneously. We did not find these variants to produce qualitatively better reconstruction results than the Joint PPGN-$h$. Interestingly, in a PPGN variant trained with no noise at all,
the $h$-autoencoder given by $G(E(.))$ still appears to be contractive,
i.e. robust to a large amount of noise (\figref{noise_robustness}). This is beneficial during sampling; if ``unrealistic'' codes appear, $G$ could map them back to realistic-looking images. We believe this property might emerge for multiple reasons: (1) $G$ and $E$ are not trained jointly; (2) $h$ features encode global, high-level rather than local, low-level information; (3) the presence of the adversarial cost when training $G$ could make the $h \to x$ mapping more ``many-to-one'' by pushing $x$ towards modes of the image distribution.

\textbf{Combinations of losses.} To understand the effects of each loss component, we repeat the Joint PPGN-$h$ training (\secref{default_ppgn}), but without noise added to the variables. Specifically, we test different combinations of losses and compare the quality of images $G(h)$ produced by pushing the codes $h$ of real images through $G$ (without MCMC sampling).

First, we found that removing the adversarial loss from the 4-loss combination yields blurrier images (\figref{no_gan}). Second, we compare 3 different feature matching losses: \layer{fc6}, \layer{pool5}, and both \layer{fc6} and \layer{pool5} combined, and found that \layer{pool5} feature matching loss leads to the best image quality (\secref{compare_losses}). Our result is consistent with Dosovitskiy \& Brox \cite{dosovitskiy-2016-NIPS-generating-images-with}.  Thus, the model that we found empirically to produce the best image quality is trained without noise and with three losses: a \layer{pool5} feature matching loss, an adversarial loss, and an image reconstruction loss. We call this variant ``Noiseless Joint PPGN-$h$'': it produced the results in \figsref{teaser}{cardoon} and Sections \ref{sec:noiseless_ppgn} \& \ref{sec:results}.

\textbf{Noiseless Joint PPGN-$h$.} We sample from this model with $(\epsilon_1, \epsilon_2, \epsilon_3) = (10^{-5}, 1, 10^{-17})$ following the same update rule in \eqnref{ppgn_update_rule} (we need noise to make it a proper sampling procedure, but found that infinitesimally small noise produces better and more diverse images, which is to be expected given that the DAE in this variant was trained without noise). Interestingly, the chain mixes substantially faster than DGN-AM (\figsref{sampling_PPGN_image}{sampling_DGNAM_image}) although the only difference between two treatments is the existence of the learned $p(h)$ prior. Overall, the Noiseless Joint PPGN-$h$ produces a large amount of sample diversity (\figref{cardoon}). Compared to the Joint PPGN-$h$, the Noiseless Joint PPGN-$h$ produces better image quality, but mixes slightly slower (\figsref{sampling_class_from_image}{sampling_class_from_random}). Sweeping across the noise levels during sampling, we noted that larger noise amounts often results in worse image quality, but not necessarily faster mixing speed (\figref{noise_sweep}). Also, as expected, a small $\epsilon_1$ multiplier makes the chain mix faster, and a large one pulls the samples towards being generic instead of class-specific (\figref{prior_sweep}).

\textbf{Evaluations.} Evaluating image generative models is challenging, and there is not yet a commonly accepted quantitative performance measure \cite{theis-2016-ICLR-a-note-on-the-evaluation-of-generative}. We qualitatively evaluate sample diversity of the Noiseless Joint PPGN-$h$ variant by running 10 sampling chains, each for 200 steps, to produce 2000 samples, and filtering out samples with class probability of less than $0.97$. From the remaining, we randomly pick 400 samples and plot them in a grid t-SNE \cite{van-der-maaten2008visualizing-data-using} (\figsref{tsne_pooltable}{tsne_volcano}). More examples for the reader's evaluation of sample quality and diversity are provided in Figs.~\ref{fig:diversity_random_images},~\ref{fig:many_random} \&~\ref{fig:60_gen_vs_60_real}.
To better observe the mixing speed, we show videos of sampling chains (with one sample per frame; no samples filtered out) from within classes and between 10 different classes at
\url{https://goo.gl/36S0Dy}. In addition, Table~\ref{table:quantitative} provides quantitative comparisons between PPGN, auxiliary classifier GAN \cite{odena-2016-arXiv-conditional-image-synthesis} and real ImageNet images in image quality (via Inception score \cite{salimans-2016-arXiv-improved-techniques-for-training} \& Inception accuracy \cite{odena-2016-arXiv-conditional-image-synthesis}) and diversity (via MS-SSIM metric \cite{odena-2016-arXiv-conditional-image-synthesis}).

While future work is required to fully understand why the Noiseless Joint PPGN-$h$ produces high-quality images at a high resolution for 1000-class ImageNet more successfully than other existing latent variable models~\cite{odena-2016-arXiv-conditional-image-synthesis,salimans-2016-arXiv-improved-techniques-for-training,radford-2015-arXiv-unsupervised-representation-learning}, we discuss possible explanations in Sec.~\ref{sec:discuss_ppgn}.

%


\section{Additional results}
\label{sec:results}

In this section, we take the Noiseless Joint PPGN-$h$ model and show its capabilities on several different tasks.

\subsection{Generating images with different condition\\ networks}

A compelling property that makes PPGN different from other existing generative models is that one can ``plug and play'' with different prior and condition components (as shown in \eqnref{joint_x_y}) and ask the model to perform new tasks, including challenging the generator to produce images that it has never seen before. Here, we demonstrate this feature by replacing the $p(y|x)$ component with different networks.

\vspace*{.5em}
\noindent\textbf{Generating images conditioned on classes} 

Above we showed that PPGN could generate a diversity of high quality samples for ImageNet classes (\figsref{teaser}{cardoon} \& \secref{noiseless_ppgn}). Here, we test whether the generator $G$ within the PPGN can generalize to new types of images that it has never seen before. Specifically, we sample with a different $p(y|x)$ model: an AlexNet DNN \cite{krizhevsky2012imagenet-classification-with-deep} trained to classify 205 categories of scene images from the MIT Places dataset \cite{zhou-2014-arXiv-object-detectors-emerge}. Similar to DGN-AM \cite{nguyen-2016-synthesizing-the-preferred-inputs}, the PPGN generates realistic-looking images for classes that the generator was never trained on, such as ``alley'' or ``hotel room'' (\figref{mit_places}). A side-by-side comparison between DGN-AM and PPGN are in \figref{mit_places_DGNAM_vs_PPGN}.

\begin{figure}[t]
	\centering
	\includegraphics[width=1.0\columnwidth]{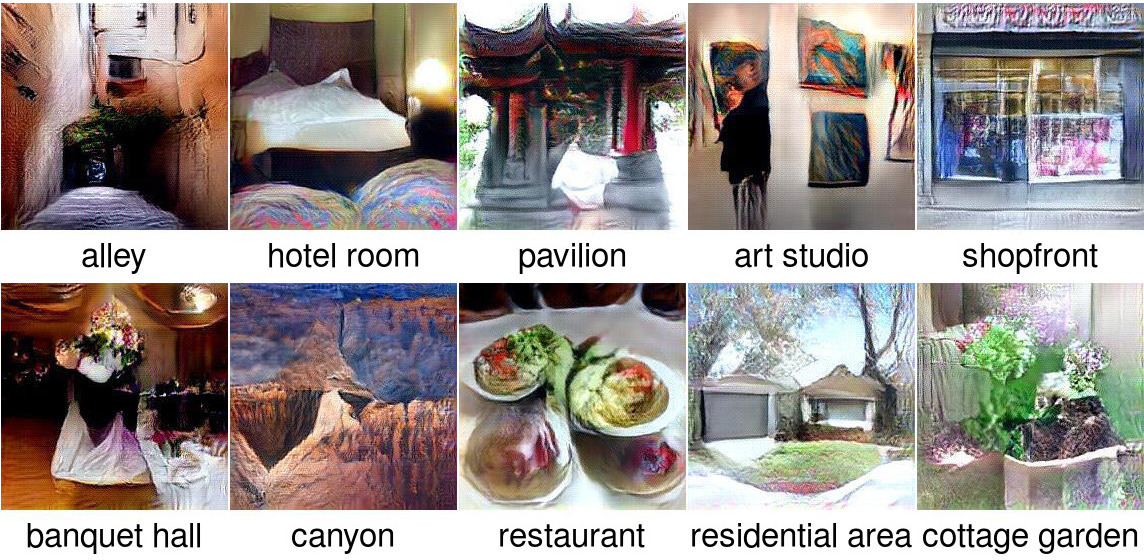}
	\caption{
		Images synthesized conditioned on MIT Places \cite{zhou-2014-arXiv-object-detectors-emerge} classes instead of ImageNet classes.
	}
	\figlabel{mit_places}
        \vspace*{-0.5em}
\end{figure}

\vspace*{.5em}
\noindent\textbf{Generating images conditioned on captions} 

Instead of conditioning on classes, we can also condition the image generation on a caption (\figref{concept}g). Here, we swap in an image-captioning recurrent network (called LRCN) from \cite{donahue-2014-arXiv-long-term-recurrent-convolutional} that was trained on the MS COCO dataset \cite{lin2014microsoft} to predict a caption $y$ given an image $x$.
Specifically, LRCN is a two-layer LSTM network that generates captions conditioned on features extracted from the output softmax layer of AlexNet \cite{krizhevsky2012imagenet-classification-with-deep}.

\begin{figure}[h]
	\centering
	\includegraphics[width=1.0\columnwidth]{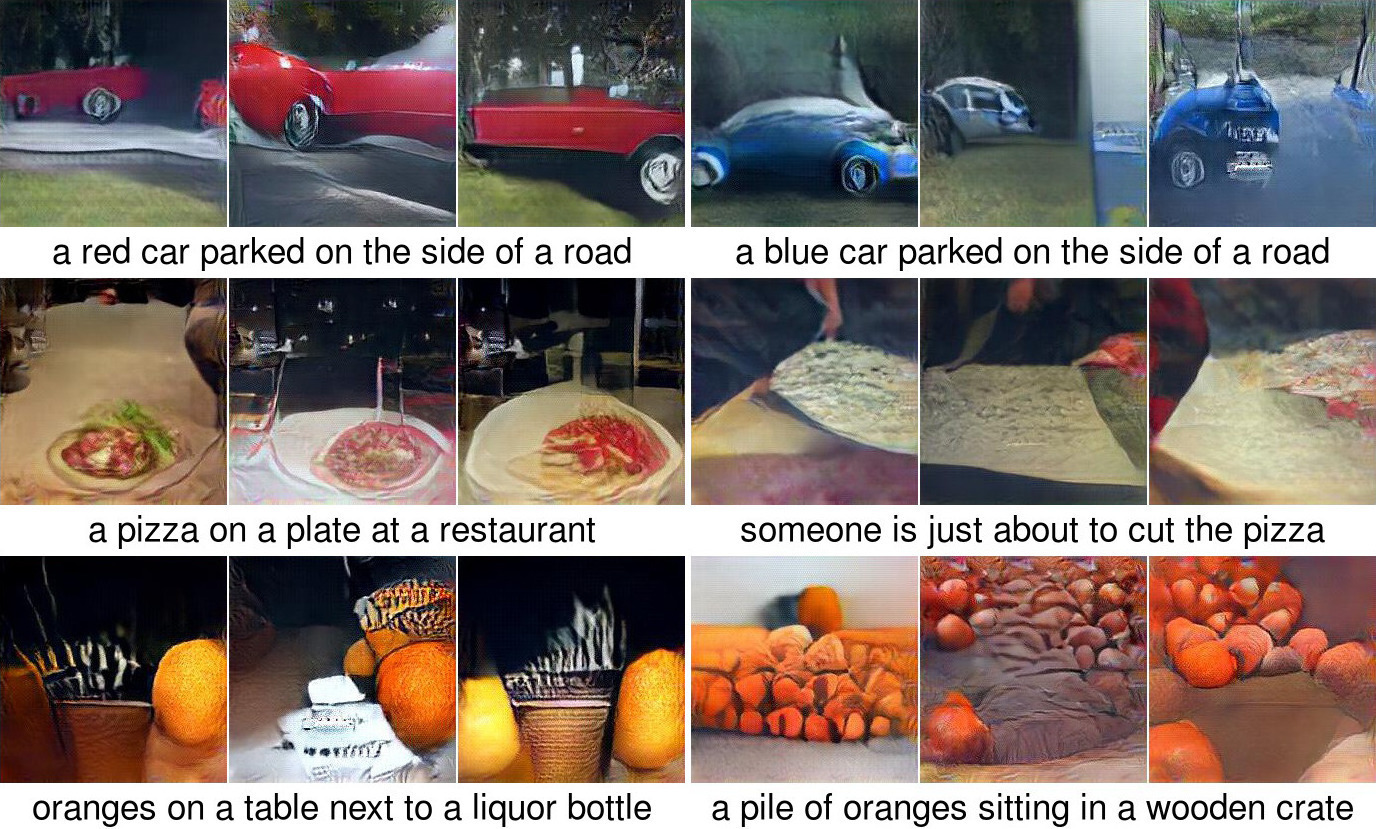}
	\caption{
		Images synthesized to match a text description. A PPGN containing the image captioning model from \cite{donahue-2014-arXiv-long-term-recurrent-convolutional} can generate reasonable images that differ based on user-provided captions (e.g. \emph{red} car vs. \emph{blue} car, oranges vs. \emph{a pile} of oranges). For each caption, we show 3 images synthesized starting from random codes (more in \figref{more_image_captioning}).
	}
	\figlabel{image_captioning}
	\vspace*{-0.3em}
\end{figure}

We found that PPGN can generate reasonable images in many cases (\figsref{image_captioning}{more_image_captioning}), although the image quality is lower than when conditioning on classes. In other cases, it also fails to generate high-quality images for certain types of images such as ``people'' or ``giraffe'', which are not categories in the generator's training set (\figref{more_image_captioning}). We also observe ``fooling'' images \cite{nguyen-2015-CVPR-deep-neural-networks}---those that look unrecognizable to humans, but produce high-scoring captions. More results are in \figref{more_image_captioning}. The challenges for this task could be: (1) the sampling is conditioned on many ($10-15$) words at the same time, and the gradients backpropagated from different words could conflict with each other; (2) the LRCN captioning model itself is easily fooled, thus
additional priors on the conversion from image features to natural language could
improve the result further; (3) the depth of the entire model (AlexNet and LRCN) impairs gradient propagation during sampling. In the future, it would be interesting to experiment with other state-of-the-art image captioning models \cite{frome-2013-NIPS-devise:-a-deep-visual-semantic,vinyals-2014-arXiv-show-and-tell:-a-neural}. Overall, we have demonstrated that PPGNs can be flexibly turned into a text-to-image model by combining the prior with an image captioning network, and this process \emph{does not} even require additional training.

\vspace*{.5em}
\noindent\textbf{Generating images conditioned on hidden neurons}
 
PPGNs can perform a more challenging form of activation maximization called Multifaceted Feature Visualization (MFV) \cite{nguyen-2016-arXiv-multifaceted-feature-visualization:}, which involves generating the \emph{set} of inputs that activate a given neuron.
Instead of conditioning on a class output neuron, here we condition on a \emph{hidden} neuron, revealing many facets that a neuron has learned to detect (\figref{faces}). 

\begin{figure}[h]
	\centering
	\includegraphics[width=1.0\columnwidth]{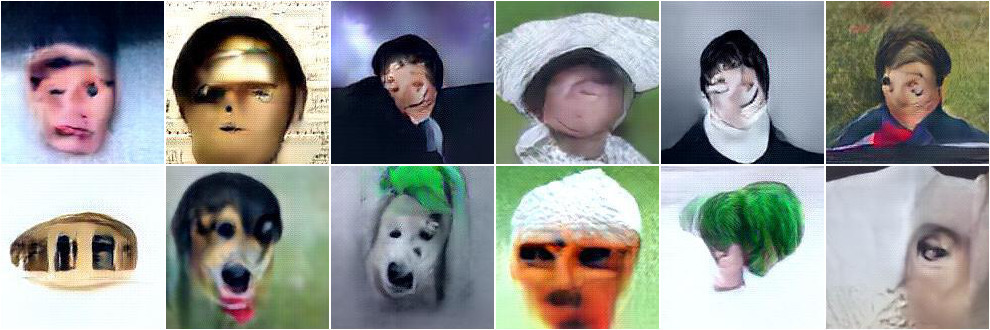}
	\caption{
		Images synthesized to activate a hidden neuron (number \unit{196}) previously identified as a ``face detector neuron'' \cite{yosinski-2015-ICML-DL-understanding-neural-networks} in the fifth convolutional layer of the AlexNet DNN trained on ImageNet. The PPGN uncovers a large diversity of types of inputs that activate this neuron, thus performing Multifaceted Feature Visualization \cite{nguyen-2016-arXiv-multifaceted-feature-visualization:}, which sheds light into what the neuron has learned to detect. The different facets include different types of human faces (top row), dog faces (bottom row), and objects that only barely resemble faces (e.g. the windows of a house, or something resembling green hair above a flesh-colored patch). More examples and details are shown in \figsref{conv5_text_detector}{conv5_face_detector}.
	}
	\figlabel{faces}
	\vspace*{-0.5em}
\end{figure}

\subsection{Inpainting}
\label{sec:inpainting}

Because PPGNs can be interpreted probabilistically, we can also
sample from them conditioned on part of an image (in addition to the class condition) to perform inpainting---filling in missing pixels given the observed context regions \cite{pathak2016context,barnes2009patchmatch,yeh2016semantic,van-den-oord-2016-arXiv-pixel-recurrent-neural}.
The model must understand the entire image to be able to reasonably fill in a large masked out region that is positioned randomly. Overall, we found that PPGNs are able to perform inpainting suggesting that the models do ``understand'' the semantics of concepts such as junco or bell pepper (\figref{inpainting}) rather than merely memorizing the images. More details and results are in Sec.~\ref{sec:SI_inpainting}.

\begin{figure}[t]
	\centering
	\includegraphics[width=1.0\columnwidth]{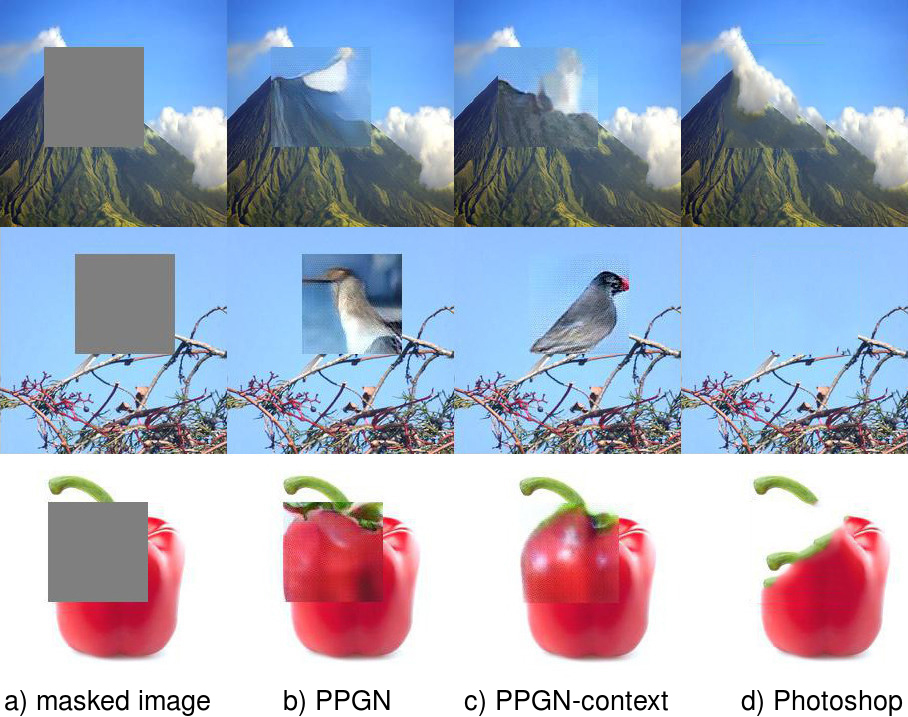}
	\caption{		
		We perform class-conditional image sampling to fill in missing pixels (see \secref{inpainting}). In addition to conditioning on a specific class
		 (PPGN), PPGN-context also constrains the code $h$ to produce an image that matches the context region.
		PPGN-context (c) matches the pixels surrounding the masked region better than PPGN (b), and semantically fills in better than the Context-Aware Fill feature in Photoshop (d) in many cases. The result shows that the class-conditional PPGN does understand the semantics of images. More PPGN-context results are in \figref{more_inpainting}.
	}
	\figlabel{inpainting}
    \vspace*{-1.0em}
\end{figure}

\section{Conclusion}
The most useful property of PPGN is the capability of ``plug and play''---allowing one to drop in a replaceable condition network and generate images according to a condition specified at test time.
Beyond the applications we demonstrated here, one could use PPGNs to synthesize images for videos or create arts with one or even multiple condition networks at the same time \cite{yahoo2016nsfw}. 
Note that DGN-AM \cite{nguyen-2016-synthesizing-the-preferred-inputs}---the predecessor of PPGNs---has previously enabled both scientists and amateurs without substantial resources to take a pre-trained condition network and generate art \cite{yahoo2016nsfw} and scientific visualizations \cite{zhou2016places}. An explanation for why this is possible is that the \layer{fc6} features that the generator was trained to invert are relatively general and cover the set of natural images. Thus, there is great value in producing flexible, powerful generators that can be combined with pretrained condition networks in a plug and play fashion. 

\subsubsection*{Acknowledgments}
We thank Theo Karaletsos and Noah Goodman for helpful discussions, and
Jeff Donahue for providing a trained image captioning model \cite{donahue-2014-arXiv-long-term-recurrent-convolutional} for our experiments. We also thank Joost Huizinga, Christopher Stanton, Rosanne Liu, Tyler Jaszkowiak, Richard Yang, and Jon Berliner for invaluable suggestions on preliminary drafts.


{\small
\bibliographystyle{ieee}
\bibliography{references}
}

\clearpage

\renewcommand{\thesection}{S\arabic{section}}
\renewcommand{\thesubsection}{\thesection.\arabic{subsection}}

\newcommand{\beginsupplementary}{%
	\renewcommand{\thetable}{S\arabic{table}}%
	\renewcommand{\thefigure}{S\arabic{figure}}%
}
\newcommand{\suptitle}{Supplementary materials for:\\\papertitle}

\newcommand{\toptitlebar}{
	\hrule height 4pt
	\vskip 0.25in
	\vskip -\parskip%
}
\newcommand{\bottomtitlebar}{
	\vskip 0.29in
	\vskip -\parskip
	\hrule height 1pt
	\vskip 0.09in%
}

\beginsupplementary

\newcommand{\maketitlesupp}{
\newpage
\twocolumn[
	\begin{@twocolumnfalse}
	\null
	\vskip .375in
	\begin{center}
		{\Large \bf \suptitle \par}
		\vspace*{24pt}
		{
			\large
			\lineskip .5em
			\par
		}
		\vskip .5em
		\vspace*{12pt}
	\end{center}
\end{@twocolumnfalse}
]
}

\maketitlesupp

\section{Markov chain Monte Carlo methods and derivation of MALA-approx}
\seclabel{samplers}

Assume a distribution $p(x)$ that we wish to produce samples from. For certain distributions with amenable structure it may be possible to write down directly an independent and identically distributed (IID) sampler, but in general this can be difficult. In such cases where IID samplers are not readily available, we may instead resort to Markov Chain Monte Carlo (MCMC) methods for sampling. Complete discussions of this topic fill books \cite{koller2009probabilistic,Goodfellow-et-al-2016-Book}. Here we briefly review the background that led to the sampler we propose.

In cases where evaluation of $p(x)$ is possible, we can write down the Metropolis-Hastings (hereafter: \emph{MH}) sampler for $p(x)$~\cite{metropolis-1953-JChemPhys-equation-of-state-calculations,hastings-1970-Biometrika-monte-carlo-sampling}. It requires a choice of proposal distribution $q(x' | x)$; for simplicity we consider (and later use) a simple Gaussian proposal distribution. Starting with an $x_0$ from some initial distribution, the sampler takes steps according to a transition operator defined by the below routine, with $N(0, \sigma^2)$ shorthand for a sample from that Gaussian proposal distribution:

\begin{enumerate}
	\item $x_{t+1} = x_t + N(0, \sigma^2)$
	\item $\alpha = p(x_{t+1}) / p(x_t)$
	\item if $\alpha < 1$, reject sample $x_{t+1}$ with probability $1-\alpha$ by setting $x_{t+1} = x_t$, else keep $x_{t+1}$
\end{enumerate}

In theory, sufficiently many steps of this simple sampling rule produce samples for any computable $p(x)$, but in practice it has two problems: it mixes slowly, because steps are small and uncorrelated in time, and it requires us to be able to compute $p(x)$ to calculate $\alpha$, which is often not possible. A Metropolis-adjusted Langevin algorithm (hereafter: \emph{MALA}) \cite{roberts1996exponential,roberts-1998-JRSS-optimal-scaling-of-discrete} addresses the first problem. This sampler follows a slightly modified procedure:

\begin{enumerate}
	\item $x_{t+1} = x_t + \sigma^2/2\nabla\log p(x_t) + N(0, \sigma^2)$
	\item $\alpha = f(x_t, x_{t+1}, p(x_{t+1}), p(x_t))$
	\item if $\alpha < 1$, reject sample $x_{t+1}$ with probability $1-\alpha$ by setting $x_{t+1} = x_t$, else keep $x_{t+1}$
\end{enumerate}

where $f(\cdot)$ is the slightly more complex calculation of $\alpha$, with the notable property that as the step size goes to 0, $f(\cdot) \to 1$.
This sampler preferentially steps in the direction of higher probability,
which allows it to spend less time rejecting low probability proposals,
but it still requires computation of $p(x)$ to calculate $\alpha$.

\setlength\cellspacetoplimit{4pt}
\setlength\cellspacebottomlimit{2pt}
\begin{table*}
	\centering
	\begin{tabular}{| Sl | c  c  l |}
		\hline
		&          & uses accept/ &  \\
		&          & reject step and  &  \\
		& mixes    & requires $p(x)$    & update rule (not including accept/reject step) \\
		\hline
		MH            & slowly   & yes       & $x_{t+1} = x_t + N(0, \sigma^2)$ \\
		MALA          & ok       & yes       & $x_{t+1} = x_t + 1/2\sigma\nabla\log p(x_t) + N(0, \sigma^2)$ \\
		MALA-approx  & ok       & no        & $x_{t+1} = x_t + \epsilon_{12}\nabla\log p(x_t) + N(0, \epsilon_3^2)$ \\
		\hline    
	\end{tabular}
	\caption{
		Samplers properties assuming Gaussian proposal distributions. Samples are drawn via MALA-approx in this paper.
	}
	\tablabel{samplers}
	\vspace{-0em}
\end{table*}

The stochastic gradient Langevin dynamics (SGLD) method \cite{welling2011bayesian,teh2014consistency} was proposed to sidestep this troublesome requirement by generating probability proposals that are based on a small subset of the data only: by using stochastic gradient descent plus noise, by skipping the accept-reject step, and by using decreasing step sizes. Inspired by SGLD, we define an approximate sampler by assuming small step size and doing away with the reject step (by accepting every sample). The idea is that the stochasticity of SGD itself introduces an implicit noise: although the resulting update does not produce asymptotically unbiased samples, it does if we also anneal the step size (or, equivalently, gradually increase the minibatch size).



While an accept ratio of 1 is only approached in the limit as the step size goes to zero, in practice we
empirically observe that this approximation produces reasonable samples even for moderate step sizes.
This approximation leads to a sampler defined by the simple update rule:

\beq
x_{t+1} = x_t + \sigma^2/2\nabla\log p(x_t) + N(0, \sigma^2)
\eeq

As explained below, we propose to decouple the two step sizes for each of the above two terms after $x_t$,
with two independent scaling factors to allow independently tuning each ($\epsilon_{12}$ and $\epsilon_3$ in \eqnref{mala-approx_2}).
This variant makes sense when we consider that the stochasticity of SGD itself introduces more noise, breaking
the direct link between the amount of noise injected and the step size under Langevin dynamics.

We note that $p(x) \propto \exp(-{\textrm{Energy}}(x))$,
$\nabla\log p(x_t)$ is just the gradient of the energy (because the partition function does not depend on $x$)
and that the scaling factor ($\sigma^2/2$ in the above equation) can be partially absorbed when changing
the {\em temperature} associated with energy, since temperature is just a multiplicative scaling factor in the energy.
Changing that link between the two terms is thus equivalent to changing temperature 
because the incorrect scale factor can
be absorbed in the energy as a change in the temperature.
Since we do not control directly the amount of noise (some of which is now produced
by the stochasticity of SGD itself), it is better to ``manually'' control the trade-off by introducing
an extra hyperparameter. Doing so also may help to compensate for the fact that the SGD noise is not
perfectly normal, which introduces a bias in the Markov chain. By manually controlling both the step
size and the normal noise, we can thus find a good trade-off between variance (or temperature level,
which would blur the
distribution) and bias (which makes us sample from a slightly different distribution).
In our experience, such decoupling has helped find better tradeoffs between sample diversity and quality, perhaps compensating for idiosyncrasies of sampling without a reject step. We call this sampler \emph{MALA-approx}:

\beq
x_{t+1} = x_t + \epsilon_{12}\nabla\log p(x_t) + N(0, \epsilon_3^2)
\eqnlabel{mala-approx_2}
\eeq

\noindent
\tabref{samplers} summarizes the samplers and their properties.

\section{Probabilistic interpretation of previous models (continued)}
\seclabel{previous_models_si}

In this paper, we consider four main representative approaches in light of the framework:

\begin{enumerate}
	\item Activation maximization with no priors \cite{nguyen-2015-CVPR-deep-neural-networks,szegedy2013intriguing-properties-of-neural,erhan2009visualizing-higher-layer-features}
	\item Activation maximization with a Gaussian prior \cite{simonyan2013deep-inside-convolutional,yosinski-2015-ICML-DL-understanding-neural-networks}
	\item Activation maximization with hand-designed priors \cite{simonyan2013deep-inside-convolutional,yosinski-2015-ICML-DL-understanding-neural-networks,nguyen-2016-arXiv-multifaceted-feature-visualization:,wei-2015-understanding-intra-class-knowledge,nguyen-2015-GECCO-innovation-engines:-automated,nguyen-2015-CVPR-deep-neural-networks,mahendran2016visualizing}
	\item Sampling in the latent space of a generator network	\cite{arulkumaran2016improving,yeh2016semantic,zhu2016generative,brock2016neural,nguyen-2016-synthesizing-the-preferred-inputs,han2016alternating} 
\end{enumerate}

Here we discuss the first three and refer readers to the main text (\secref{previous_models}) for the fourth approach.

\noindent\textbf{Activation maximization with no priors.}
From \eqnref{update_rule}, if we set $(\epsilon_1, \epsilon_2, \epsilon_3) = (0, 1, 0)$
, we obtain a sampler that follows the class gradient directly without contributions from a $p(x)$ term or the addition of noise. In a high-dimensional space, this results in adversarial or fooling images \cite{szegedy2013intriguing-properties-of-neural,nguyen-2015-CVPR-deep-neural-networks}. We can also interpret the sampling procedure in \cite{szegedy2013intriguing-properties-of-neural,nguyen-2015-CVPR-deep-neural-networks} as a sampler with non-zero $\epsilon_1$ but with a $p(x)$ such that $\frac{\partial\log p(x)}{\partial{x}} = 0$; in other words, a uniform $p(x)$ where all images are equally likely.

\setlength\cellspacetoplimit{4pt}
\setlength\cellspacebottomlimit{2pt}
\definecolor{hilightclr}{rgb}{0.0,0.6,0.4}
{
	\begin{table*}
		\center
		\begin{tabular}{>{\raggedright}m{.5\textwidth} | >{\arraybackslash}m{.4\textwidth} }  
			\hline
			\emph{a. Derivative of logit.} Has worked well in practice \cite{nguyen-2016-synthesizing-the-preferred-inputs,erhan2009visualizing-higher-layer-features} but not quite the right term to maximize under the sampler framework set out in this paper. &
			\begin{align*}
			{\color{hilightclr} \frac{\partial l_i}{\partial x}}
			\end{align*} \\
			\hline
			\emph{b. Derivative of softmax.} Previously avoided due to poor performance \cite{simonyan2013deep-inside-convolutional,yosinski-2015-ICML-DL-understanding-neural-networks}, but poor performance may have been due to ill-conditioned optimization rather than the inclusion of logits from other classes. In particular, the term goes to 0 as $s_i$ goes zero. 
			&
			\begin{align*}
			\frac{\partial s_i}{\partial x} = s_i\left({\color{hilightclr} \frac{\partial l_i}{\partial x}} - \sum_j s_j \frac{\partial l_j}{\partial x}\right)
			\end{align*} \\
			\hline
			\emph{c. Derivative of log of softmax.} Correct term under the sampler framework set out in this paper. Well-behaved under optimization, perhaps due to the $\partial l_i/\partial x$ term untouched by the $s_i$ multiplier. &
			{\begin{align*}
				\frac{\partial \log s_i}{\partial x} & = \frac{\partial \log p(y = y_i | x_t)}{\partial x}  \\
				& = {\color{hilightclr} \frac{\partial l_i}{\partial x}} - \frac{\partial}{\partial x} \log \sum_j \exp(l_j)
				\end{align*}} \\
			\hline
		\end{tabular}
		\caption{A comparison of derivatives for use in activation maximization experiments. The first has most commonly been used, the second has worked in the past but with some difficulty, but the third is correct under the sampler framework set out in this paper. We perform experiments in this paper with the third variant. 
		}
		\tablabel{logit_vs_softmax}
		\vspace{-0em}
	\end{table*}
}

\noindent\textbf{Activation maximization with a Gaussian prior.} To combat the fooling problem \cite{nguyen-2015-CVPR-deep-neural-networks}, several works have used $L_2$ decay, which can be thought of as a simple Gaussian
prior over images \cite{simonyan2013deep-inside-convolutional,yosinski-2015-ICML-DL-understanding-neural-networks,wei-2015-understanding-intra-class-knowledge}.
From \eqnref{update_rule}, if we define a Gaussian $p(x)$ centered at the origin (assume the mean image has been subtracted) and set $(\epsilon_1, \epsilon_2, \epsilon_3) = (\lambda, 1, 0)$, pulling Gaussian constants into $\lambda$, we obtain the following noiseless update rule:

\beq
x_{t+1} = (1-\lambda) x_t + \frac{\partial \log p(y = y_c | x_t)}{\partial x_t}
\eqnlabel{update_rule_gaussian}
\eeq

The first term decays the current image slightly toward the origin, as appropriate under a Gaussian image prior, and the second term pulls the image toward higher probability regions for the chosen class. Here, the second term is computed as the derivative of the log of a softmax unit in the output layer of the classification network, which is trained to model $p(y | x)$.
If we let $l_i$ be the logit outputs of a classification network, where $i$ indexes over the classes, then the softmax outputs are given by $s_i = \exp(l_i) / \sum_j \exp(l_j)$, and the value $p(y = y_c | x_t)$ is modeled by the softmax unit $s_c$.

Note that the second term is similar, but not identical, to the gradient of logit term used by \cite{simonyan2013deep-inside-convolutional,yosinski-2015-ICML-DL-understanding-neural-networks,mahendran2016visualizing}. There are three variants of computing this class gradient term: 1) derivative of logit; 2) derivative of softmax; and 3) derivative of log of softmax. Previously
mentioned papers empirically reported that derivative of the logit unit $l_i$ produces better visualizations than the derivative of the softmax unit $s_i$ (\tabref{logit_vs_softmax}a vs. b), but this observation had not been fully justified \cite{simonyan2013deep-inside-convolutional}. In light of our probablistic interpretation (discussed in \secref{mcmc_interpretation}), we consider activation maximization as performing noisy gradient descent to minimize the energy function $E(x,y)$:
\begin{align}
E(x,y) &= -log(p(x,y))\nonumber\\
&= -log(p(x)p(y|x))\nonumber\\
&= -(log(p(x)) + log(p(y|x)))
\end{align}

To sample from the joint model $p(x,y)$, we follow the energy gradient:

\begin{align}
\frac{\partial E(x,y)}{\partial x} &= -\bigg( \frac{\partial log(p(x))}{\partial x} + \frac{\partial log(p(y|x))}{\partial x} \bigg) 
\end{align}

\noindent which derives the class gradient term that matches that in our framework (\eqnref{update_rule_gaussian}, second term). In addition, recall that the classification network is trained to model $p(y|x)$ via softmax, thus the class gradient variant (the derivative of log of softmax) is the most theoretically justifiable in light of our interpretation. We summarize all three variants in \tabref{logit_vs_softmax}.
In overall, we found the proposed class gradient term a) theoretically justifiable under the probabilistic interpretation, and b) working well empirically, and thus suggest it for future activation maximization studies.

\noindent\textbf{Activation maximization with hand-designed priors.} In an effort to outdo the simple Gaussian prior, many works have proposed more creative, hand-designed image priors such as Gaussian blur \cite{yosinski-2015-ICML-DL-understanding-neural-networks},
total variation \cite{mahendran2016visualizing}, jitter \cite{mordvintsev2015inceptionism}, and data-driven patch priors \cite{wei2015understanding}. These priors effectively serve as a simple $p(x)$ component. Those that cannot be explicitly expressed in the mathematical $p(x)$ form (e.g. jitter \cite{mordvintsev2015inceptionism} and center-biased regularization \cite{nguyen-2016-arXiv-multifaceted-feature-visualization:}) 
can be written as a general regularization function $r(.)$ as in \cite{yosinski-2015-ICML-DL-understanding-neural-networks}, in which case the noiseless update becomes:

\beq
x_{t+1} = r(x_t) + \frac{\partial \log p(y = y_c | x_t)}{\partial x_t}
\eeq

Note that all methods considered in this section are noiseless and therefore produce samples showing diversity only by starting the optimization process at different initial conditions. The effect is that samples tend to converge to a single mode or a small number of modes \cite{erhan2009visualizing-higher-layer-features,nguyen-2016-arXiv-multifaceted-feature-visualization:}.

\section{Comparing feature matching losses}
\seclabel{compare_losses}
The addition of \emph{feature matching} losses (i.e. the differences between a real image and a generated image not in pixel space, but in a feature space, such as a high-level code in a deep neural network) to the training cost has been shown to substantially improve the quality of samples produced by generator networks, e.g. by producing sharper and more realistic images \cite{dosovitskiy-2016-NIPS-generating-images-with,larsen-2015-arXiv-autoencoding-beyond-pixels,johnson2016perceptual}. 

Dosovitskiy \& Brox \cite{dosovitskiy-2016-NIPS-generating-images-with} used the feature matching loss measured in the \layer{pool5} layer code space of AlexNet 
deep neural network (DNN) \cite{krizhevsky2012imagenet-classification-with-deep} trained to classify 1000-class ImageNet images \cite{deng2009imagenet:-a-large-scale-hierarchical}.
Here, we empirically compare several feature matching losses computed in different layers of the AlexNet DNN. Specifically, we follow the training procedure in 
Dosovitskiy \& Brox
\cite{dosovitskiy-2016-NIPS-generating-images-with}, and train 3 generator networks, each with a different feature matching loss computed in different layers from the pretrained AlexNet DNN: a) \layer{pool5}, b) \layer{fc6} and c)
both \layer{pool5} and \layer{fc6} losses. 
We empirically found that matching the \layer{pool5} features leads to the best image quality (\figref{compare_perceptual}), and chose the generator with this loss for the main experiments in the paper.

\begin{figure*}
	\centering
	\begin{subfigure}{1.0\linewidth}
		\centering
		\includegraphics[width=1.0\linewidth]{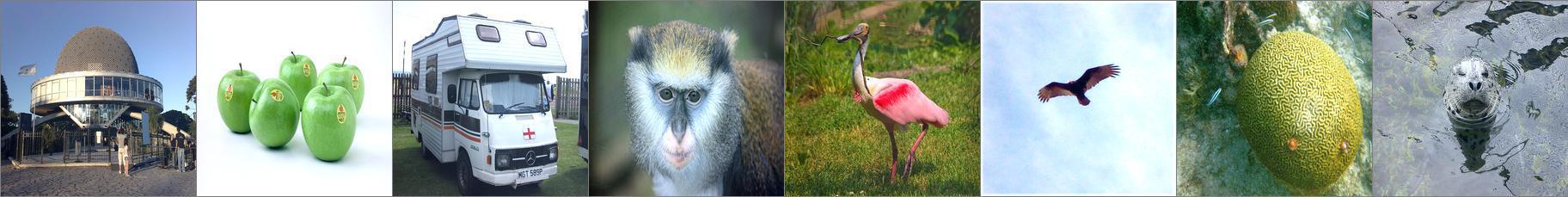}
		\caption{Real images}
		\vspace{0.2cm}
	\end{subfigure}
	\hspace{1mm}
	\begin{subfigure}{1.0\linewidth}
		\centering
		\includegraphics[width=1.0\linewidth]{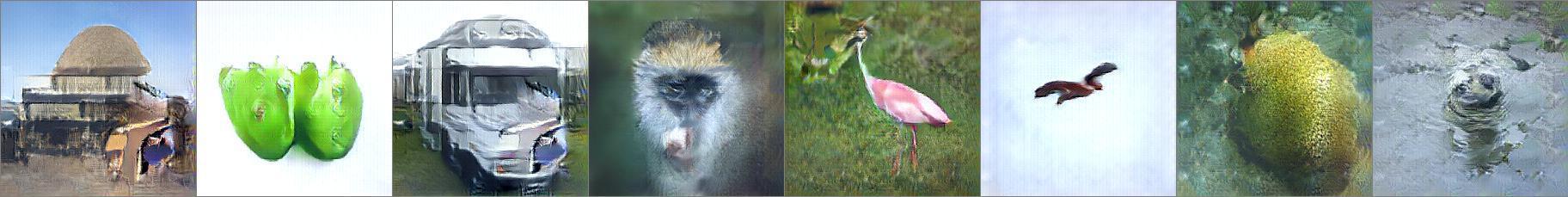}
		\caption{Joint PPGN-$h$ ($L_{img} + L_{h_1} + L_h + L_{GAN}$)}
		\vspace{0.2cm}
	\end{subfigure}	
	\hspace{1mm}
	\begin{subfigure}{1.0\linewidth}
		\centering
		\includegraphics[width=1.0\linewidth]{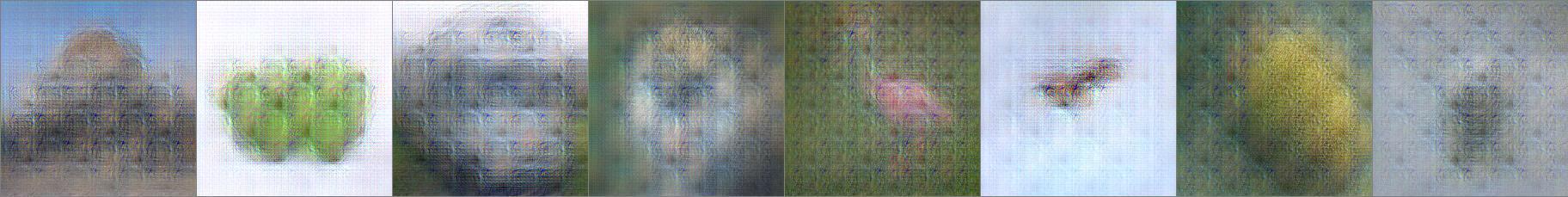}
		\caption{$L_{GAN}$ removed ($L_{img} + L_{h_1} + L_h$)}
		\figlabel{no_gan}
		\vspace{0.2cm}
	\end{subfigure}	
	\hspace{1mm}
	\begin{subfigure}{1.0\linewidth}
		\centering
		\includegraphics[width=1.0\linewidth]{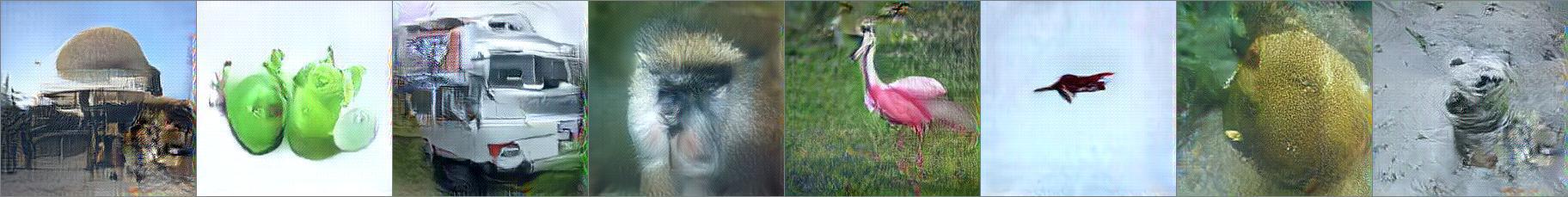}
		\caption{$L_{h_1}$ removed: $L_{img} + L_h + L_{GAN}$}
		\vspace{0.2cm}
	\end{subfigure}	
	\hspace{1mm}
	\begin{subfigure}{1.0\linewidth}
		\centering
		\includegraphics[width=1.0\linewidth]{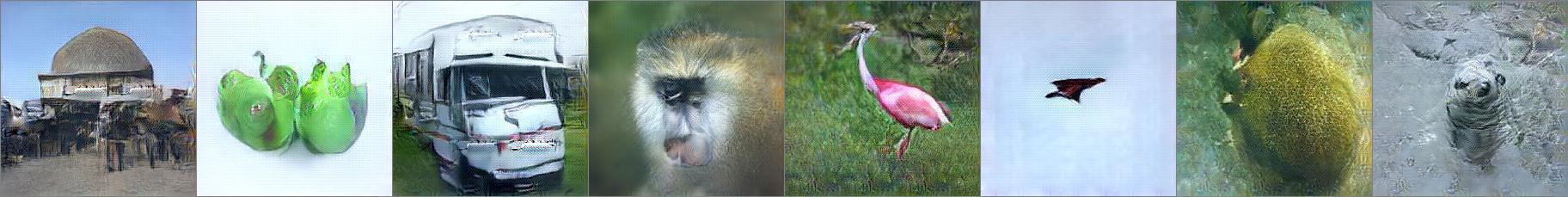}
		\caption{$L_h$ removed: $L_{img} + L_{h_1} + L_{GAN}$}
		\vspace{0.2cm}
	\end{subfigure}	
	\caption{
		A comparison of images produced by different generators $G$, each trained with a different loss combination (below each image). $L_{img}$, $L_{h_1}$, and $L_h$ are $L_2$ reconstruction losses respectively in the pixel ($x$), \layer{pool5} feature ($h_1$) and \layer{fc6} feature ($h$) space. $G$ is trained to map $h \to x$, i.e. reconstructing images from \layer{fc6} features.
		In the Joint PPGN-$h$ treatment (\secref{default_ppgn}), $G$ is trained with a combination of 4 losses (panel b). Here, we perform an ablation study on this loss combination to understand the effect of each loss, and find a combination that produces the best image quality.
		We found that removing the GAN loss yields blurry results (panel c). The Noiseless Joint PPGN-$h$ variant (\secref{noiseless_ppgn}) is trained with the loss combination that produces the best image quality (panel e).		
		Compared to \layer{pool5}, \layer{fc6} feature matching loss often produce the worse image quality because it is effectively encouraging generated images to match the high-level abstract statistics of real images instead of low-level statistics \cite{gretton2006kernel}. Our result is in consistent with Dosovitskiy \& Brox \cite{dosovitskiy-2016-NIPS-generating-images-with}.
	}
	\vspace*{8em} 
	\figlabel{compare_perceptual}
\end{figure*}

\section{Training details}
\label{sec:train_details}

\subsection{Common training framework}
\label{sec:common_training}
We use the Caffe framework \cite{jia2013caffe:-an-open-source} to train the networks. All networks are trained with the Adam
optimizer \cite{kingma2014adam} with momentum $\beta_1=0.9$, $\beta_2 = 0.999$, and $\gamma = 0.5$, and an initial learning rate of $0.0002$ following \cite{dosovitskiy-2016-NIPS-generating-images-with}. The batch size is $64$. To stabilize the GAN training, we follow heuristic rules based on the ratio of the discriminator loss over generator loss $r = loss_D / loss_G$ and pause the training of the generator or discriminator if one of them is winning too much.
In most cases, the heuristics are a) pause training D if $r < 0.1$; b) pause training G if $r > 10$. We did not find BatchNorm \cite{ioffe-2015-arXiv-batch-normalization:-accelerating} helpful in further stabilizing the training as found in Radford et al. \cite{radford-2015-arXiv-unsupervised-representation-learning}. We have not experimented with all of the techniques discussed in Salimans et al. \cite{salimans-2016-arXiv-improved-techniques-for-training}, some of which could further improve the results.

\subsection{Training PPGN-$x$}
\label{sec:train_x_DAE}

We train a DAE for images and incorporate it to the sampling procedure as a $p(x)$ prior to avoid fooling examples \cite{nguyen-2016-synthesizing-the-preferred-inputs}. The DAE is a 4-layer convolutional network that encodes an image to the layer \layer{conv1} 
of AlexNet \cite{krizhevsky2012imagenet-classification-with-deep} and decodes it back to images with 3 upconvolutional layers.
We add an amount of Gaussian noise $\sim N(0, \sigma^2)$ with $\sigma=25.6$ to images during training. The network is trained via the common training framework described in \secref{common_training} for $25,000$ mini-batch iterations. We use $L_2$ regularization of $0.0004$.

\subsection{Training PPGN-$h$}
\label{sec:train_h_DAE}
For the PPGN-$h$ variant, we train two separate networks: a generator $G$ (that maps codes $h$ to images $x$) and a prior $p(h)$.
$G$ is trained via the same procedure described in \secref{train_PPGN}.
We model $p(h)$ via a multi-layer perceptron DAE with 7 hidden layers of size: $4096 - 2048 - 1024 - 500 - 1024 - 2048 - 4096$. We experimented with larger networks but found this to work the best. We sweep across different amounts of Gaussian noise $N(0, \sigma^2)$, and empirically chose $\sigma=1$ (i.e. $\sim$10\% of the mean \layer{fc6} feature activation).
The network is trained via the common training framework described in \secref{common_training} for $100,000$ mini-batch iterations. We use $L_2$ regularization of $0.001$.

\subsection{Training Noiseless Joint PPGN-$h$}
\label{sec:train_PPGN}

Here we describe the training details of the generator network $G$ used in the main experiments in Sections \ref{sec:ppgn_h}, \ref{sec:noiseless_ppgn}, \ref{sec:default_ppgn}.
The training procedure follows closely the framework by Dosovitskiy \& Brox \cite{dosovitskiy-2016-NIPS-generating-images-with}. 

The purpose is to train a generator network $G$ to reconstruct images from an abstract, high-level feature code space of an encoder network $E$---here, the first fully connected layer (\layer{fc6}) of an AlexNet DNN \cite{krizhevsky2012imagenet-classification-with-deep} pre-trained to perform image classification on the ImageNet dataset \cite{deng2009imagenet:-a-large-scale-hierarchical} (\figref{training_diagram}a)
We train $G$ as a decoder for the encoder $E$, which is kept frozen. In other words, $E+G$ form an image autoencoder (\figref{training_diagram}b).

\textbf{Training losses.}
G is trained with 3 different losses as in Dosovitskiy \& Brox \cite{dosovitskiy-2016-NIPS-generating-images-with}, namely, an adversarial loss $L_{GAN}$, an image reconstruction loss $L_{img}$, and a feature matching loss $L_{h_1}$ measured in the spatial layer \layer{pool5} (\figref{training_diagram}b)
:

\beq
L_G = L_{img} + L_{h_1} + L_{GAN}
\eeq

$L_{img}$ and $L_{h_1}$ are $L_2$ reconstruction losses in their respective space of images $x$ and $h_1$ (\layer{pool5}) codes : 

\begin{align}
L_{img} &= || \hat{x} - x||^2 \\
L_{h_1} &= || \hat{h_1} - h_1||^2
\end{align}

The adversarial loss for $G$ (which intuitively maximizes the chance $D$ makes mistakes) follows the original GAN paper \cite{goodfellow2014generative-adversarial-networks}:

\beq
L_{GAN} = -\sum_{i}{log(D_\rho (G_\theta(h_i)))}
\eeq

where $x_i$ is a training image, and $h_i = E(x_i)$ is a code. As in Goodfellow et al. \cite{goodfellow2014generative-adversarial-networks}, $D$ tries to tell apart real and fake images, and is trained with the adversarial loss as follows:

\beq
L_{D} = -\sum_{i}{log(D_\rho (x_i)) + log(1-D_\rho(G_\theta(h_i)))}
\eeq

\textbf{Architecture.} $G$, an upconvolutional (also ``deconvolutional'') network \cite{dosovitskiy-2015-CVPR-learning-to-generate-chairs} with 9 upconvolutional and 3 fully connected layers. $D$ is a regular convolutional network for image classification with a similar architecture to AlexNet \cite{krizhevsky2012imagenet-classification-with-deep} with 5 convolutional layers followed by 3 fully connected layers, and 2 outputs (for ``real'' and ``fake'' classes).

The networks are trained via the common training framework described in \secref{common_training} for $10^6$ mini-batch iterations. We use $L_2$ regularization of $0.0004$.

\textbf{Specifics of DGN-AM reproduction.}
Note that while the original set of parameters in Nguyen et al. \cite{nguyen-2016-synthesizing-the-preferred-inputs} (including a small number of iterations, an $L_2$ decay on code $h$, and a step size decay) produces high-quality images, it does not allow for a long sampling chain, traveling from one mode to another. For comparisons with other models within our framework, we sample from DGN-AM with $(\epsilon_1, \epsilon_2, \epsilon_3) = (0, 1, 10^{-17})$, which is slightly different from $(\lambda, 1, 0)$ in \eqnref{dgnam_update_rule}, but produces qualitatively the same result.

\subsection{Training Joint PPGN-$h$}
\label{sec:train_PPGN_noise}

Via the same existing network structures from DGN-AM \cite{nguyen-2016-synthesizing-the-preferred-inputs}, we train the generator $G$ differently by treating the entire model as being composed of 3 interleaved DAEs: one for $h$, $h_1$, and $x$ respectively (see \figref{training_diagram}c).
Specifically, we add Gaussian noise to these variables during training, and by incorporating three corresponding $L_2$ reconstruction losses (see \figref{training_diagram}c). Adding noise to an AE can be considered as a form of regularization that encourages an autoencoder to extract more useful features \cite{vincent-2008-ICML-extracting-and-composing-robust}. Thus, here, we hypothesize that adding a small amount of noise to $h_1$ and $x$ might slightly improve the result. In addition, the benefits of adding noise to $h$ and training the pair $G$ and $E$ as a DAE for $h$ are two fold: 1) it allows us to formally estimate the quantity $\partial logp(h)/\partial h$ (see \eqnref{rx_x})
following a previous mathematical justification from Alain \& Bengio \cite{alain-2014-what-regularized-auto-encoders}; 2) it allows us to sample with a larger noise level, which might improve the mixing speed. 

We add noise to $h$ during training, and train $G$ with a $L_2$ reconstruction loss for $h$:

\beq
L_{h} = || \hat{h} - h||^2
\eeq





Thus, generator network $G$ is trained with 4 losses in total:

\beq
L_G = L_{img} + L_{h} + L_{h_1} + L_{GAN}
\eeq

Three losses $L_{img}$, $L_{h_1}$, and $L_{GAN}$ remain the same as in the training of Noiseless Joint PPGN-$h$ (\secref{train_PPGN}). Network architectures and other common training details remain the same as described in \secref{train_PPGN}.

The amount of Gaussian noise $N(0,\sigma^2)$ added to 
the 3 different variables $x$, $h_1$, and $h$ are respectively $\sigma = \{1, 4, 1\}$ which are $\sim$1\% of the mean pixel values and $\sim$10\% of the mean activations respectively in \layer{pool5} and \layer{fc6} space computed from the training set. We experimented with larger noise levels, but were not able to train the model successfully as large amounts of noise resulted in training instability. We also tried training without noise for $x$, i.e. treating the model as being composed of 2 DAEs instead of 3, but did not obtain qualitatively better results.

Note that while we did not experiment in this paper, jointly training both the generator $G$ and the encoder $E$ via their respective maximum likelihood training algorithms is possible. Also, Xie et al. \cite{xie2016cooperative} has proposed a training regime that alternatively updates these two networks. That cooperative training scheme indeeds yields a generator that synthesizes impressive results for multiple image datasets \cite{xie2016cooperative}.

\begin{figure*}
	\centering
	\includegraphics[width=2.0\columnwidth]{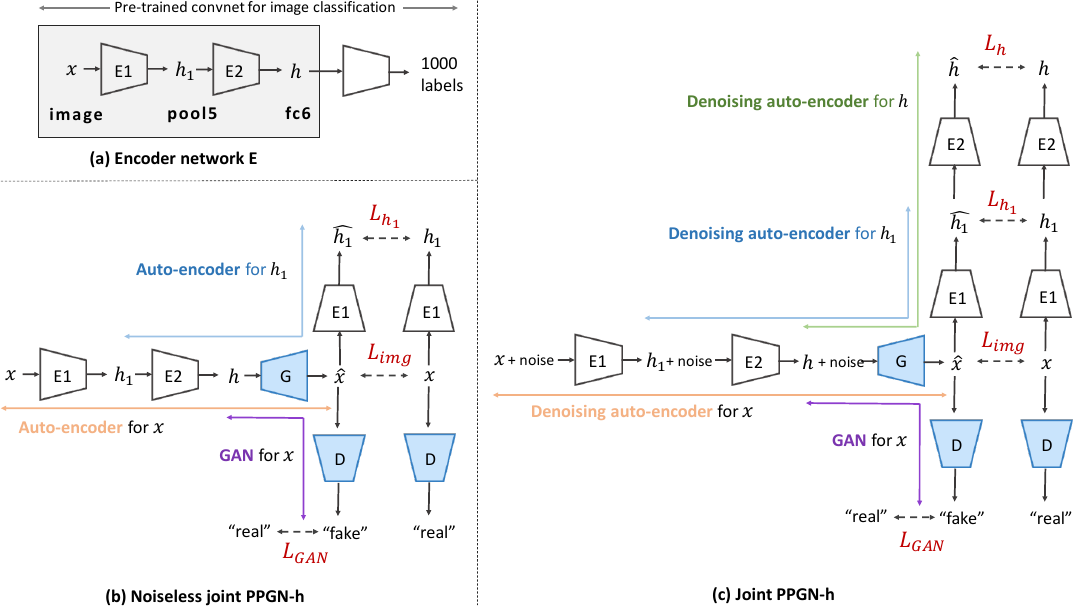}
	\caption{In this paper, we propose a class of models called PPGNs that are composed of 1) a generator network $G$ that is trained to draw a wide range of image types, and 2) a replaceable ``condition'' network $C$ that tells $G$ what to draw (\figref{concept}). Panel (b) and (c) show the components involved in the training of the generator network $G$ for two main PPGN variants experimented in this paper. Only shaded components ($G$ and $D$) are being trained while others are kept frozen.
		\textbf{b)} For the Noiseless Joint PPGN-$h$ variant (\secref{noiseless_ppgn}),
		we train a generator $G$ to reconstruct images $x$ from compressed features $h$ produced by a pre-trained encoder network $E$. Specifically, $h$ and $h_1$ are, respectively, features extracted at layer \layer{fc6} and \layer{pool5} of AlexNet \cite{krizhevsky2012imagenet-classification-with-deep} trained to classify ImageNet images (a). 
		$G$ is trained with 3 losses: an image reconstruction loss $L_{img}$, a feature matching loss \cite{dosovitskiy-2016-NIPS-generating-images-with} $L_{h_1}$ and an adversarial loss \cite{goodfellow2014generative-adversarial-networks} $L_{GAN}$. As in Goodfellow et al. \cite{goodfellow2014generative-adversarial-networks}, $D$ is trained to tell apart real and fake images. This PPGN variant produces the best image quality and thus used for the main experiments in this paper (\secref{results}). After $G$ is trained, we
		sample from this model following an iterative sampling procedure described in \secref{noiseless_ppgn}. \textbf{c)} For the Joint PPGN-$h$ variant (\secref{default_ppgn}),
		we train the entire model as being composed of 3 interleaved DAEs respectively for $x$, $h_1$ and $h$. In other words, we add noise to each of these variables and train the corresponding AE with a $L_2$ reconstruction loss. 
		The loss for $D$ remains the same as in (a), while the loss for $G$ is now composed of 4 components: $L = L_{img} + L_{h_1} + L_{h} + L_{GAN}$. 
		The sampling procedure for this PPGN variant is provided in \secref{default_ppgn}.
		See \secref{train_details} for more training and architecture details of the two PPGN variants.
	}
	\figlabel{training_diagram}
\end{figure*}

\begin{figure*}[h]
	\hspace{0.5em}
	(a) Real: top 9                                                 \hspace{3.5em}
	(b) DGN-AM \cite{nguyen-2016-synthesizing-the-preferred-inputs} \hspace{2.7em}
	(c) Real: random 9                                               \hspace{3.0em}
	(d) PPGN (\textbf{this})
	\centering
	\includegraphics[width=1.8\columnwidth]{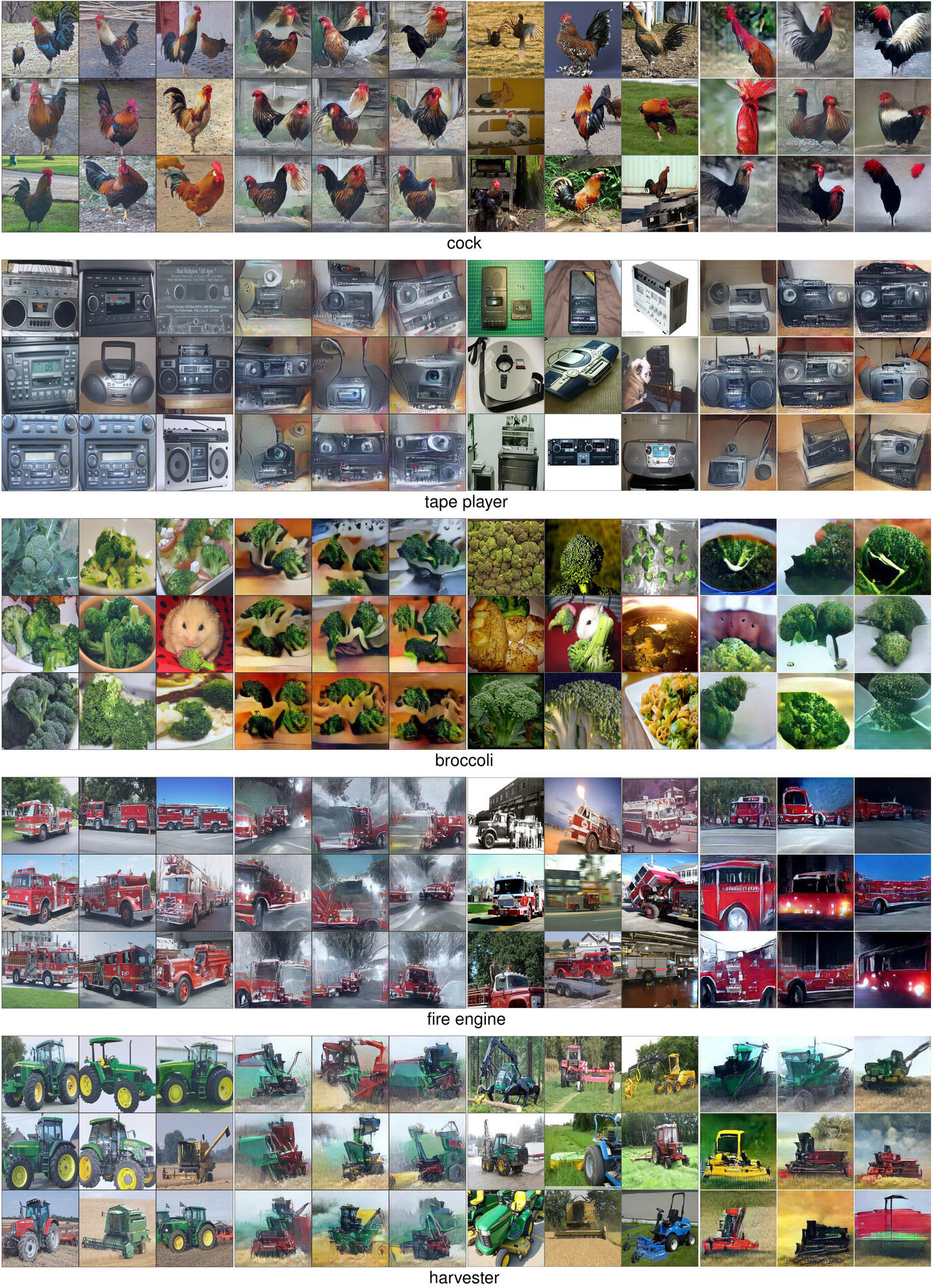}
	\caption{(a) The 9 training set images that most highly activate a given class output neuron (e.g. fire engine). (b) DGN-AM \cite{nguyen-2016-synthesizing-the-preferred-inputs} synthesizes high-quality images, but they often converge to the mode of high-activating images (the top-9 mode). (c) 9 training set images randomly picked from the same class.  (d) Our new method PPGN produces samples with better quality and substantially larger diversity than DGN-AM, thus better representing the diversity of images from the class.
	}
	\figlabel{DGNAM_vs_PPGN}
	\vspace{0em}
\end{figure*}

\begin{figure*}
	\centering
	\begin{subfigure}{1.0\linewidth}
		\centering
		\includegraphics[width=0.95\columnwidth]{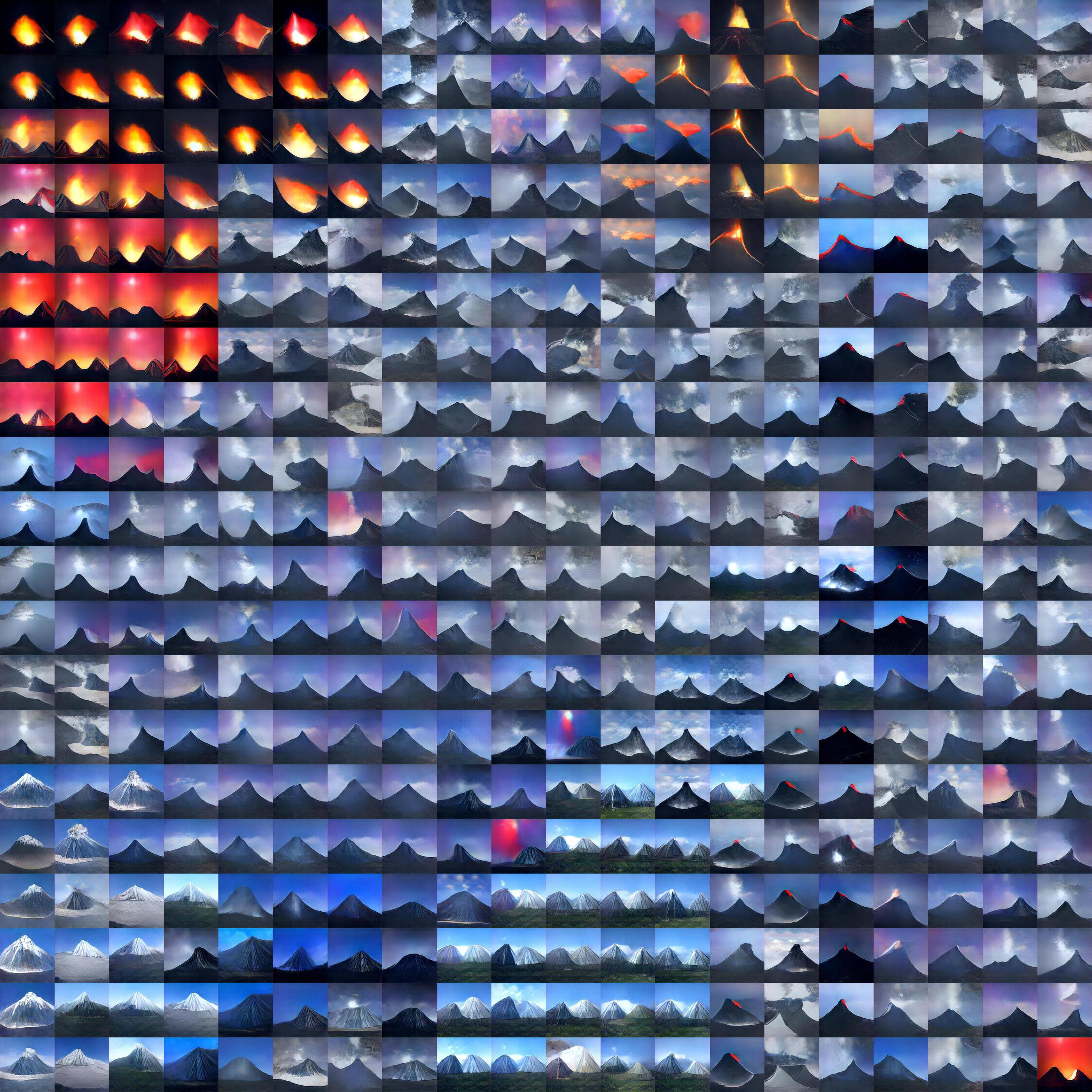}
		\caption{Samples produced by PPGN visualized in a grid t-SNE \cite{van-der-maaten2008visualizing-data-using}}.
		\vspace{0.2cm}
	\end{subfigure}		
	\hspace{1mm}
	\begin{subfigure}{1.0\linewidth}
		\centering
		\includegraphics[width=0.95\linewidth]{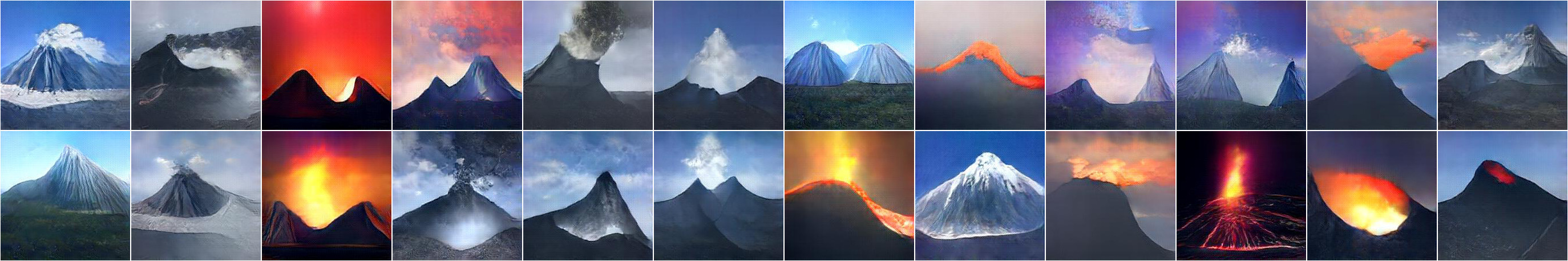}
		\caption{Samples hand-picked from (a) to showcase the diversity and quality of images produced by PPGN.}
		\vspace{0.2cm}
	\end{subfigure}	
	
	\caption{
		We qualitatively evaluate sample diversity by running 10 sampling chains (conditioned on class ``volcano''), each for 200 steps, to produce 2000 samples, and filtering out samples with class probability of less than $0.97$. From the remaining, we randomly pick 400 samples and plot them in a grid t-SNE \cite{van-der-maaten2008visualizing-data-using} (top panel). From those, we chose a selection to highlight the quality and diversity of the samples (bottom panel). There is a tremendous amount of detail in each image and diversity across images. Samples include dormant volcanos and active volcanoes with smoke plumes of different colors from white to black to fiery orange. Some have two peaks and others one, and underneath are scree, green forests,  or glaciers (complete with crevasses). The sky changes from different shades of mid-day blue through different sunsets to pitch black night.
	}
	\figlabel{tsne_volcano}
	\vspace{0em}
\end{figure*}

\begin{figure*}
	\centering
	\begin{subfigure}{1.0\linewidth}
		\centering
		\includegraphics[width=0.95\columnwidth]{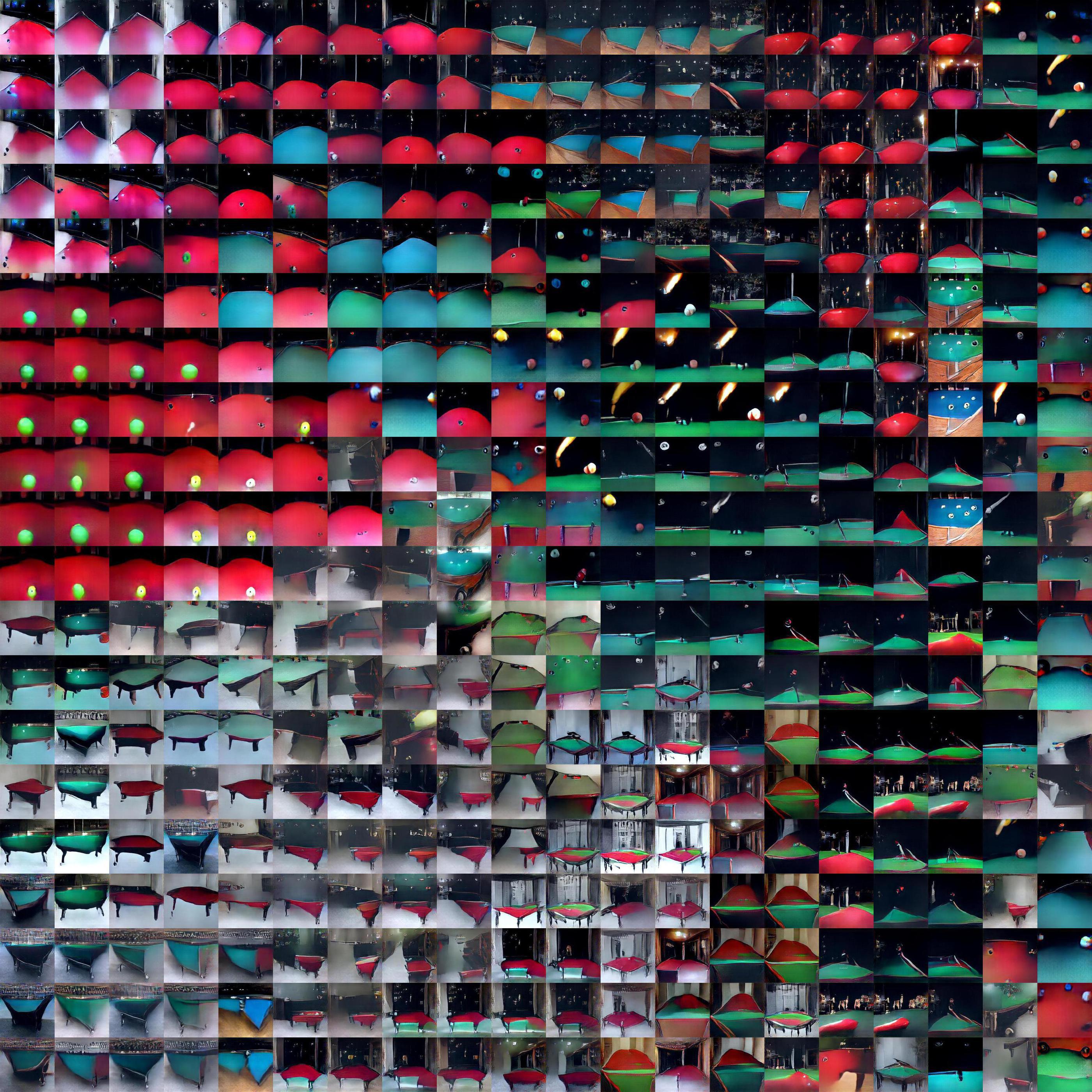}
		\caption{Samples produced by PPGN visualized in a grid t-SNE \cite{van-der-maaten2008visualizing-data-using}}.
		\vspace{0.2cm}
	\end{subfigure}		
	\hspace{1mm}
	\begin{subfigure}{1.0\linewidth}
		\centering
		\includegraphics[width=0.95\linewidth]{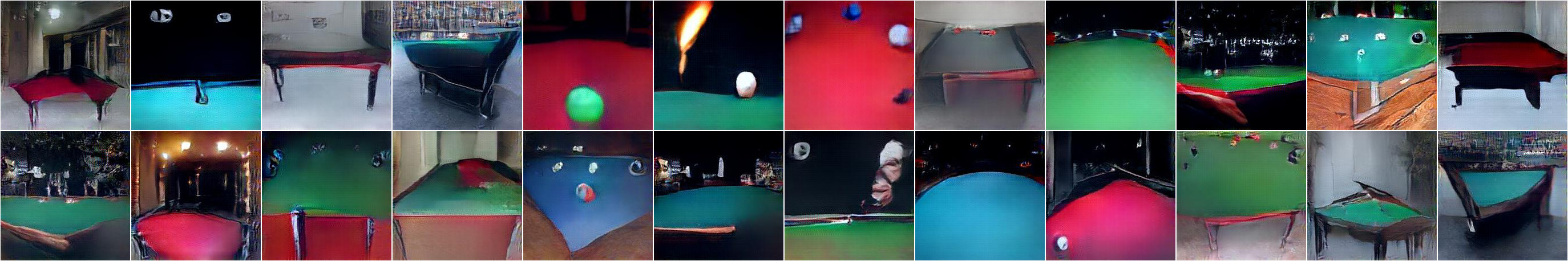}
		\caption{Samples hand-picked from (a) to showcase the diversity and quality of images produced by PPGN.}
		\vspace{0.2cm}
	\end{subfigure}	
	
	\caption{
		The figures are selected and plotted in the same way as \figref{tsne_volcano}, but here for the ``pool table'' class. Once again, we observe a high degree of both image quality and diversity. Different felt colors (green, blue, and red), lighting conditions, camera angles, and interior designs are apparent. 
	}
	\figlabel{tsne_pooltable}
	\vspace{0em}
\end{figure*}

\begin{figure*}
	\centering
	\begin{subfigure}{1.0\linewidth}
		\centering
		\includegraphics[width=1.0\linewidth]{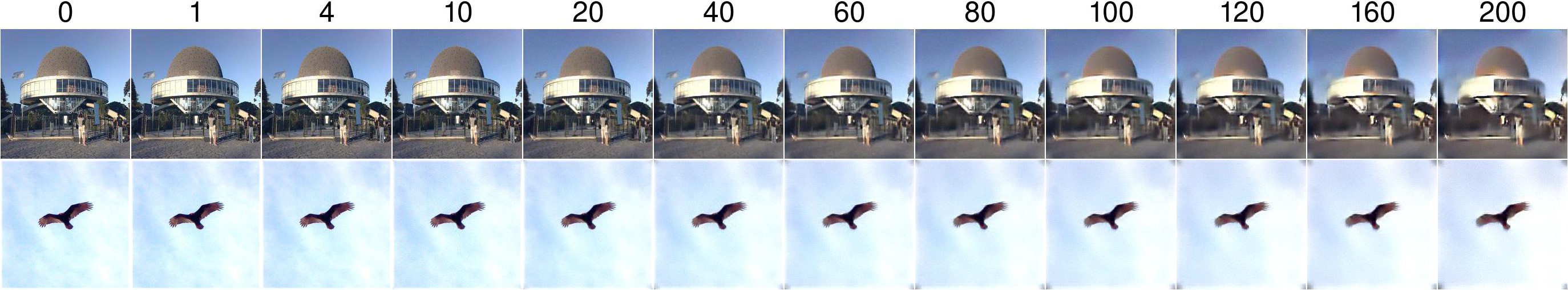}
		\caption{PPGN-$x$ with a DAE model of $p(x)$}
		\vspace{0.2cm}
		\figlabel{sampling_x_image}
	\end{subfigure}	
	\hspace{1mm}
	\begin{subfigure}{1.0\linewidth}
		\centering
		\includegraphics[width=1.0\linewidth]{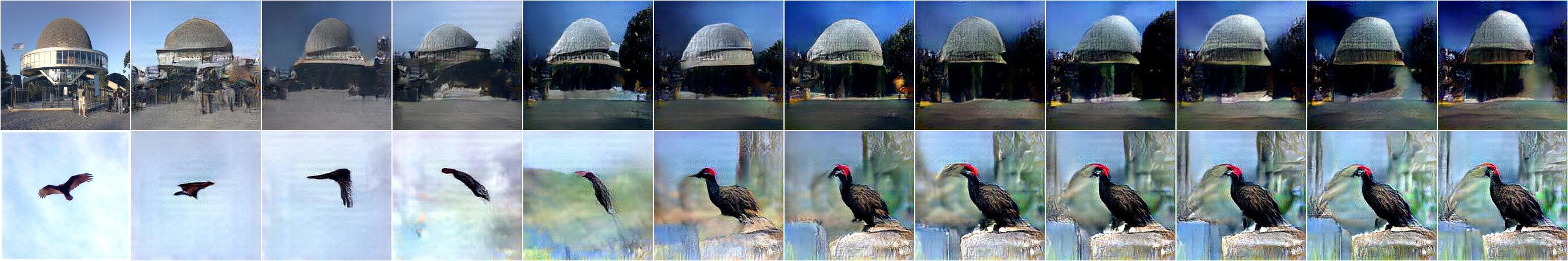}
		\caption{DGN-AM \cite{nguyen-2016-synthesizing-the-preferred-inputs} (which has a hand-designed Gaussian $p(h)$ prior)}
		\vspace{0.2cm}		
		\figlabel{sampling_DGNAM_image}
	\end{subfigure}		
	\hspace{1mm}
	\begin{subfigure}{1.0\linewidth}
		\centering
		\includegraphics[width=1.0\linewidth]{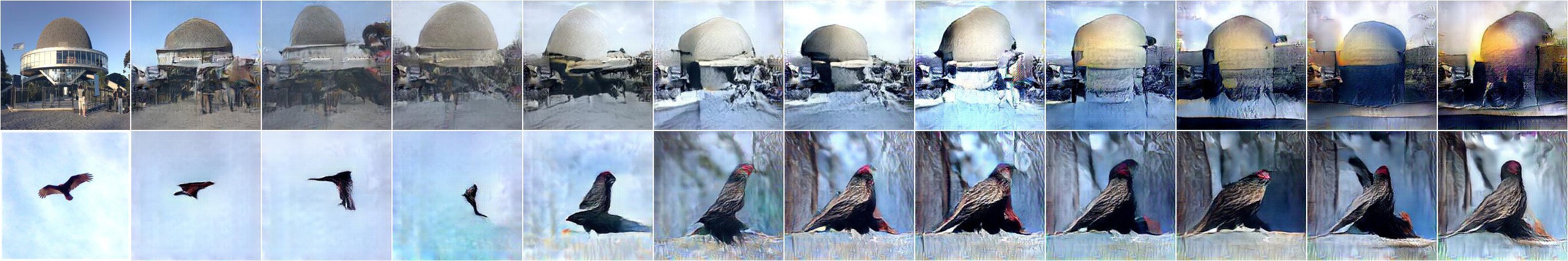}
		\caption{PPGN-$h$: Generator and multi-layer perceptron DAE model of $p(h)$}
		\vspace{0.2cm}
		\figlabel{sampling_h_image}
	\end{subfigure}	
	\hspace{1mm}
	\begin{subfigure}{1.0\linewidth}
		\centering
		\includegraphics[width=1.0\linewidth]{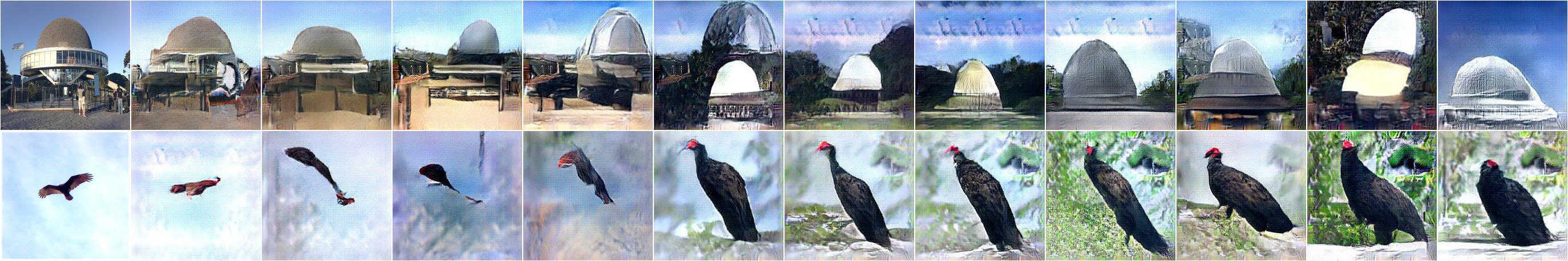}
		\caption{Joint PPGN-$h$: joint Generator and DAE}
		\vspace{0.2cm}		
		\figlabel{sampling_inter_image}
	\end{subfigure}	
	\hspace{1mm}
	\begin{subfigure}{1.0\linewidth}
		\centering
		\includegraphics[width=1.0\linewidth]{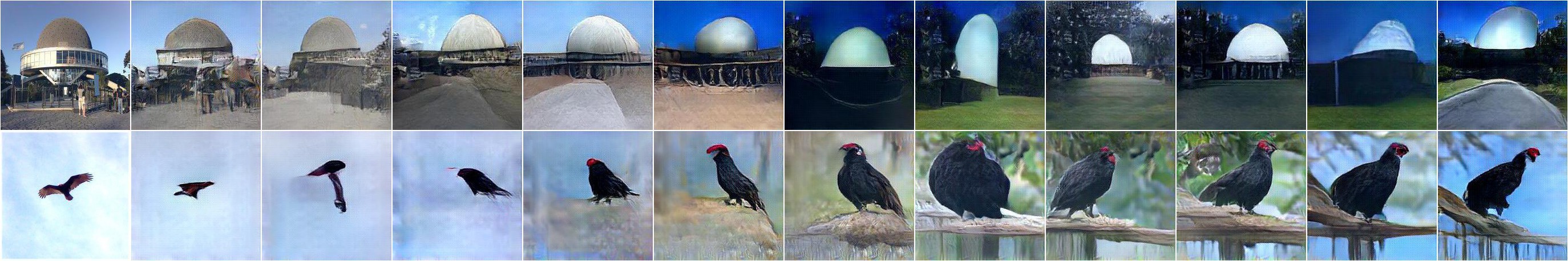}
		\caption{Noiseless Joint PPGN-$h$: joint Generator and AE}
		\vspace{0.2cm}		
		\figlabel{sampling_PPGN_image}
	\end{subfigure}			
	\caption{
		A comparison of samples generated from a single sampling chain (starting from a real image on the left) across different models. Each panel shows two sampling chains for that model: one conditioned on the ``planetarium'' class and the other conditioned on the ``kite'' (a type of bird) class. The iteration number of the sampling chain is shown on top.
		(a) The sampling chain in the image space mixes poorly.
		(b) The sampling chain from DGN-AM \cite{nguyen-2016-synthesizing-the-preferred-inputs} (in the $h$ code space with a hand-designed Gaussian $p(h)$ prior) produces better images, but still mixes poorly, as evidenced by similar samples over many iterations. (c) To improve sampling, we tried swapping in a $p(h)$ model represented by a 7-layer DAE for $h$. However, the sampling chain does not mix faster or produce better samples. (d) We experimented with a better way to model $p(h)$, i.e. modeling $h$ via $x$. We treat the generator $G$ and encoder $E$ as an autoencoder for $h$ and call this treatment ``Noiseless Joint PPGN-$h$'' (see \secref{noiseless_ppgn}). This is also our best model that we use for experiments in \secref{results}. 
		This substantially improves the mixing speed and sample quality.
		(e) We train the entire model as being composed of 3 DAEs and sample from it by adding noise to the image, \layer{fc6} and \layer{pool5} variables. The chain mixes slightly faster compared to (d), but generates slightly worse samples.
	}
	\figlabel{sampling_class_from_image}
\end{figure*}

\begin{figure*}
	\centering
	\begin{subfigure}{1.0\linewidth}
		\centering
		\includegraphics[width=1.0\linewidth]{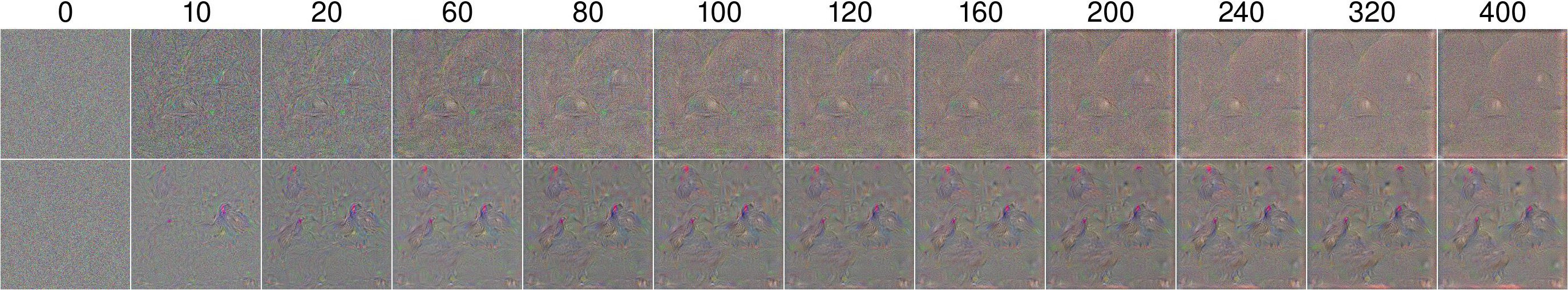}
		\caption{PPGN-$x$ with a DAE model of $p(x)$}
		\vspace{0.2cm}
		\figlabel{sampling_x_random}
	\end{subfigure}	
	\hspace{1mm}
	\begin{subfigure}{1.0\linewidth}
		\centering
		\includegraphics[width=1.0\linewidth]{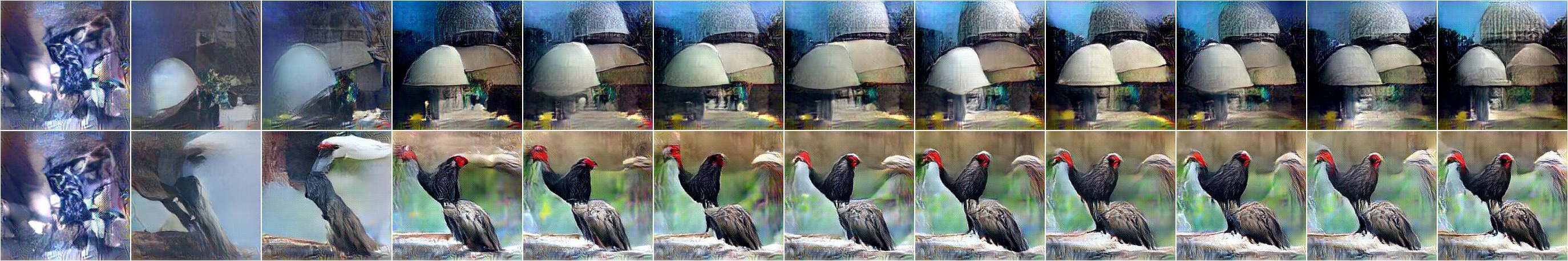}
		\caption{DGN-AM \cite{nguyen-2016-synthesizing-the-preferred-inputs} (which has a hand-designed Gaussian $p(h)$ prior)}
		\vspace{0.2cm}		
		\figlabel{sampling_DGNAM_random}
	\end{subfigure}		
	\hspace{1mm}
	\begin{subfigure}{1.0\linewidth}
		\centering
		\includegraphics[width=1.0\linewidth]{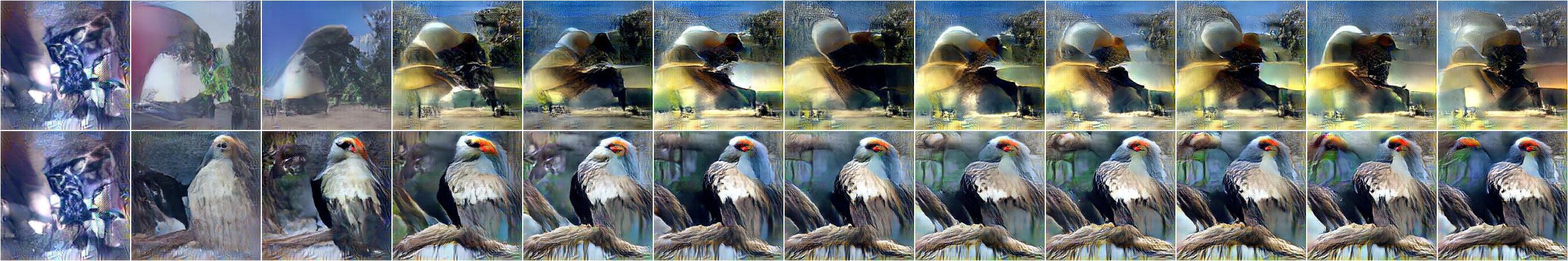}
		\caption{PPGN-$h$: Generator and a multi-layer perceptron DAE model of $p(h)$}
		\vspace{0.2cm}		
		\figlabel{sampling_h_random}
	\end{subfigure}	
	\hspace{1mm}
	\begin{subfigure}{1.0\linewidth}
		\centering
		\includegraphics[width=1.0\linewidth]{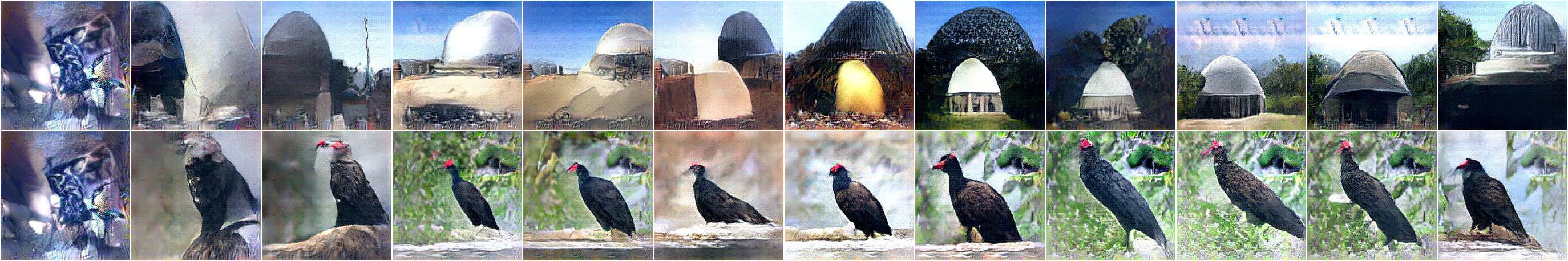}
		\caption{Joint PPGN-$h$: joint Generator and DAE}
		\vspace{0.2cm}		
		\figlabel{sampling_inter_random}
	\end{subfigure}	
	\hspace{1mm}
	\begin{subfigure}{1.0\linewidth}
		\centering
		\includegraphics[width=1.0\linewidth]{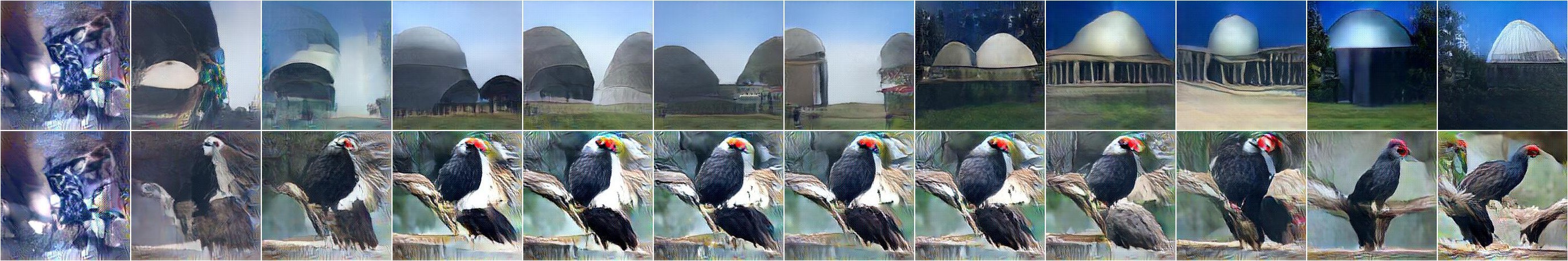}
		\caption{Noiseless Joint PPGN-$h$: joint Generator and AE}
		\vspace{0.2cm}		
		\figlabel{sampling_PPGN_random}
	\end{subfigure}			
	\caption{
		Same as \figref{sampling_class_from_image}, but starting from a random code $h$ (which when pushed through generator network $G$ produces the leftmost images) except for (a) which starts from random images as the sampling operates directly in the pixel space. All of our qualitative conclusions are the same as for \figref{sampling_class_from_image}. Note that the samples in (b) appear slightly worse than the images reported in Nguyen et al. \cite{nguyen-2016-synthesizing-the-preferred-inputs}. The reason is that in the new framework introduced in this paper we perform an infinitely long sampling chain at a constant learning rate to travel from one mode to another in the space. In contrast, 
		the set of parameters (including the number of iterations, an $L_2$ decay on code $h$, and a learning rate decay) in Nguyen et al. \cite{nguyen-2016-synthesizing-the-preferred-inputs}
		is carefully tuned for the best image quality, but does not allow for a long sampling chain (\figref{cardoon}).
	}
	\figlabel{sampling_class_from_random}
\end{figure*}

\begin{figure*}
	\centering
	\begin{subfigure}{1.0\linewidth}
		\centering
		\includegraphics[width=1.0\linewidth]{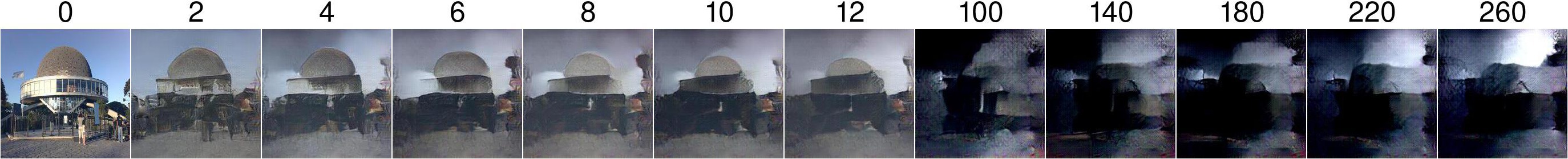}
		\caption{Very large noise ($\epsilon_3=10^{-1}$)}
		\vspace{0.2cm}
	\end{subfigure}		
	\hspace{1mm}
	\begin{subfigure}{1.0\linewidth}
		\centering
		\includegraphics[width=1.0\linewidth]{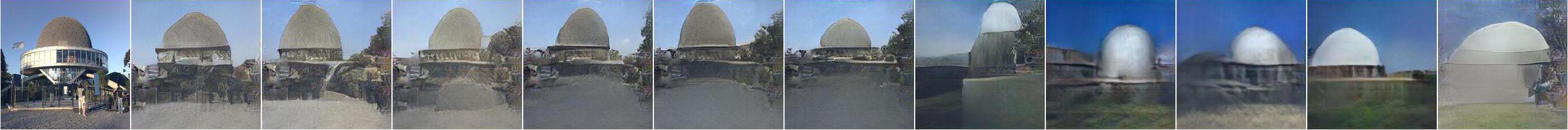}
		\caption{Large noise ($\epsilon_3=10^{-5}$)}
		\vspace{0.2cm}
	\end{subfigure}	
	\hspace{1mm}
	\begin{subfigure}{1.0\linewidth}
		\centering
		\includegraphics[width=1.0\linewidth]{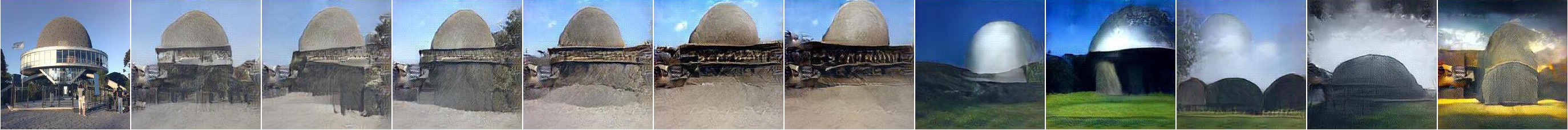}
		\caption{Medium noise ($\epsilon_3=10^{-9}$)}
		\vspace{0.2cm}
	\end{subfigure}	
	\hspace{1mm}
	\begin{subfigure}{1.0\linewidth}
		\centering
		\includegraphics[width=1.0\linewidth]{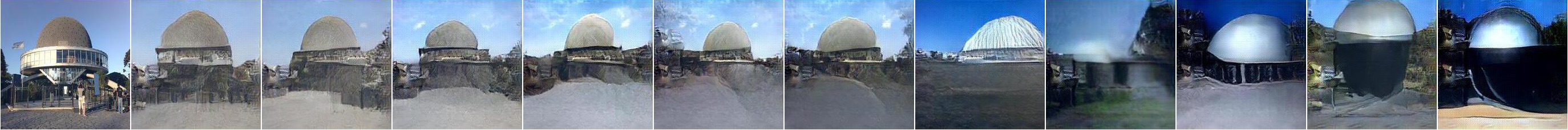}
		\caption{Small noise ($\epsilon_3=10^{-13}$)}
		\vspace{0.2cm}		
	\end{subfigure}	
	\hspace{1mm}
	\begin{subfigure}{1.0\linewidth}
		\centering
		\includegraphics[width=1.0\linewidth]{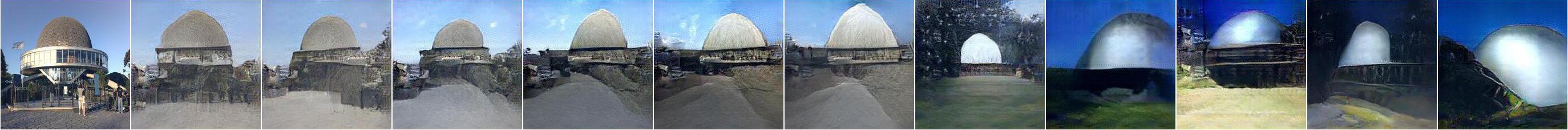}
		\caption{Infinitesimal noise ($\epsilon_3=10^{-17}$)}
		\vspace{0.2cm}
	\end{subfigure}	
	\caption{
		Sampling chains with the \emph{noiseless} PPGN model starting from the code of a real image (\emph{left}) and conditioning on class ``planetarium'' i.e. $(\epsilon_1, \epsilon_2) = (1, 10^{-5})$ for different noise levels $\epsilon_3$. The sampling step numbers are shown on top. Samples are better with a tiny amount of noise (e) than with larger noise levels (a,b,c \& d), so we chose that as our default noise level for all sampling experiments with the Noiseless Joint PPGN-$h$ variant (\secref{noiseless_ppgn}).
		These results suggest that a certain amount of noise added to the DAE during training might help the chain mix faster, and thus partly motivated our experiment in \secref{default_ppgn}.
	}
	\figlabel{noise_sweep}
\end{figure*}

\begin{figure*}
	\centering
	\includegraphics[width=1.85\columnwidth]{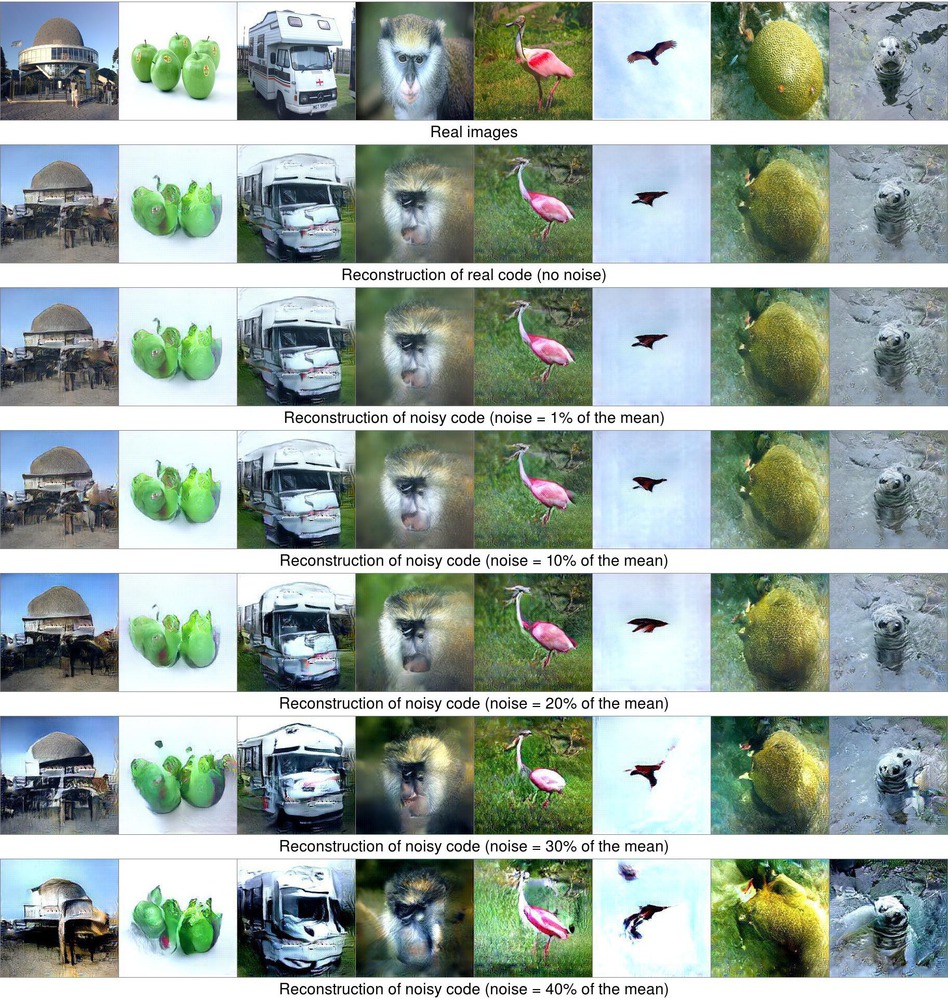}
	\caption{
		The default generator network $G$ in our experiments (used in Sections \ref{sec:ppgn_h} \& \ref{sec:noiseless_ppgn}) was trained to reconstruct images from compressed \layer{fc6} features extracted from AlexNet classification network \cite{krizhevsky2012imagenet-classification-with-deep} with three different losses: 
		adversarial loss \cite{goodfellow2014generative-adversarial-networks}, feature matching loss \cite{dosovitskiy-2016-NIPS-generating-images-with}, and image reconstruction loss (more training details are in \secref{train_PPGN}). Here, we test how robust $G$ is to Gaussian noise added to an input code $h$ of a real image.
		We sweep across different levels of Gaussian noise $N(0,\sigma^2)$ with $\sigma=\{1\%, 10\%, 20\%, 30\%, 40\%\}$ of the mean \layer{fc6} activation computed by the activations of validation set images.
		We observed that $G$ is robust to even a large amount of noise up to $\sigma=20\%$ despite being trained without explicit regularizations (i.e. with noise \cite{vincent-2008-ICML-extracting-and-composing-robust} or a contractive penalty \cite{rifai2011contractive}).
	}
	\figlabel{noise_robustness}
	\vspace{0em}
\end{figure*}

\begin{figure*}
	\centering
	\begin{subfigure}{1.0\linewidth}
		\centering
		\includegraphics[width=1.0\linewidth]{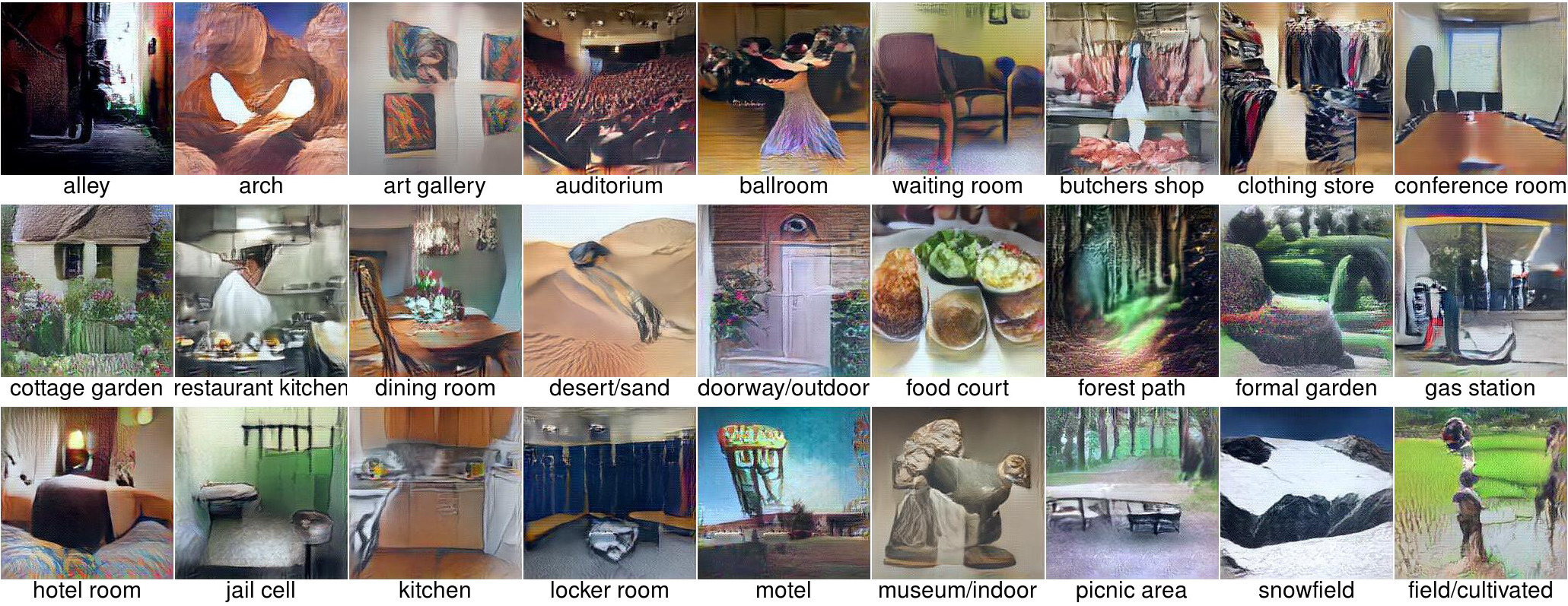}
		\caption{Samples produced by the DGN-AM method \cite{nguyen-2016-synthesizing-the-preferred-inputs}}
		\vspace{0.2cm}
	\end{subfigure}	
	\hspace{1mm}
	\begin{subfigure}{1.0\linewidth}
		\centering
		\includegraphics[width=1.0\linewidth]{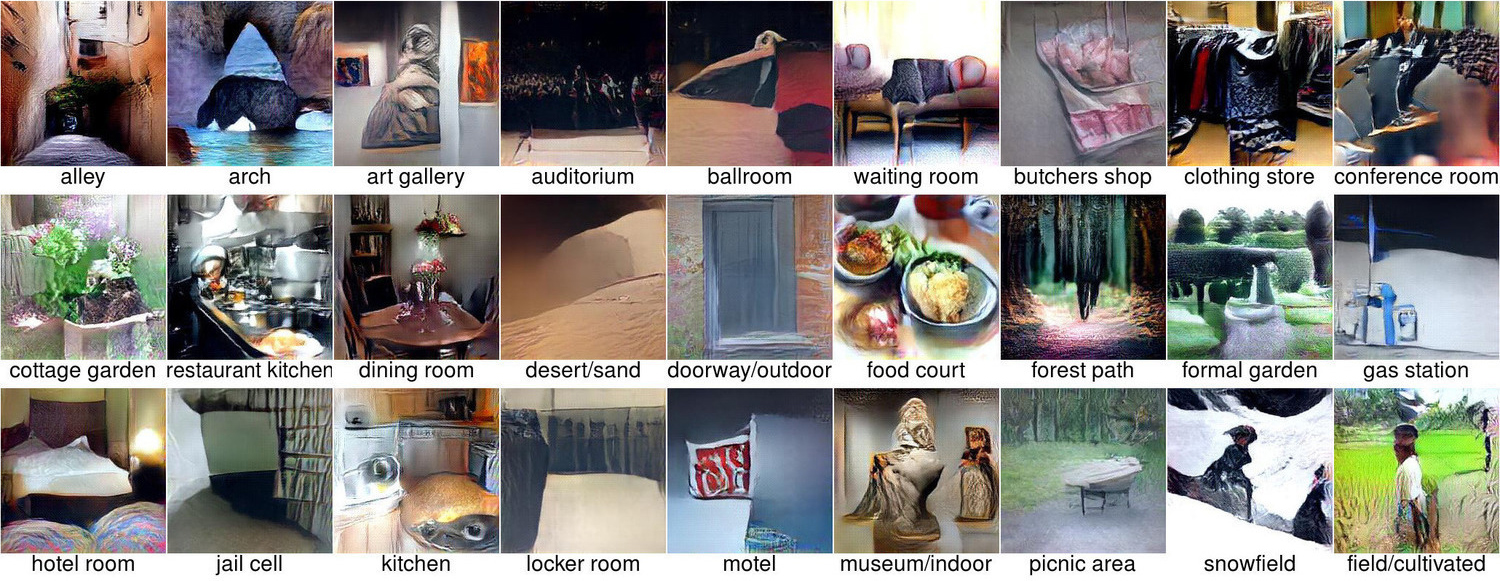}
		\caption{Samples produced by PPGN (the new model proposed in this paper)}
		\vspace{0.2cm}
	\end{subfigure}	
	\caption{
		A comparison of images produced by the DGN-AM method \cite{nguyen-2016-synthesizing-the-preferred-inputs} (top) and the new PPGN method we introduce in this paper (bottom). Both methods synthesize images conditioned on classes of scene images that the generator was never trained on. Specifically, the condition model $p(y|x)$ is AlexNet trained to classify 205 categories of scene images from the MIT Places dataset \cite{zhou-2014-arXiv-object-detectors-emerge}, while the prior model $p(x)$ 
		is trained to generate ImageNet images. Despite having a strong, learned prior (represented by a DAE trained on ImageNet images), the PPGN (like DGN-AM) produces high-quality images for an unseen dataset.
	}
	\figlabel{mit_places_DGNAM_vs_PPGN}
\end{figure*}

\begin{figure*}
	\centering
	\includegraphics[width=1.75\columnwidth]{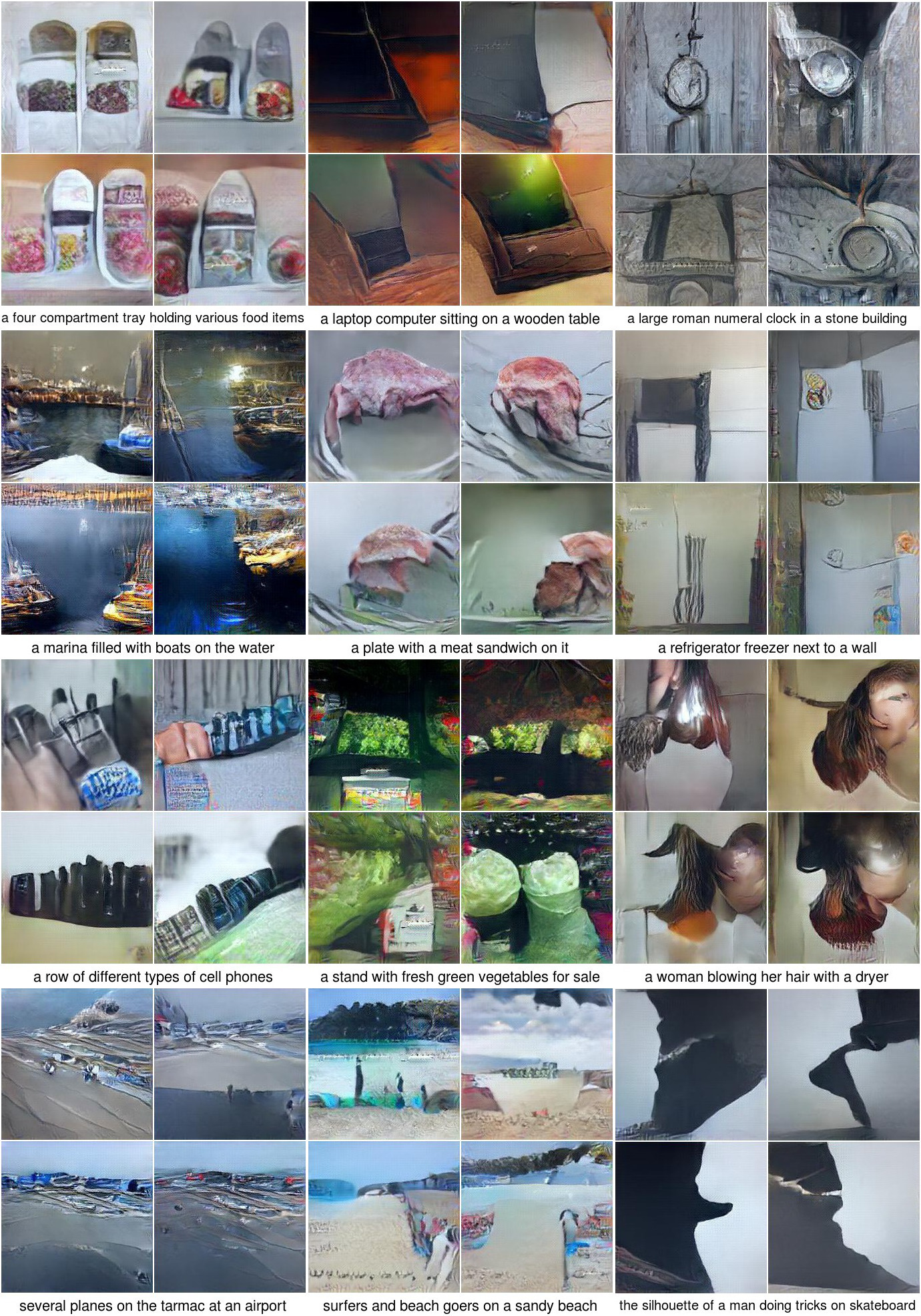}
	\caption{
		The model can be given a text description of an image and asked to generate the described image. Technically, that involves the same PPGN model, but conditioning on a caption instead of a class. Here the condition network is the LRCN image captioning model from Donahue et al. \cite{donahue-2014-arXiv-long-term-recurrent-convolutional}, which can generate reasonable captions for images. For each caption, we show 4 images synthesized by starting from random initializations. Note that it reasonably draws ``tarmac'', ``silhouette'' or ``woman'' although these are not categories in the ImageNet dataset \cite{deng2009imagenet:-a-large-scale-hierarchical}.
	}
	\figlabel{more_image_captioning}
\end{figure*}

\begin{figure*}
	\centering
	\includegraphics[width=2.0\columnwidth]{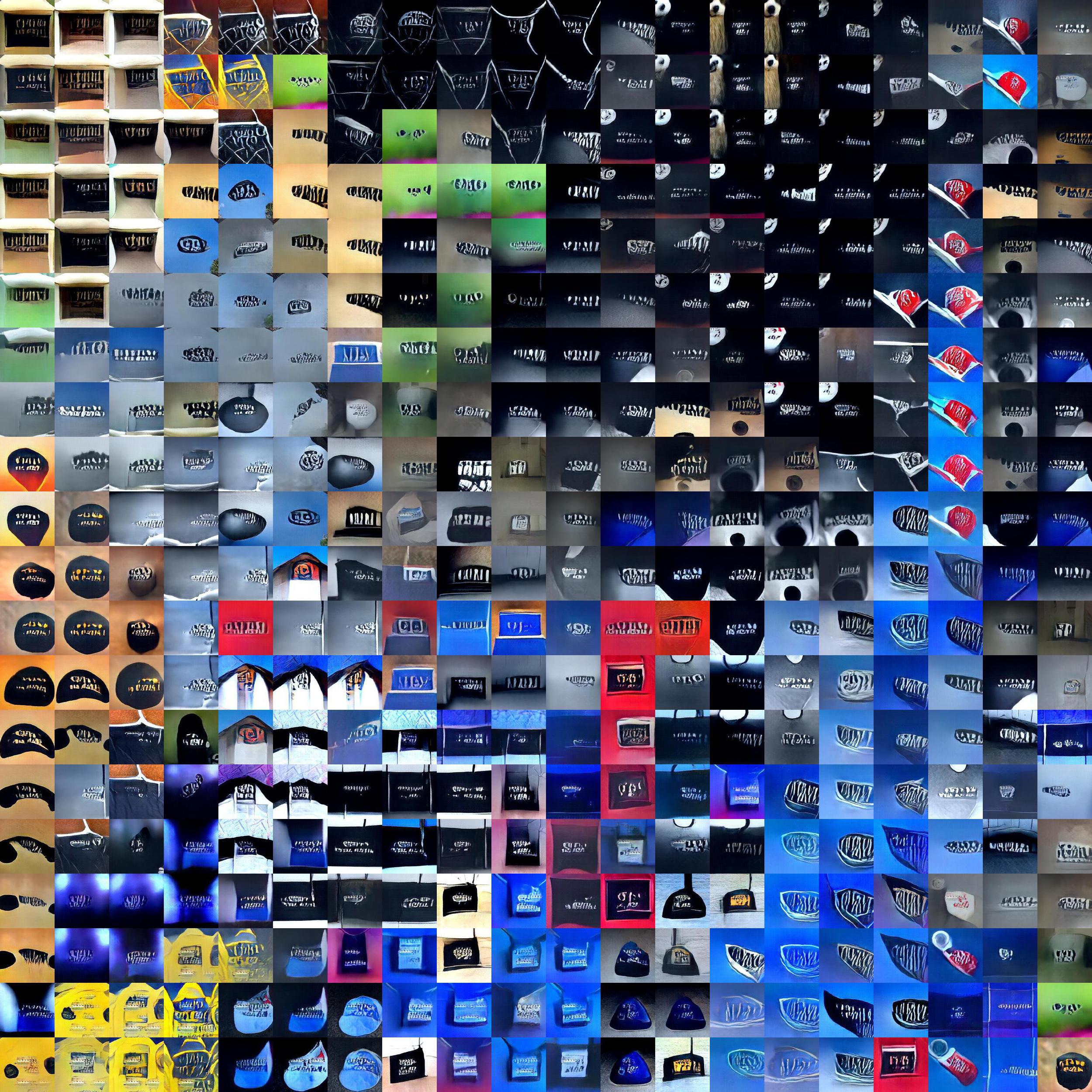}
	\caption{
		PPGNs have the ability to perform `multifaceted feature visualization,' meaning they can generate the set of inputs that activate a given hidden neuron, which improves our ability to understand what type of features that neuron has learned to detect \cite{nguyen-2016-arXiv-multifaceted-feature-visualization:,yosinski-2015-ICML-DL-understanding-neural-networks}.
		To demonstrate that capability, instead of conditioning on a class from the dataset, here we generate images conditioned on a hidden neuron previously identified as detecting text \cite{yosinski-2015-ICML-DL-understanding-neural-networks}: neuron number \unit{243} in layer \layer{conv5} of the AlexNet DNN.
		We run 10 sampling chains, each for 200 steps, to produce 2000 samples, and filtering out samples with a softmax probability 				(taken over all depth columns at the same spatial location) of less than $0.97$. From the remaining, we randomly pick 400 samples and plot them in a grid t-SNE \cite{van-der-maaten2008visualizing-data-using}.
		These images can be interpreted as the preferred stimuli for this text detector unit \cite{nguyen-2016-synthesizing-the-preferred-inputs}. The diversity of samples is substantially improved vs. previous methods \cite{nguyen-2016-arXiv-multifaceted-feature-visualization:,nguyen-2016-synthesizing-the-preferred-inputs,yosinski-2015-ICML-DL-understanding-neural-networks} uncovering different facets that the neuron has learned to detect. In other words, while previous methods produced one type of sample per neuron \cite{yosinski-2015-ICML-DL-understanding-neural-networks,nguyen-2016-synthesizing-the-preferred-inputs}, or lower quality samples with greater diversity \cite{nguyen-2016-arXiv-multifaceted-feature-visualization:}, PPGNs produce a diversity of high-quality samples, and thus represent the state of the art in multifaceted feature visualization.
	}
	\figlabel{conv5_text_detector}
\end{figure*}

\begin{figure*}
	\centering
	\includegraphics[width=2.0\columnwidth]{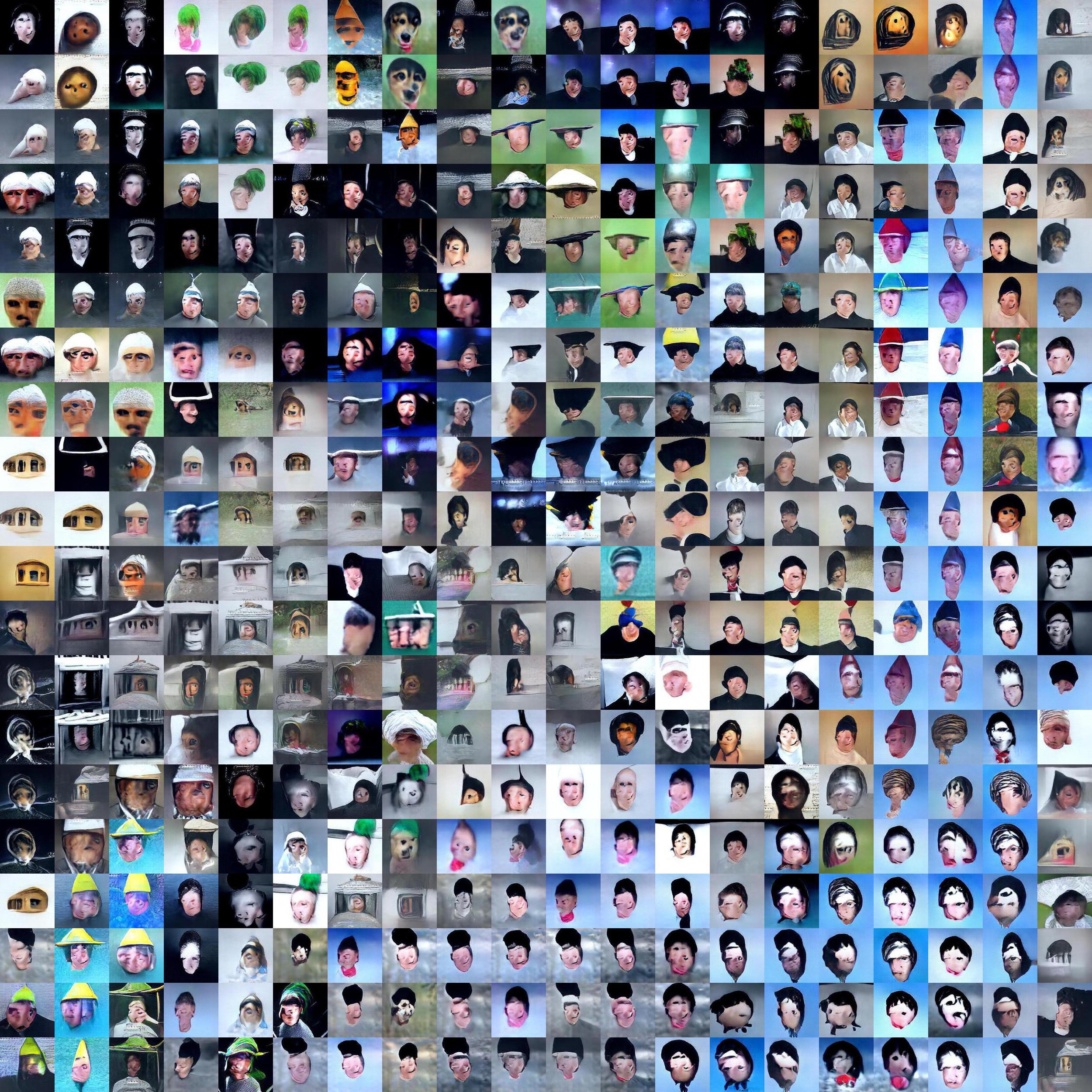}
	\caption{
		This figure shows the same thing as \figref{conv5_text_detector}, except in this case for a hidden neuron previously identified to be a face detector \cite{yosinski-2015-ICML-DL-understanding-neural-networks} neuron (number \unit{196}) in layer \layer{conv5} of the AlexNet DNN.
		One can see different types of faces that the neuron has learned to detect, including everything from dog faces (top row) to masks (left columns), and human faces from different angles, against different backgrounds, with and without hats, and with different shirt colors. Interestingly, we see that certain types of houses with windows that resemble eye sockets also activate this neuron (center left). This demonstrates the value of PPGNs as tools to identify unexpected facets, which aids in understanding the network, predicting some failure cases, and providing hints for how the network may be improved.
	}
	\figlabel{conv5_face_detector}
\end{figure*}

\begin{figure*}
	\centering
	\begin{subfigure}{1.0\linewidth}
		\centering
		\includegraphics[width=1.0\linewidth]{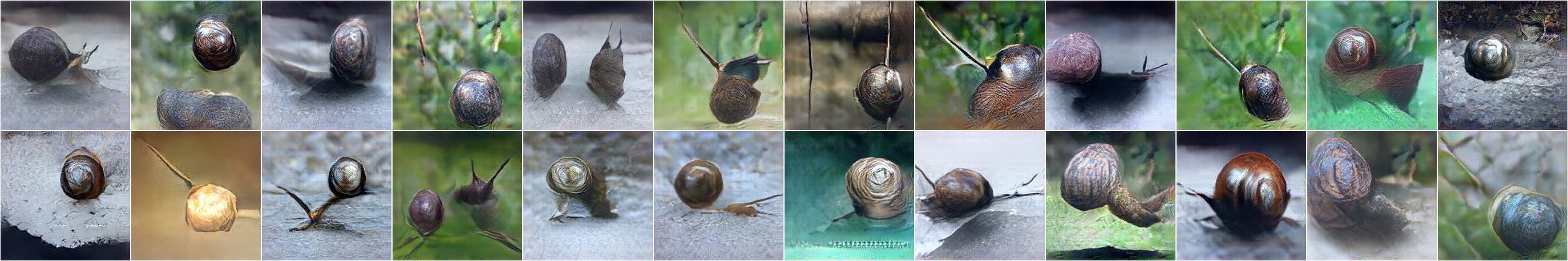}
		\caption{Snail}
		\vspace{0.2cm}
	\end{subfigure}		
	\hspace{1mm}
	\begin{subfigure}{1.0\linewidth}
		\centering
		\includegraphics[width=1.0\linewidth]{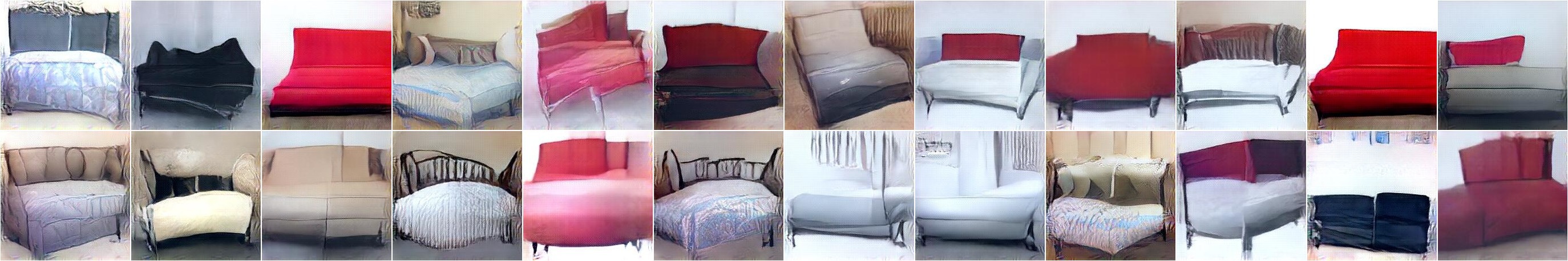}
		\caption{Studio couch}
		\vspace{0.2cm}
	\end{subfigure}	
	\hspace{1mm}
	\begin{subfigure}{1.0\linewidth}
		\centering
		\includegraphics[width=1.0\linewidth]{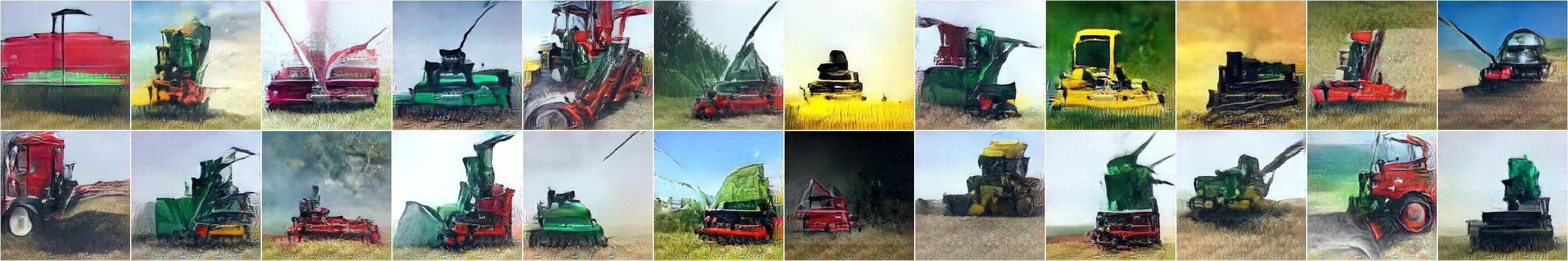}
		\caption{Harvester}
		\vspace{0.2cm}
	\end{subfigure}		
	\hspace{1mm}
	\begin{subfigure}{1.0\linewidth}
		\centering
		\includegraphics[width=1.0\linewidth]{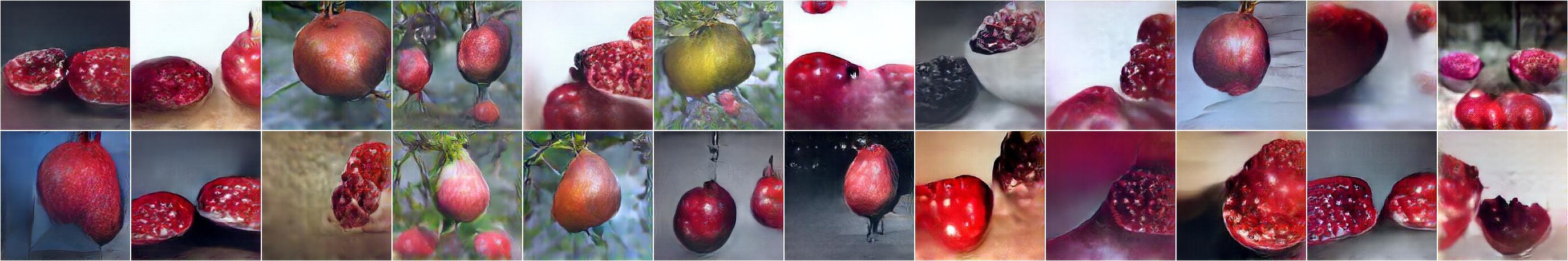}
		\caption{Pomegranate}
		\vspace{0.2cm}
	\end{subfigure}	
	\hspace{1mm}
	\begin{subfigure}{1.0\linewidth}
		\centering
		\includegraphics[width=1.0\linewidth]{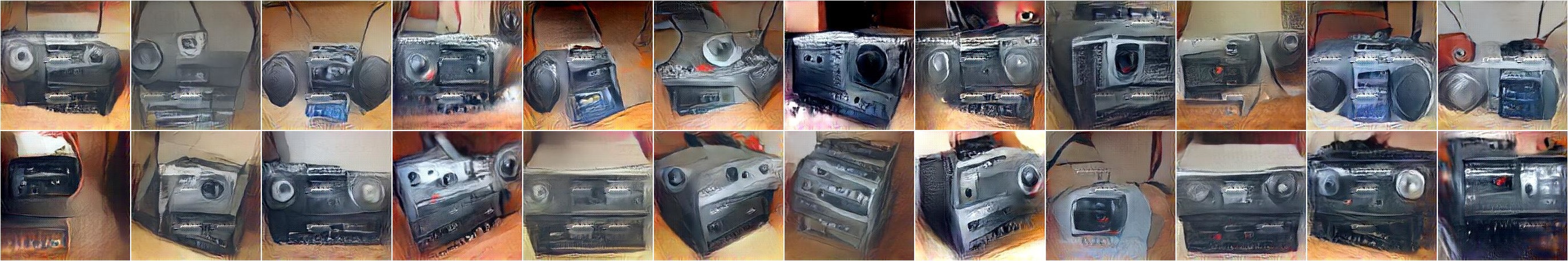}
		\caption{Tape player}
		\vspace{0.2cm}
	\end{subfigure}
	\caption{
		To evaluate the diversity of images within a class, here we show randomly chosen images from 5 different classes (a class label shown below each panel). Each image is the last sample produced from a 200-iteration sampling chain starting from a random initialization. The PPGN model is described in \secref{noiseless_ppgn}. We chose this method because it is simple, intuitive and straightforward to interpret and compare to other image generative models that do not require MCMC sampling. Another method to produce samples is to run a long sampling chain and record images that are produced at every sampling step; however, as done in \figref{tsne_volcano}, that method would require additional processing (including heuristic filtering and clustering) to obtain a set of different images because a well-known issue with MCMC sampling is that mixing is slow i.e. subsequent samples are often correlated. Note that one could obtain a larger diversity by running each sampling chain with a different set of parameters ($\epsilon$ multipliers in \eqnref{update_rule}); however, here we use the same set of parameters as previously reported in \secref{noiseless_ppgn} for simplicity and reproducibility.
	}
	\figlabel{diversity_random_images}
	\vspace{0em}
\end{figure*}

\begin{figure*}
	\centering
	\includegraphics[width=1.93\columnwidth]{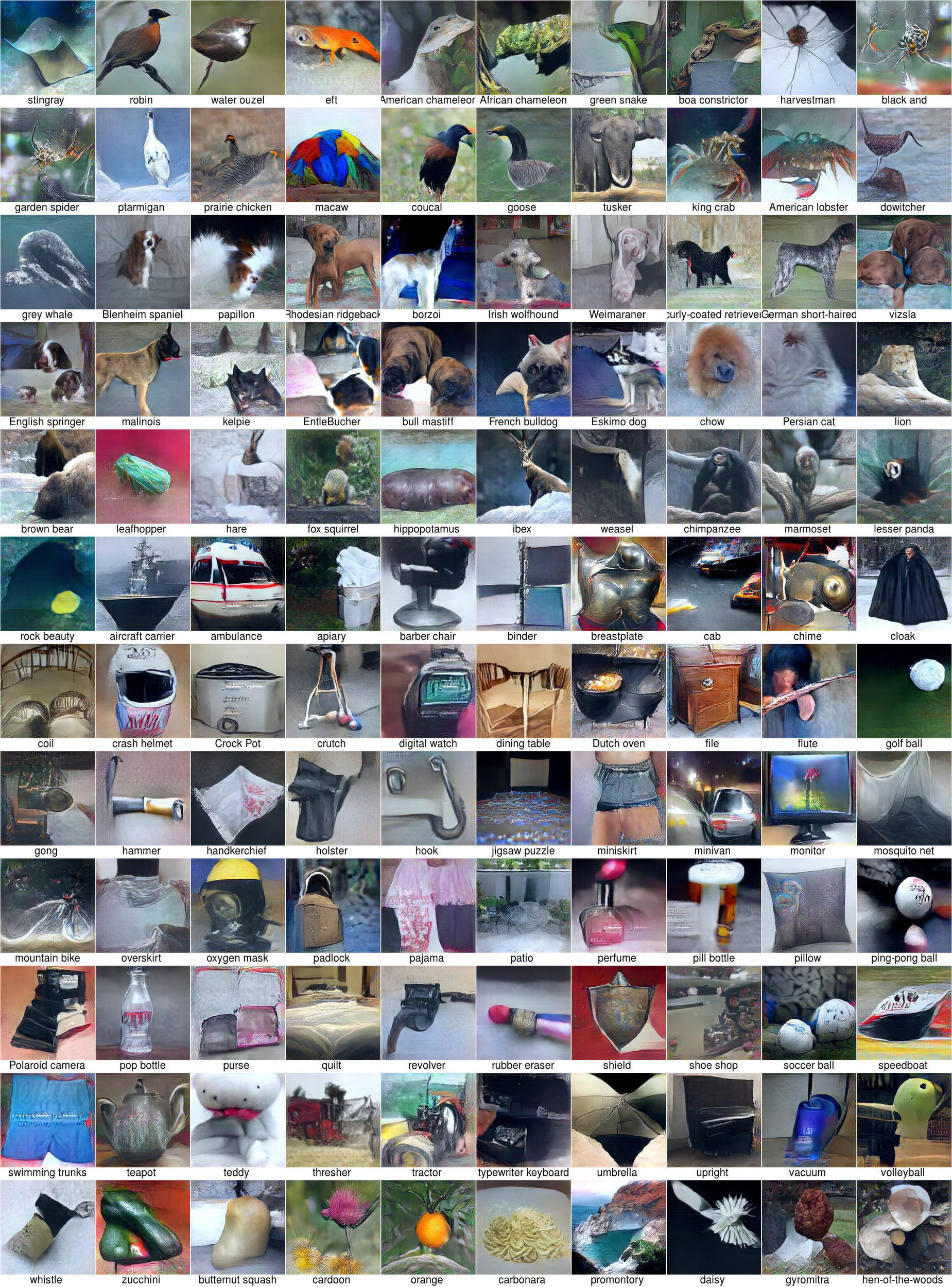}
	\caption{
		For a fair evaluation of the image quality produced by PPGN, here we show one randomly chosen sample for each of 120 random ImageNet classes (neither cherry-picked). Each image shown is the last sample produced from a 200-iteration sampling chain starting from a random initialization. The PPGN model is described in \secref{noiseless_ppgn}.
	}
	\label{fig:many_random}
\end{figure*}

\begin{figure*}
	\centering
	\begin{subfigure}{1.0\linewidth}
		\centering
		\includegraphics[width=1.0\linewidth]{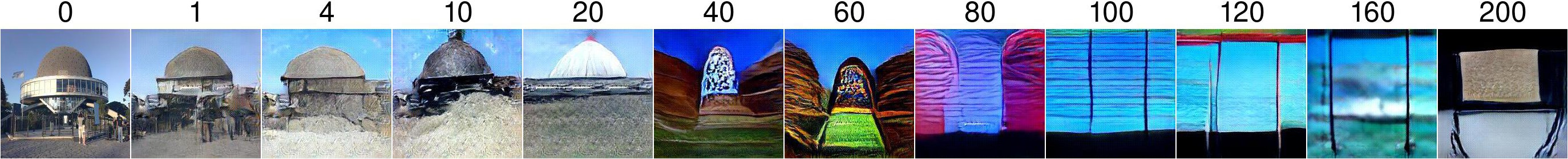}
		\caption{$\epsilon_1=10^{-1}$}
		\vspace{0.2cm}
	\end{subfigure}		
	\hspace{1mm}
	\begin{subfigure}{1.0\linewidth}
		\centering
		\includegraphics[width=1.0\linewidth]{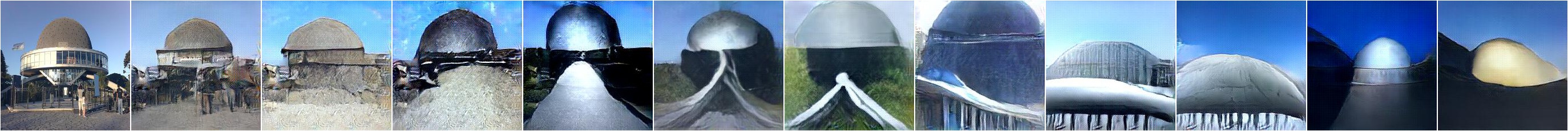}
		\caption{$\epsilon_1=10^{-3}$}
		\vspace{0.2cm}
	\end{subfigure}	
	\hspace{1mm}
	\begin{subfigure}{1.0\linewidth}
		\centering
		\includegraphics[width=1.0\linewidth]{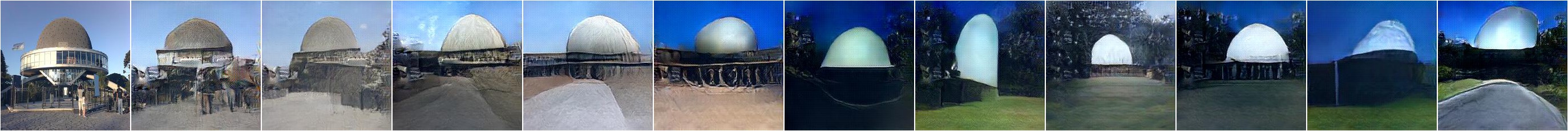}
		\caption{$\epsilon_1=10^{-5}$ (Noiseless Joint PPGN-$h$)}
		\vspace{0.2cm}
	\end{subfigure}	
	\hspace{1mm}
	\begin{subfigure}{1.0\linewidth}
		\centering
		\includegraphics[width=1.0\linewidth]{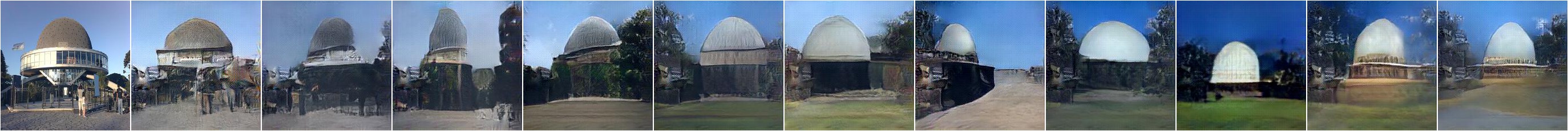}
		\caption{$\epsilon_1=10^{-7}$}
		\vspace{0.2cm}		
	\end{subfigure}	
	\hspace{1mm}
	\begin{subfigure}{1.0\linewidth}
		\centering
		\includegraphics[width=1.0\linewidth]{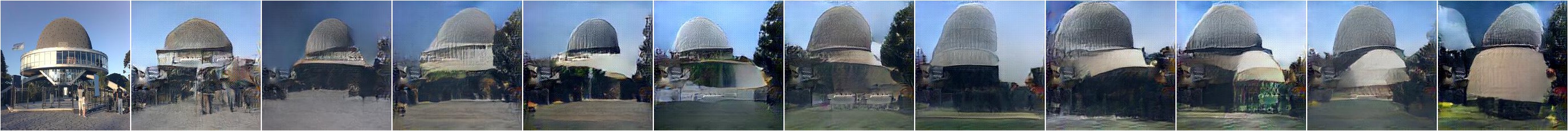}
		\caption{$\epsilon_1=10^{-11}$}
		\vspace{0.2cm}
	\end{subfigure}	
	\hspace{1mm}
	\begin{subfigure}{1.0\linewidth}
		\centering
		\includegraphics[width=1.0\linewidth]{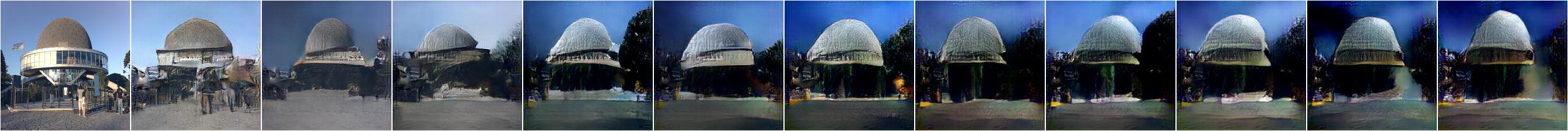}
		\caption{$\epsilon_1=0$ (no contribution from the prior)}
		\vspace{0.2cm}
	\end{subfigure}	
	\caption{
		To evaluate the effect of the prior term $\epsilon_1$ in the sampling, here we sweep across different values of this multiplier.
		We sample from the Noiseless Joint PPGN-$h$ model (\secref{noiseless_ppgn}) starting from the code of a real image (\emph{left}) and conditioning on class ``planetarium'' with a fixed amount of noise i.e. $(\epsilon_2, \epsilon_3) = (1, 10^{-17})$ for different $\epsilon_1$ values. The sampling step numbers are shown on top. Without the learned prior $p(h)$ (f), we arrive at the DGN-AM treatment results where the chain does not mix at all (the same result as in \figref{sampling_DGNAM_image}). Increasing $\epsilon_1$ up to a small value (c-e) results in a chain that mixes faster, from one planetarium to another. When the contribution of the prior is too high which overwrites the class gradients, yielding a chain that mixes from one mode of generic images to another (a). We empirically chose $\epsilon_1 = 10^{-5}$ as the default value for the Noiseless Joint PPGN-$h$ experiments in this paper as it produces the best image quality and diversity for many classes.
	}
	\figlabel{prior_sweep}
\end{figure*}

\section{Inpainting}
\label{sec:SI_inpainting}

We first randomly mask
out a $100\times100$ patch of a real $227\times227$ image $x_{real}$ (\figref{inpainting}a). The patch size is chosen following Pathak et al. \cite{pathak2016context}. We perform the same update rule as in \eqnref{ppgn_update_rule} (conditioning on a class, e.g. ``volcano''), but with an additional step updating image $x$ during the forward pass:

\vspace*{-.5em}
\beq
x = M\odot x + (1-M)\odot x_{real}
\eeq
\vspace*{-1.0em}

\noindent where $M$ is the binary mask for the corrupted patch, \mbox{$(1-M)\odot x_{real}$} is the uncorrupted area of the real image, and $\odot$ denotes the Hadamard (elementwise) product.
Intuitively, we clamp the observed parts of the synthesized image and
then sample only the unobserved portion in each pass. The 
DAE $p(h)$ model and the image classification network $p(y|h)$ model see progressively refined versions of the final, filled in image.
This approach tends to fill in semantically correct content, but it often fails to match the local details of the surrounding context (\figref{inpainting}b, the predicted pixels often do not transition smoothly to the surrounding context). An explanation is that we are sampling in the fully-connected \layer{fc6} feature space, which mostly encodes information of the global structure of objects instead of local details \cite{yosinski-2015-ICML-DL-understanding-neural-networks}.

To encourage the synthesized image to match the context of the real image, we
can add an extra condition in pixel space in the form of an additional term to the update rule in \eqnref{update_rule} to update $h$ in the direction of minimizing the cost:
$||(1-M) \odot x_{real} - (1-M)\odot x||^2_2$.
This helps the filled-in pixels match the surrounding context better (\figref{inpainting} b vs. c). Compared to the Context-Aware Fill feature in Photoshop CS6, which is based on the PatchMatch technique \cite{barnes2009patchmatch}, our method often performs worse in matching the local features of the surrounding context, but can fill in semantic objects better in many cases (\figref{inpainting}, bird \& bell pepper). More inpainting results are provided in the \figref{more_inpainting}.

\begin{figure*}[h]
	\centering
	\includegraphics[width=1.9\columnwidth]{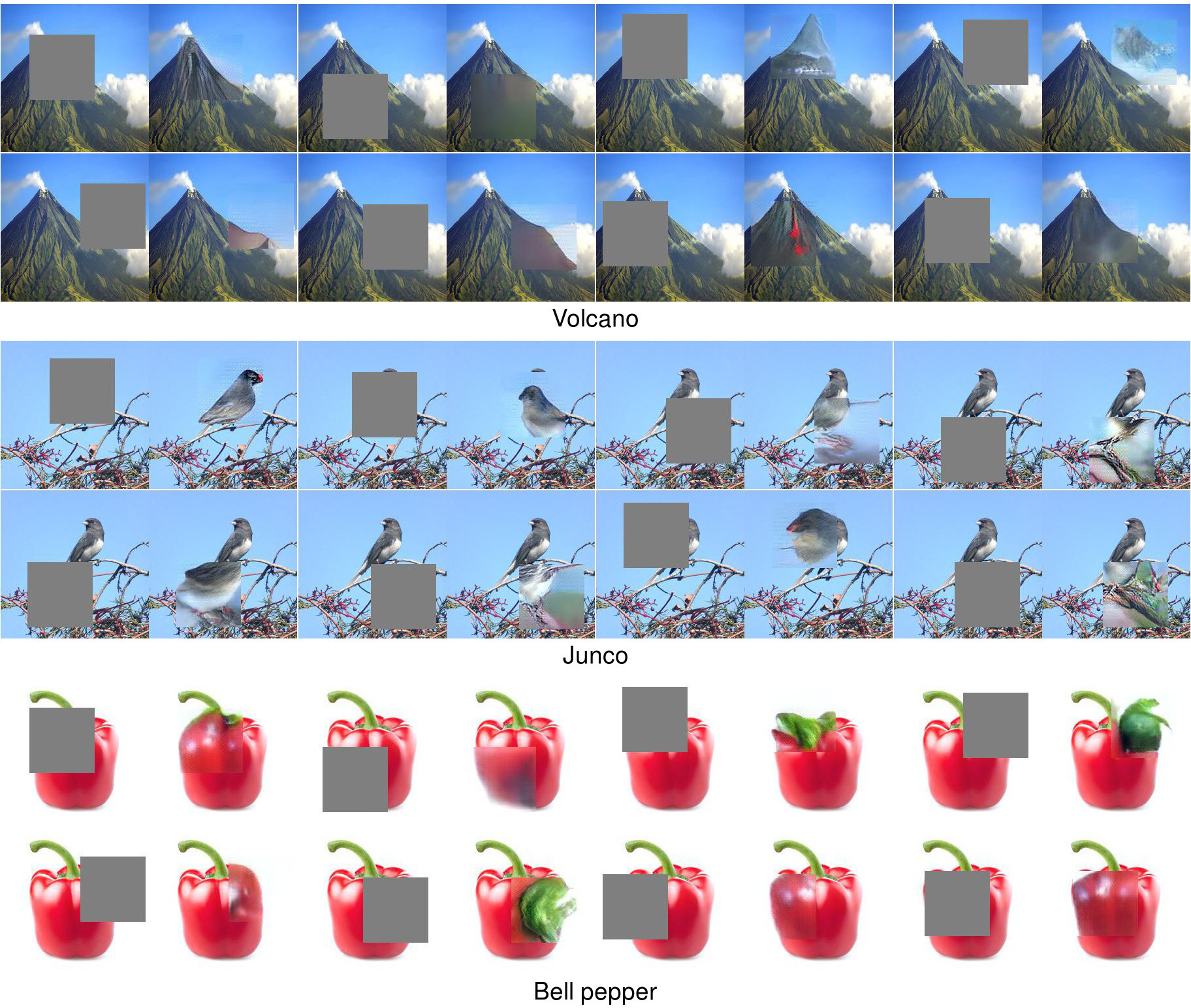}
	\caption{
		To test the ability of PPGNs to perform ``inpainting'', we randomly mask out a $100\times100$ patch in a real image, and perform class-conditional image sampling via PPGN-context (described in \secref{inpainting}) to fill in missing pixels. In addition to conditioning on a specific class (here, ``volcano'', ``junco'' and ``bell pepper'' respectively), we put an additional constraint on the code $h$ that it has to produce an image that matches the context region.
		PPGN-context performs semantically well in many cases. However, sometimes it does not match the local features of the surrounding regions. The result shows that the class-conditional image model does understand the semantics of images.
	}
	\figlabel{more_inpainting}
\end{figure*}

\section{PPGN-$x$: DAE model of $p(x)$}
\label{sec:SI_ppgn_x}

We investigate the effectiveness of using a DAE to model $p(x)$ directly (\figref{concept}a).
This DAE is a 4-layer convolutional network trained on unlabeled images from ImageNet. We sweep across different noise amounts for training the DAE and empirically find that a noise level of $20\%$ of the pixel value range, corresponding to $\epsilon_3 = 25.6$, produces the best results.
Full training and architecture details are provided in \secref{train_x_DAE}.

We sample from this chain following \eqnref{update_rule_DAE} with $(\epsilon_1, \epsilon_2, \epsilon_3) = (1, 10^5, 25.6)$\footnote{ The $\epsilon_1$ and $\epsilon_3$ correspond to the noise level used while training the DAE, and the $\epsilon_2$ value is chosen manually to produce the best samples.} and show samples in \figsref{sampling_x_image}{sampling_x_random}. 
PPGN-$x$
exhibits two expected problems: first, it models the data distribution poorly, evidenced by the images becoming blurry over time. Second, the chain mixes slowly, changing only slightly in hundreds of steps. 

Note that, instead of training the above DAE, one can also form an $x$-DAE by combining a pair of separately trained encoder $E$ and a generator $G$ into a composition $E(G(.))$. We also experiment with this model and call it Joint PPGN-$x$. The details of network $E$ and $G$ and how they can be combined are described in  Sec.~\ref{sec:default_ppgn} (Joint PPGN-$h$). For sampling, we sample in the image space, similarly to the PPGN-$x$ in this section. We found that Joint PPGN-$x$ model performs better than PPGN-$x$, but worse than Joint PPGN-$h$ (data not shown).

\section{Why PPGNs produce high-quality images}
\label{sec:discuss_ppgn}

One practical question is why Joint PPGN-$h$ produces high-quality images at a high resolution for 1000-class ImageNet more successfully than other existing latent variable models~\cite{odena-2016-arXiv-conditional-image-synthesis,salimans-2016-arXiv-improved-techniques-for-training,radford-2015-arXiv-unsupervised-representation-learning}. We can consider this question from two perspectives. 

First, from the perspective of the training loss, $G$ is trained with the combination of three losses (\figref{training_diagram}b), which may
be a beneficial approach to model $p(x)$. 
The GAN \cite{goodfellow2014generative-adversarial-networks} loss, which is the gradient of $\log(1-D(x))$, that is used to train $G$ 
pushes each reconstruction $G(h)$
toward a mode of real images $p_{\textrm{data}}(x)$ and away from the current reconstruction distribution. This can be seen by noting that the Bayes optimal $D$ is $p_{\textrm{data}}(x)/(p_{\textrm{data}}(x) + p_{\textrm{model}}(x))$ \cite{goodfellow2014generative-adversarial-networks}.
Since $x \sim G(h)$ is already near a mode of $p_{\textrm{model}}(x)$, the net effect is to push $G(h)$ towards one of the modes of $p_{\textrm{data}}$, thus making the reconstructions sharper and more plausible.
If one uses only the GAN objective and no reconstruction objectives ($L_2$ losses in the pixel or feature space), $G$ may bring the sample far from the original $x$, possibly collapsing several modes of $x$ into fewer modes.
This is the typical, known ``missing-mode'' behavior of GANs \cite{salimans-2016-arXiv-improved-techniques-for-training,goodfellow2014generative-adversarial-networks}
that arises in part because GANs
minimize the Jensen-Shannon divergence rather than Kullback-Leibler divergence between $p_{\textrm{data}}$ and $p_{\textrm{model}}$, leading to an over-memorization of modes \cite{theis-2016-ICLR-a-note-on-the-evaluation-of-generative}.
The reconstruction losses
are important to combat this missing mode problem and may also serve to enable better convergence of the feature space autoencoder to the distribution it models, which is necessary
in order to make the $h$-space reconstruction properly estimate $\partial \log p(h) / \partial h$ \cite{alain-2014-what-regularized-auto-encoders}.

Second, from the perspective of the learned $h \to x$ mapping, 
we train the $G$ parameters of the $E+G$ pair of networks as an $x$-AE, mapping $x \to h \to x$ (see \figref{training_diagram}b).
In this configuration, as in VAEs \cite{kingma-2013-arXiv-auto-encoding-variational-bayes} and regular DAEs \cite{vincent-2008-ICML-extracting-and-composing-robust}, the one-to-one mapping helps prevent the typical latent $\to$ input missing mode collapse that occurs in GANs, where some input images are not representable using any code \cite{goodfellow2014generative-adversarial-networks,salimans-2016-arXiv-improved-techniques-for-training}.
However, unlike in VAEs and DAEs, where the latent distribution is learned in a purely unsupervised manner, we leverage the labeled ImageNet data to train $E$ in a supervised manner that yields a distribution of features $h$ that we hypothesize to be semantically meaningful and useful for building a generative image model. To further understand the effectiveness of using deep, supervised features, it might be interesting future work to train PPGNs with other feature distributions $h$ such as random features or shallow features (e.g. produced by PCA).

\begin{table*}[]
	\centering
	\begin{tabular}{lccccc}
		\hline
		\textbf{Model}                 & \textbf{Image size} & \textbf{Inception accuracy} & \textbf{Inception score} & \textbf{MS-SSIM} & \textbf{Percent of classes} \\ \hline
		Real ImageNet images              & $256\times256$             & 76.1\%                      & 210.4 $\pm$ 4.6              & 0.10 $\pm$ 0.06       & 999 / 1000                  \\
		AC-GAN \cite{odena-2016-arXiv-conditional-image-synthesis}                         & $128\times128$             & 10.1\%                      & N/A                      &N/A              & 847 / 1000                  \\
		PPGN                           & $256\times256$             & 59.6\%                      & 60.6 $\pm$ 1.6               & 0.23 $\pm$ 0.11      & 829 / 1000                  \\
		PPGN samples resized to $128\times128$ & $128\times128$             & 54.8\%                      & 47.7 $\pm$ 1.0               & 0.25 $\pm$ 0.11      & 770 / 1000                  \\ \hline
	\end{tabular}
	\caption{A comparison between real ImageNet validation set images, AC-GAN \cite{odena-2016-arXiv-conditional-image-synthesis} samples, PPGN samples and their resized $128\times128$ versions. Following the literature, we report Inception scores \cite{salimans-2016-arXiv-improved-techniques-for-training} (higher is better) and Inception accuracies \cite{odena-2016-arXiv-conditional-image-synthesis} (higher is better) to evaluate sample quality, and MS-SSIM score \cite{odena-2016-arXiv-conditional-image-synthesis} (lower is better), which measures sample diversity within each class. As in Odena et al. \cite{odena-2016-arXiv-conditional-image-synthesis}, the last column (``Percent of classes'', higher is better) shows the number of classes that are more diverse (by MS-SSIM metric) than the least diverse class in ImageNet. 
	Overall, PPGN samples are of substantially higher quality quality than AC-GAN samples (by Inception accuracy, i.e. PPGN samples are far more recognizable by the Google Inception network \cite{szegedy-2016-arXiv-inception-v4-inception-resnet-and-the-impact} than AC-GAN samples). Their diversity scores are similar (last column, $847 / 1000$ vs. $829 / 1000$).
	However, by all 4 metrics, PPGN samples have substantially lower diversity and quality than real images. This result aligns with our qualitative observations in \figsref{60_gen_vs_60_real}{DGNAM_vs_PPGN}.
	\newline\underline{Row 2:} Note that we chose to compare with AC-GAN \cite{odena-2016-arXiv-conditional-image-synthesis} because, this model is also class-conditional and, to the best of our knowledge, it produces the previous highest resolution ImageNet images ($128\times128$) in the literature. 
	\newline\underline{Row 3:} For comparison with ImageNet $256\times256$ images, the spatial dimension of the samples from the generator $G$ is $256\times256$ and we did not crop it to $227\times227$ as done in other experiments in the paper.
	\newline\underline{Row 4:} Although imperfect, we resized PPGN $256\times256$ samples down to $128\times128$ (last row) for comparison with AC-GAN. 	
	}
	\label{table:quantitative}
\end{table*}

\begin{figure*}
	\centering
	\begin{subfigure}{1.0\linewidth}
		\centering
		\includegraphics[width=0.95\columnwidth]{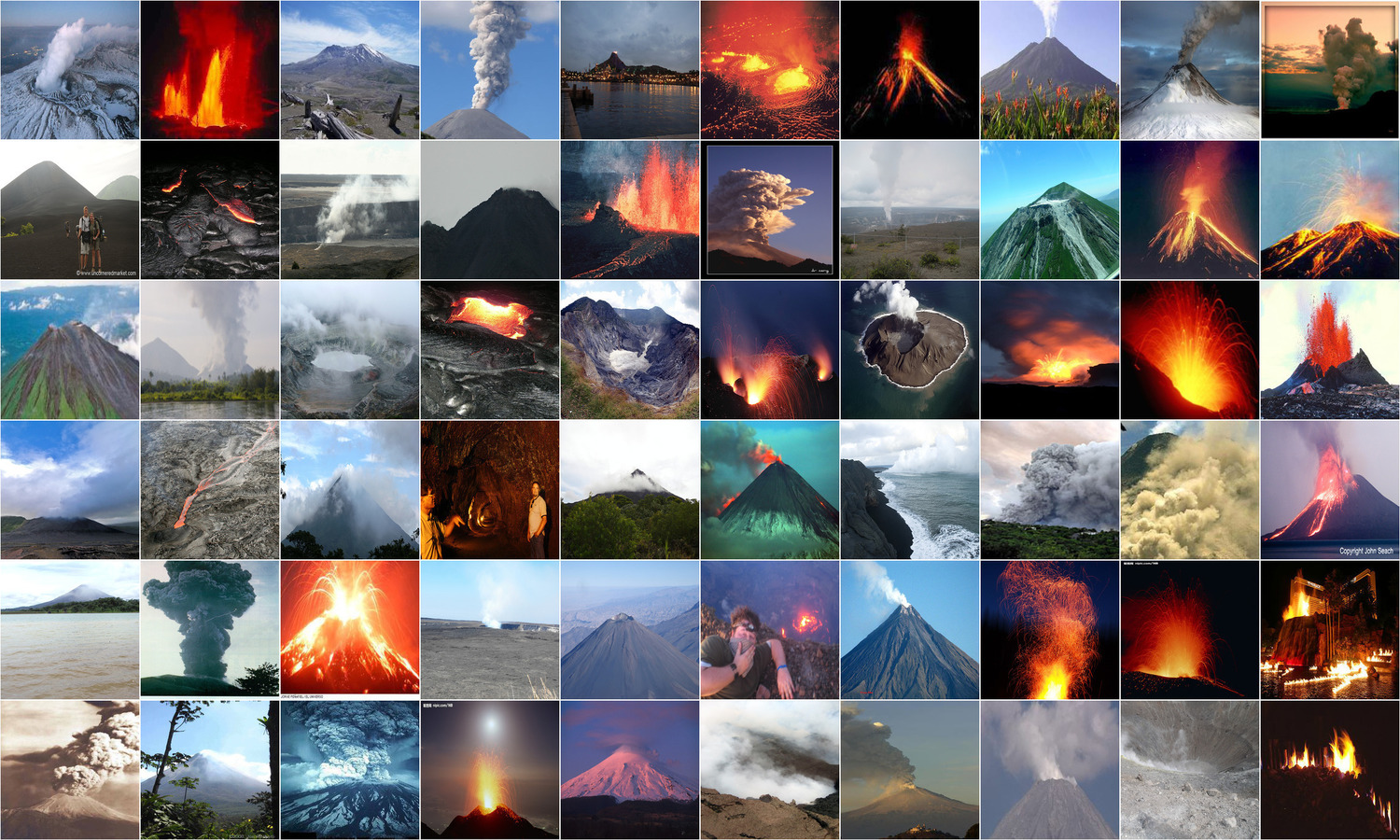}
		\caption{60 training set images randomly taken from the ``volcano'' class.}.
		\vspace{0.2cm}
	\end{subfigure}		
	\hspace{1mm}
	\begin{subfigure}{1.0\linewidth}
		\centering
		\includegraphics[width=0.95\linewidth]{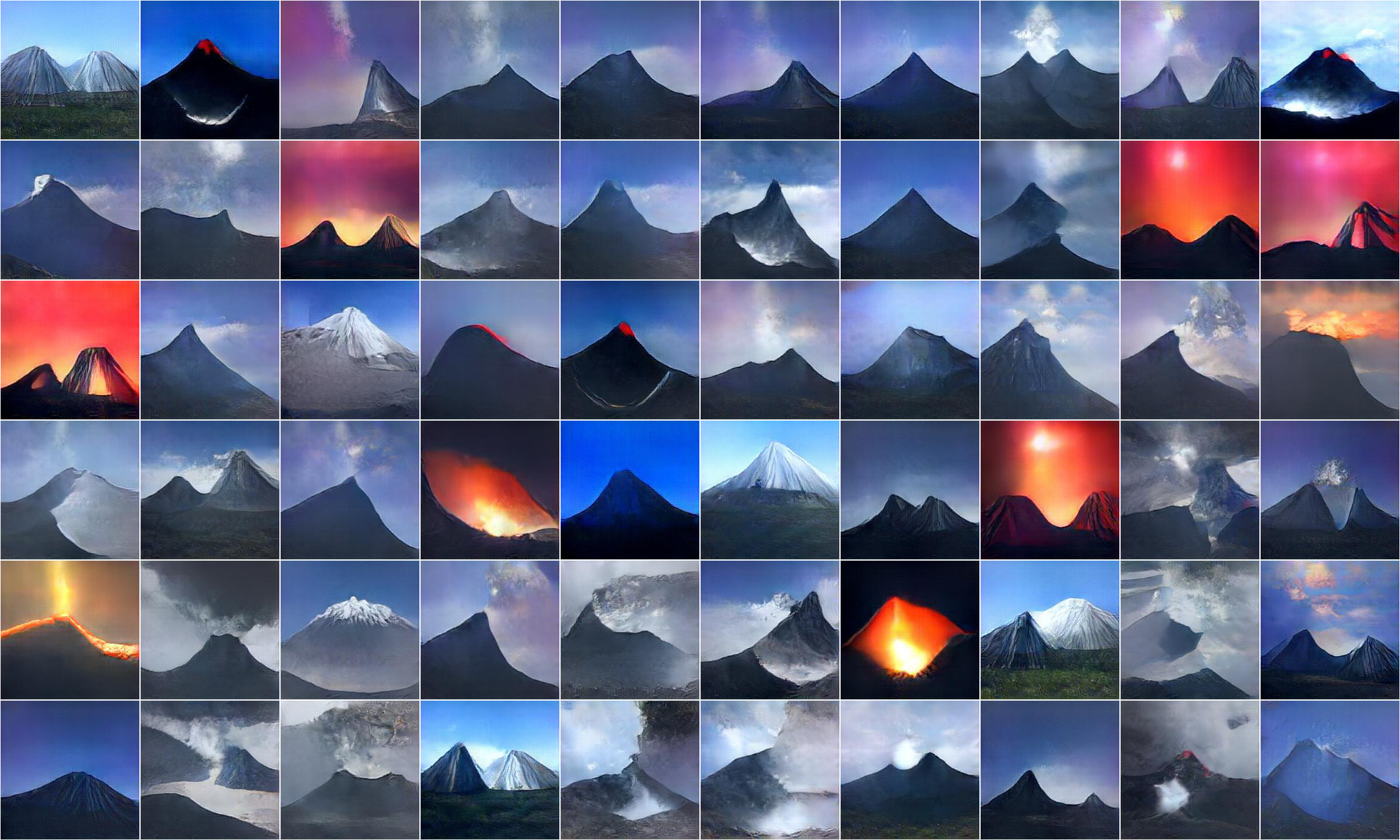}
		\caption{60 PPGN samples randomly selected from 2000 samples, which are produced from 10 200-step sampling chains.}
		\vspace{0.2cm}
	\end{subfigure}	
	
	\caption{
		To evaluate how well PPGN samples represent the training set images, we compare 60 real images (top) vs. 60 PPGN generated images (bottom). All images are randomly selected. While the set of generated images are high-quality, they have a much lower diversity compared to the set of real images. This observation aligns with our quantitative measure in Table~\ref{table:quantitative}.
	}
	\figlabel{60_gen_vs_60_real}
	\vspace{0em}
\end{figure*}

\end{document}